\pdfoutput=1
\documentclass{article}

\usepackage[nonatbib, final]{neurips_2023}

\usepackage[english]{babel}

\usepackage{amsmath}
\usepackage{booktabs}
\usepackage{comment}
\usepackage{graphicx}
\usepackage{subcaption}
\usepackage{tikz}
\usepackage[colorlinks=true, allcolors=blue]{hyperref}
\usepackage[para]{footmisc}
\usepackage[para]{footmisc}
\usepackage{ragged2e}
\usepackage{lipsum}
\usepackage{mdframed}
\usepackage{tcolorbox}
\usepackage{framed}

\newtcolorbox{formattedquote}{colback=blue!3!white, colframe=blue!20!white, fontupper=\ttfamily\small}

\newtcolorbox{formattedresponse}{colback=yellow!3!white, colframe=yellow!50!white, fontupper=\ttfamily\small}

\newtcbox{\mybox}[1][blue]{on line, arc=3pt, colback=blue!3!white, fontupper=\ttfamily\small, colframe=blue!20!white, boxsep=0pt, left=1pt, right=1pt, top=2pt, bottom=2pt, boxrule=0.5pt}

\title{GPT4GEO: How a Language Model \\ Sees the World's Geography}
\author{%
  Jonathan Roberts \\
  CAML Lab\\
  University of Cambridge\\
  \texttt{jdr53@cam.ac.uk} \\
  \And
  Timo L{\"u}ddecke \\
  Institute of Computer Science and CIDAS \\
  University of G{\"o}ttingen \\
  \texttt{timo.lueddecke@uni-goettingen.de}
  \And
  Sowmen Das \\
  CAML Lab\\
  University of Cambridge\\
  \texttt{sd973@cam.ac.uk} \\
  \And
  Kai Han \\
  Visual AI Lab \\
  The University of Hong Kong \\
  \texttt{kaihanx@hku.hk} \\
  \And
  Samuel Albanie \\
  CAML Lab\\
  University of Cambridge\\
  \texttt{samuel.albanie.academic@gmail.com} \\
}

\begin{document}
\maketitle

\begin{abstract}

Large language models (LLMs) have shown remarkable capabilities across a broad range of tasks involving question answering and the generation of coherent text and code.
Comprehensively understanding the strengths and weaknesses of LLMs is beneficial for safety, downstream applications and improving performance.
In this work, we investigate the degree to which GPT-4 has acquired factual geographic knowledge and is capable of using this knowledge for interpretative reasoning, which is especially important for applications that involve geographic data, such as geospatial analysis, supply chain management, and disaster response.
To this end, we design and conduct a series of diverse experiments, 
starting from factual tasks such as location, distance and elevation estimation to more complex questions such as generating country outlines and travel networks, route finding under constraints and supply chain analysis. 
We provide a broad characterisation of what GPT-4 (without plugins or Internet access) knows about the world, highlighting both potentially surprising capabilities but also limitations. 
\end{abstract}

\section{Introduction}

In recent years, large language models (LLMs), such as the Generative Pre-trained Transformer (GPT) series \cite{radford2018improving,radford2019language,brown2020language,openai2023gpt4}, LLaMA \cite{touvron2023llama}, PaLM \cite{chowdhery2022palm}, and BLOOM \cite{scao2022bloom}, have demonstrated remarkable capabilities in understanding and generating natural language text across a broad range of tasks. The recently released GPT-4 \cite{openai2023gpt4} model displays a broad range of skills, arguably significantly beyond those of its predecessors and contemporaries alike, opening up new avenues for application in both commercial and scientific domains. 

Despite only being publicly released less than two months ago, numerous studies have been carried out revealing the ability of GPT-4 to perform a diverse set of tasks such as generating artwork, writing code \cite{bubeck2023sparks}, search \cite{sun2023chatgpt}, recipe generation \cite{noever2023multimodal}, %
mathematics \cite{yuan2023well} and architecture optimisation \cite{zheng2023can}, as well as achieve strong performance on many examinations \cite{openai2023gpt4, nori2023capabilities}.
However, there is yet to be a detailed investigation into what the model knows about the world's geography.

A comprehensive understanding of the geographic capabilities of GPT-4 is important for: (1) \textbf{Safety}. As AI models become more powerful, so do the potential dangers and safety risks \cite{amodei2016concrete}. An awareness of the full range of the acquired knowledge and skills of GPT-4 is essential to safe, widespread deployment in society. Similarly, an understanding of when the model hallucinates \cite{ji2023survey} %
is critical to safe and reliable usage. (2) \textbf{Improvements}. Accurate knowledge of cases where GPT-4 succeeds and fails are vital to drive research in model training and architecture. (3) \textbf{Downstream applications}. Only after a reasonable understanding of GPT-4's abilities is it possible to maximally leverage it for downstream tasks, applications and products. Strong geographic knowledge and skills would enable many commercial opportunities in the travel and navigation sectors as well as boost research, for example in environmental sciences. 

\begin{figure}
    \centering
    \includegraphics[width=\textwidth]{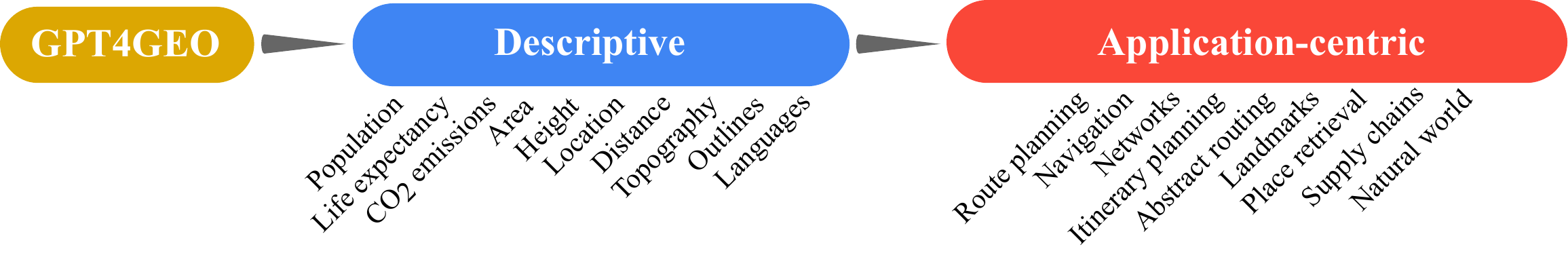}
    \vspace{-5mm}
    \caption{\textbf{GPT4GEO experiments taxonomy}. 
    We initially focus on factual tasks before moving towards application-centric tasks requiring logic and reasoning that build on this knowledge.}
    \label{fig:taxonomy}
\end{figure}

Despite these compelling incentives, actually characterising the capabilities of GPT-4, even in a specific domain, is challenging for a number of reasons: (i) \textbf{Closed source}. Only limited technical details of the GPT-4 model have been publicly released, making it difficult to estimate capabilities via scaling laws. (ii) \textbf{Training volume}.
It is highly likely GPT-4 was trained on a very large-scale text corpus. Even if details of this data were available, given the scale, it would be challenging to fully understand the distribution of content 
and correspondingly infer specifics about the knowledge acquired by the model. (iii) \textbf{Breadth of capability}. As models become more generalised, the range of possible tasks they can perform greatly increases. (vi) \textbf{Combinatorial explosion}. The scale of the diversity of the world's geography means that attaining a comprehensive understanding of even a subset of geographic capability requires experimentation across many different factors, quickly resulting in an unfeasible number of permutations \cite{10.5555/2678074}. 

In this work, we provide a broad profile of what GPT-4 knows about the world. To overcome the challenges of characterising the capability of an LLM like GPT-4, we design experiments of varying degrees of depth, providing a mixture of both qualitative and quantitative results. We initially probe low-level \textit{descriptive} knowledge before transitioning towards more complex, \textit{application-centric} geographic tasks that build upon the factual knowledge and incorporate it into logical answers.

\section{Related Work}

\paragraph{GPT-4 capabilities.}
Recent works have characterised GPT-4's ability to solve tasks within a narrow focus. In the computer science domain, GPT-4 has been tasked with coding \cite{poldrack2023ai}, information retrieval \cite{sun2023chatgpt}, spear phishing \cite{hazell2023large}, and neural architecture search \cite{zheng2023can}. Many works have explored the performance of GPT-4 in the medical domain \cite{dash2023evaluation, 
giorgi2023clinical, tang2023gersteinlab, kasai2023evaluating, wang2023large}. Other areas investigated include mathematics \cite{yuan2023well}, education \cite{kortemeyer2023can} and literature \cite{chang2023speak, garrido2023simulating}, as well as receipe generation \cite{noever2023multimodal}, climate science \cite{vaghefi2023chatclimate} and logical reasoning \cite{liu2023evaluating}. More generalised profiles of GPT-4's capability are provided by OpenAI's technical report \cite{openai2023gpt4} and \cite{bubeck2023sparks}, from which we draw inspiration. 
In \cite{openai2023gpt4}, numerous test scores are reported, including the AP Environmental Science exam, in which GPT-4 scores in the 91st percentile. 
This encompasses a small amount of geographical awareness combined with environmental knowledge but covers entirely different question types to our experiments.
Concurrent work \cite{li2023autonomous} creates a prototype system using GPT-4 as the reasoning core of an autonomous geospatial agent; however, system design is central to this work and geographic capability is only peripherally probed. Our work focuses on providing a broad profile of what GPT-4 knows about the world.

\paragraph{Geographic capabilities of other language models.}
Cultural commonsense knowledge research indirectly covers geographic knowledge, with the focus being evaluating the \textit{cultural knowledge} of language models, such as GPT-3 \cite{nguyen2022extracting} or other multilingual models \cite{yin2022geomlama}. Query point-of-interest matching is more directly related to geography with a focus on location. Pretraining language models with geographic context \cite{ding2023multi, huang2022ernie} results in relatively strong performances in various location-based geographic tasks. 
Leveraging the superior capabilities of GPT-4, our work includes these tasks as a subset of a wide selection of geographic knowledge capability characterisation.

\section{Method}

\subsection{GPT-4} We use the GPT-4 model \cite{openai2023gpt4} in all of our experimentation. Citing competitive and safety considerations, OpenAI has not released technical details for GPT-4. However, it is likely GPT-4 follows (in some form) the same 2-stage training strategy of previous models: (1) \textbf{Training}. Next word prediction training from a large corpus of text taken from the Internet. (2) \textbf{Fine-tuning}. Fine-tuning via Reinforcement Learning from Human Feedback. %
For this work, we interact with GPT-4 using three methods: the ChatGPT interface\footnote{https://chat.openai.com}, the OpenAI Playground\footnote{https://platform.openai.com} and API\footnote{https://platform.openai.com/docs/api-reference}, using the default settings.

For each experiment, we provide an \mybox{example of the prompt structure} used to query GPT-4, and include details of our experimental setup, ground-truth data and specific prompts in the \hyperref[app]{Appendix}. For each experiment we generate all 
responses from the same GPT-4 instance, and unless otherwise stated, we use outputs from single prompts. For a discussion of reproducibility see Sec. \ref{sec:iand_fd}.

\subsection{Experimental Design}
\label{sec:experimental_design}

To characterise what GPT-4 knows about the world, we devise a set of progressively more challenging experiments that aim to provide a broad profile of capabilities across key geographic aspects. Due to the breadth and complexity of the world's geography, coupled with GPT-4's stochasticity, we curate a representative set of quantitative and qualitative experiments, see Fig. \ref{fig:taxonomy}.

\textbf{Descriptive}. We begin with low-level tasks probing factual knowledge that is essential for downstream applications. 
We arrange these tasks according to increasing difficulty, progressing from simple estimations toward more complex topography and mapping.

\textbf{Application-centric}. Building on the prior experiments, we investigate GPT-4's ability to leverage the acquired descriptive knowledge for application-centric reasoning tasks. We extensively explore travel and navigation, as well as, supply chains, networks, wildlife ranges and many more.

\textbf{Evaluation}. We use a range of data sources to determine either qualitatively or quantitatively (depending on the nature of each experiment) the degree to which GPT-4 is correct. For geospatial experiments, we frequently leverage the Cartopy library\footnote{https://scitools.org.uk/cartopy/} and Google Maps as a source of ground truth. For the quantiative experiments, we report the relative error\footnote{$RE = 100\% \cdot \frac{|x_t - x_p|}{x_t}$, where $x_t$ and $x_p$ are the true and predicted values.}.

\section{Experiments}

\subsection{Descriptive}

We initially investigate GPT-4's ability to solve descriptive tasks by conducting simple spatial and human knowledge retrieval experiments. Next, we increase the complexity and explore more challenging tasks that require additional reasoning. Finally, we discuss our findings.

\begin{figure}[]
    \centering
    \begin{subfigure}{0.32\textwidth}
        \raggedright
        \includegraphics[height=4.2cm]{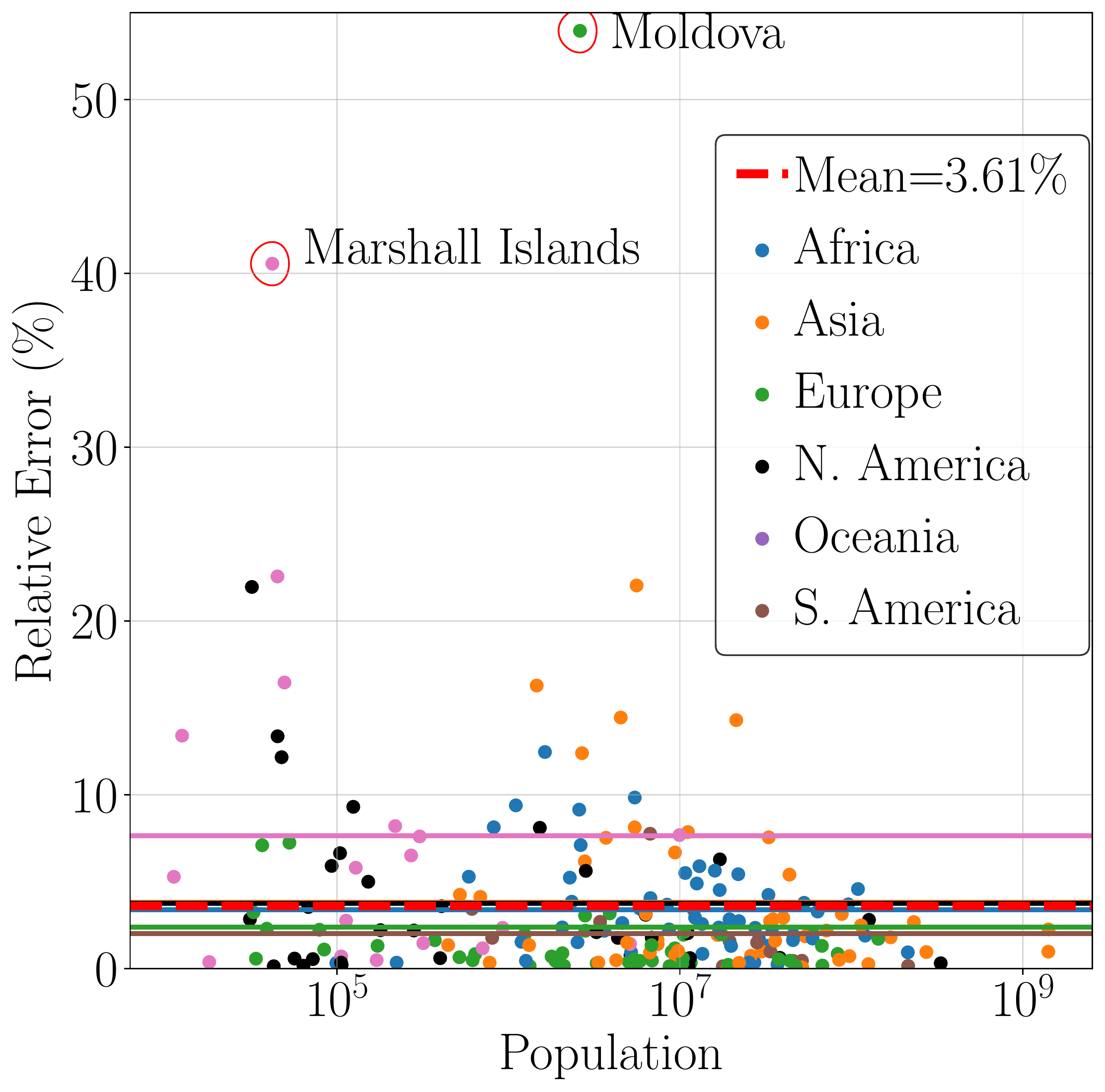}
        \caption{Population (2021).}
        \label{fig:population}
    \end{subfigure}
    \begin{subfigure}{0.32\textwidth}
        \centering
        \includegraphics[height=4.2cm]{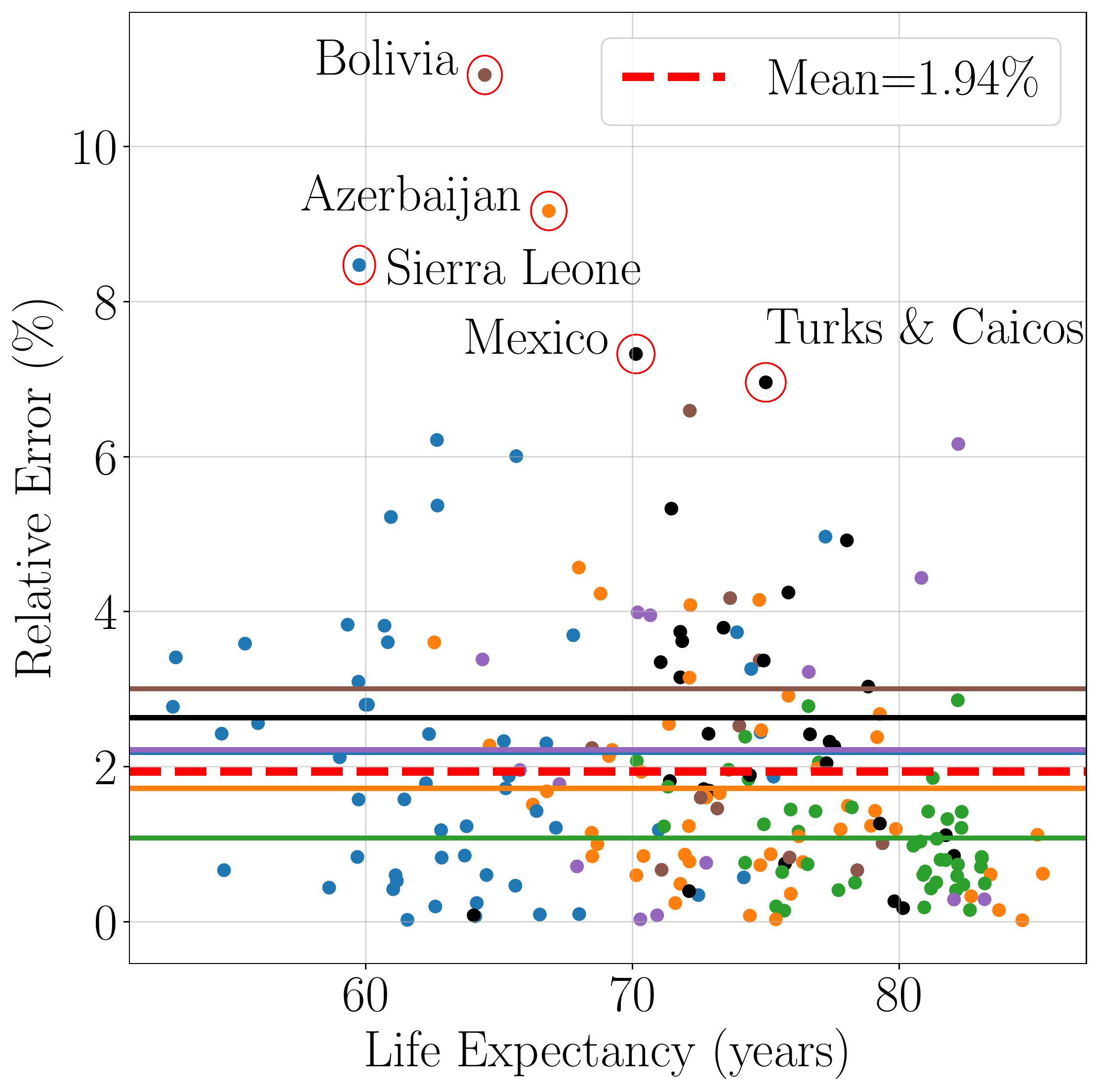}
        \caption{Life Expectancy (2020).}
        \label{fig:life_expectancy}
    \end{subfigure}
    \begin{subfigure}{0.32\textwidth}
        \raggedleft
        \includegraphics[height=4.2cm]{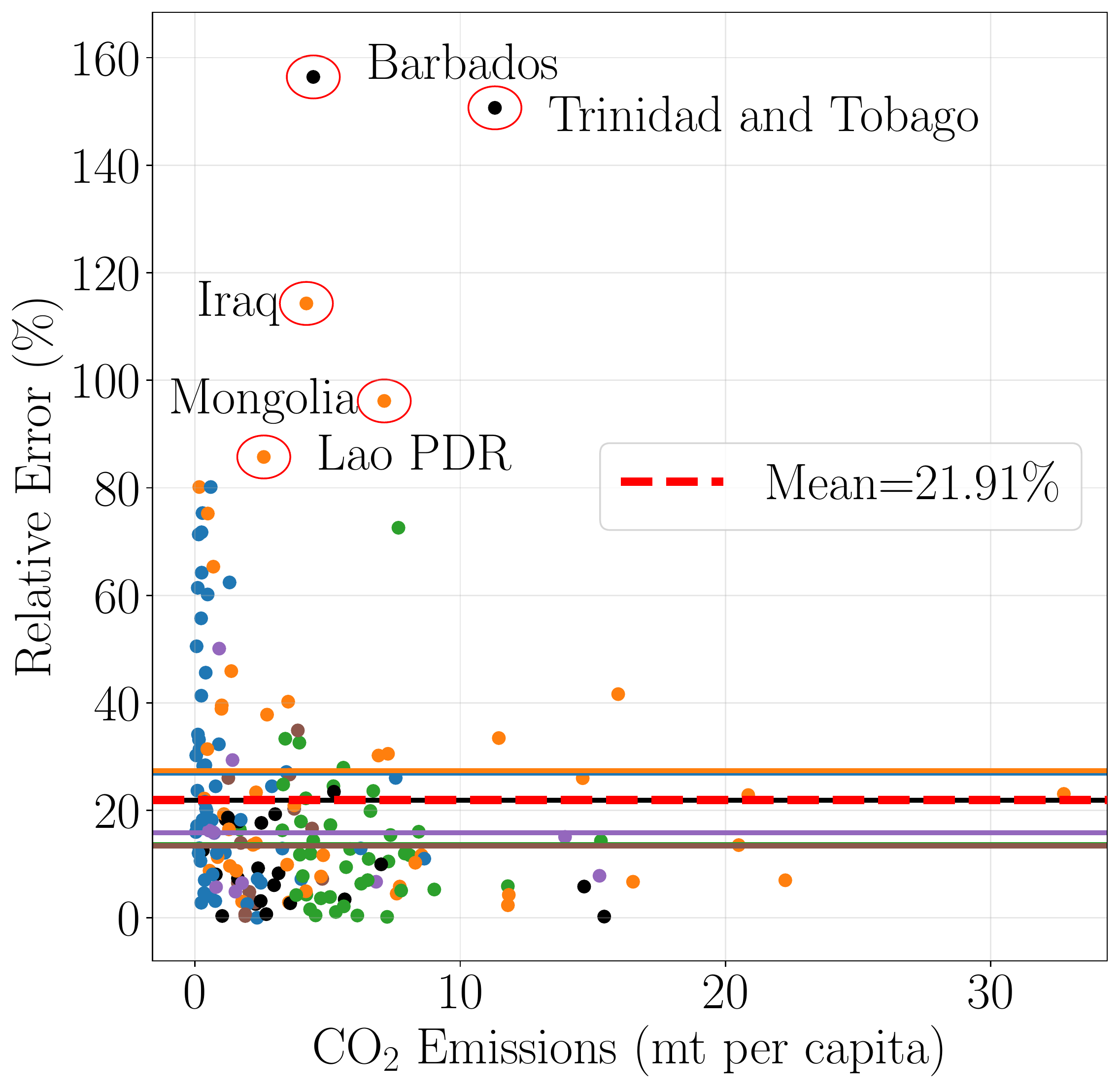}
        \caption{CO\textsubscript{2} Emissions (2019).}
        \label{fig:co2}
    \end{subfigure}
\caption{\textbf{Socioeconomic indicators}. A quantitative evaluation of GPT-4's understanding of country-level human populations and their impact on the environment, including $-$ (a) country populations, (b) life expectancies, and (c) CO\textsubscript{2} emissions per capita. The red circles denote outliers.}
\label{fig:human_features}
\end{figure}

\textbf{Population, life expectancy and CO\textsubscript{2} emissions}. We evaluate GPT-4's understanding of country-level socioeconomic indicators -- i.e., population, life expectancy and CO\textsubscript{2} emissions -- by calculating the relative error against ground-truth data from \cite{worldbank2023pop}, \cite{worldbank2023le} and \cite{worldbank2023co2}, respectively (see Fig. \ref{fig:human_features}). For populations, GPT-4 performs relatively well with a mean relative error (MRE) of 3.61\%. However, significantly higher errors are recorded for less populated countries. For country life expectancies, GPT-4 performs even better, with an MRE $<$2\% and a worst error of just over 10\%. GPT-4's estimations for CO\textsubscript{2} emissions per capita are an order of magnitude worse with an MRE of $>$20\% and individual errors of $>$150\% for two countries. We observe minor variations in MRE across continents for each experiment, however, there is no consistent overall trend. We use the same format of prompt for each experiment, e.g., for population:

\begin{formattedquote}
    For each of the following countries, provide their population in \textit{<Year>} as a python list in the following format: 
    \\\
    [Population\_of\_Country\_1, \# Country 1 \\
    Population\_of\_Country\_2, \# Country 2, ...]
    \\ \\
    \textit{[<Country\_1>, <Country\_2>, ...]}
\end{formattedquote}
Where we obtain \textit{\textless Country\_Names\textgreater} from the respective ground truths.

\begin{figure}
    \centering
    \begin{subfigure}{0.32\textwidth}
        \raggedright
        \includegraphics[height=4.2cm]{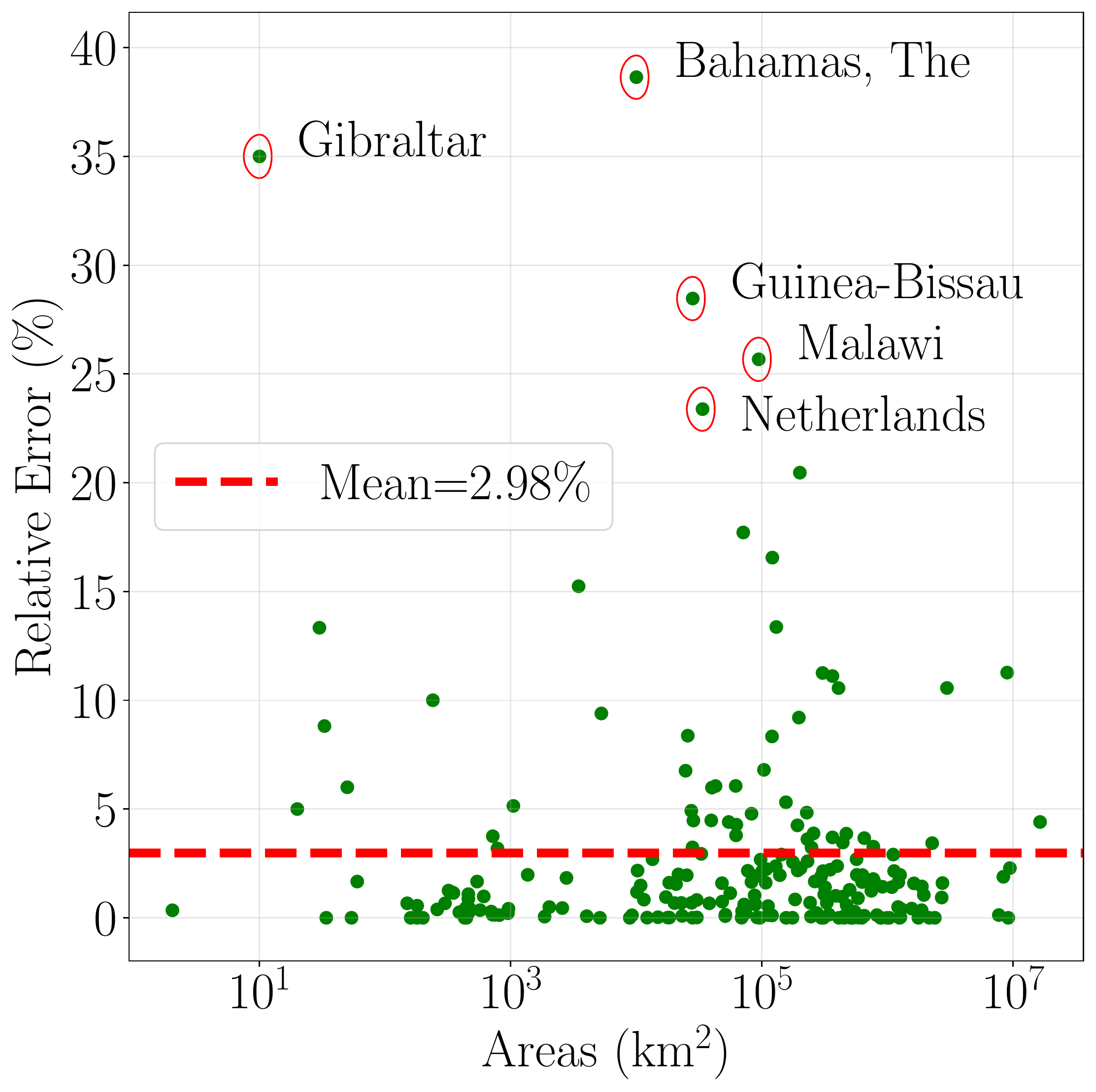}
        \caption{Area.}
        \label{fig:country_areas}
    \end{subfigure}
    \begin{subfigure}{0.32\textwidth}
        \centering
        \includegraphics[height=4.2cm]{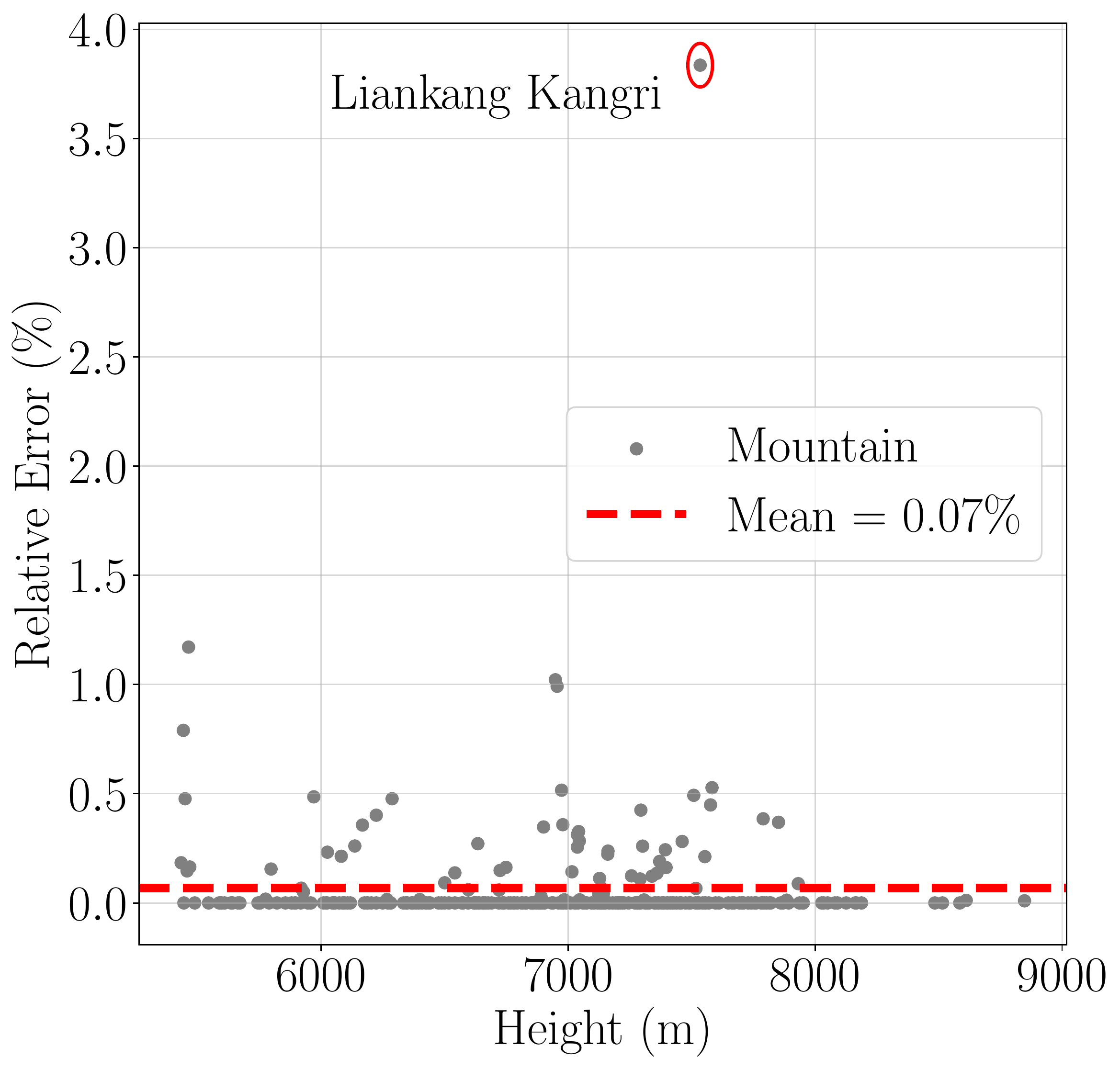}
        \caption{Height.}
        \label{fig:mountain_height}
    \end{subfigure}
    \begin{subfigure}{0.32\textwidth}
        \raggedleft
        \includegraphics[height=4.2cm]{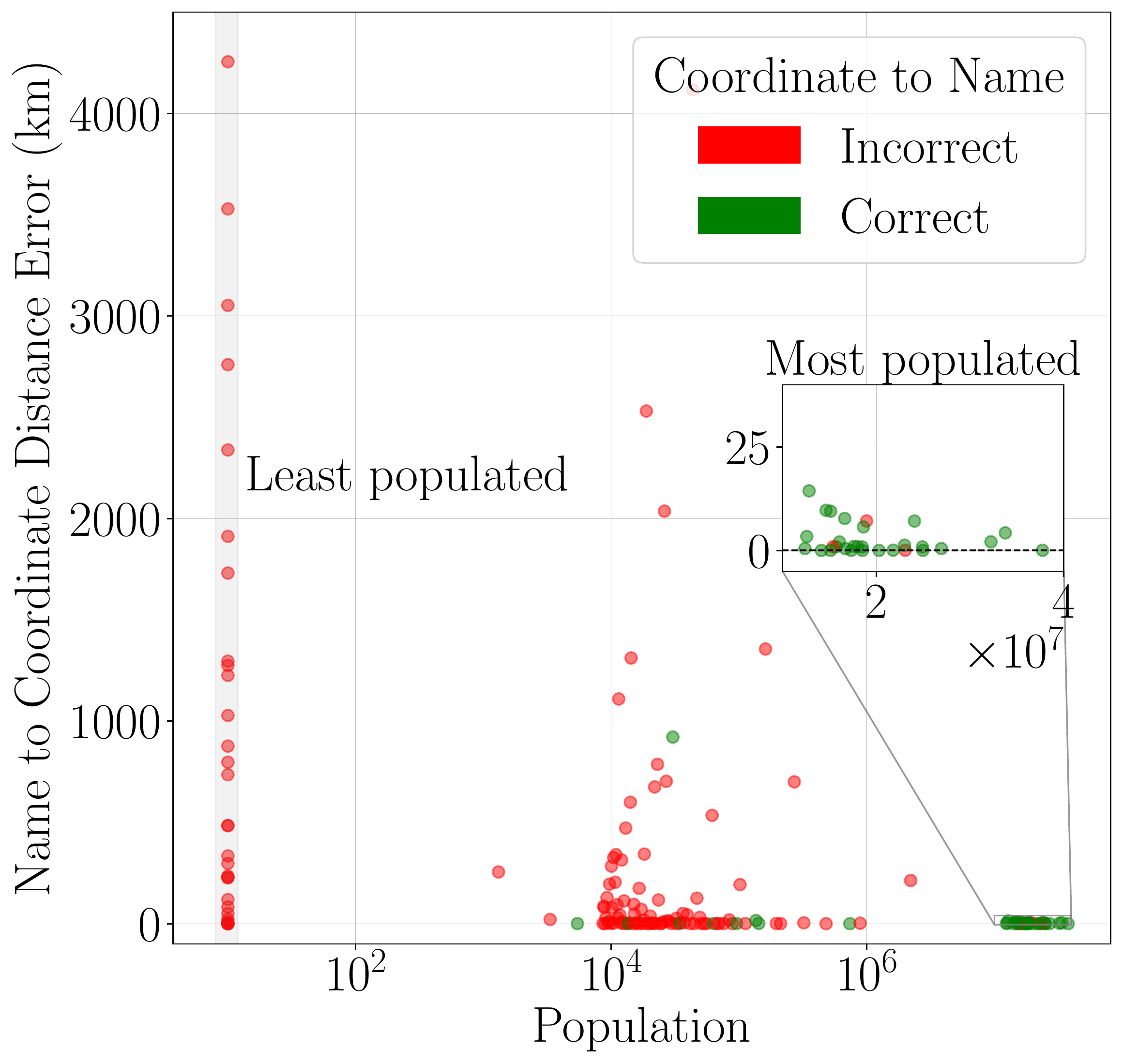}
        \caption{Location.}
        \label{fig:settlement_location}
    \end{subfigure}
    \caption{\textbf{Spatial features}. An evaluation of GPT-4's understanding of
     country areas (a), heights of the 300 tallest mountains (b), and locations of settlements of different populations (c).}
    \label{fig:geographic_features}
\end{figure}

\textbf{Area and height}. We evalute GPT-4's knowledge of country areas and heights of the 300 tallest mountains by calculating the relative error against ground-truth data from \cite{worldbank2023area} and \cite{kaggle_mountains}, respectively -- see Figs. \ref{fig:country_areas}-\ref{fig:mountain_height}. For country areas, GPT-4 attains an MRE of $\sim$3\% with reasonable spread in accuracy, and relative errors of $>$20\% for 6 countries. Very strong performance is shown for the mountain heights, with an MRE of 0.07\% and only one outlying error at $\sim$4\%. We prompt GPT-4 for areas and heights in the same way as the socioeconomic indicators, e.g., for areas:

\begin{formattedquote}
    For each of the following countries, provide the land area in sq. km as of \textit{<Year>}. Provide the areas as a python list in the following format:
    \\\
    [Area\_of\_Country\_1, \# Country 1 \\
    Area\_of\_Country\_2, \# Country 2, ...]
    \\ \\
    \textit{[<Country\_1>, <Country\_2>, ...]}
\end{formattedquote}

\textbf{Location}. 
We compile a set of the 30 most and least populated settlements, as well as a representative sample of 100 settlements with populations distributed in between (all data is taken from \cite{kaggle_cities}). We conduct two experiments, the results for which are displayed in Fig. \ref{fig:settlement_location}. First, we provide the settlement names to GPT-4, asking for the coordinates of each and we calculate the distance error to the true coordinates using the haversine formula. We observe the location accuracy clearly decreases with decreasing settlement population, with the worst case out by 4,000 km. Next, we conduct the reverse -- providing the coordinates to GPT-4 and asking for the names. This proves much more difficult, with incorrect settlement names predicted in most cases (red points). For the Name $\rightarrow$ Coordinates setting, we query GPT-4 using the following prompt (and use the reverse for Coordinate $\rightarrow$ Name setting):

\begin{formattedquote}
    In a code block, provide a python list of tuples for the latitude and longitude coordinates for each of these settlements - e.g., [(Lat,Lon), \# Settlement 1 ...]. Maintain the same order.
    \newline
    \textit{[<Country\_1>, <Country\_2>, ...]}
\end{formattedquote} 

\textbf{Distance}. We obtain coordinates and population of cities from the GeoNames databases \cite{geonames}. We randomly sample 40 city pairs from each continent and population group and ask GPT-4 to estimate the corresponding distances. We calculate the relative error between the prediction and ground truth distances using the haversine formula from the coordinates. We used the following prompt for distance estimation. In cases where the number of outputs did not match the number of provided cities, we repeated the prompt.
\begin{formattedquote}
    What are the straight-line distances between the following pairs of cities? Output a csv list of the distances only in km.\\
    <\textit{City 1 (Country)}> to \textit{<City 2 (Country)>}\\
    ... \textit{[repeated for all 40 pairs]}
\end{formattedquote}

Our results (Tab.~\ref{tab:distances}) indicate mediocre performance for distance estimation. Average errors can exceed 50\% for small cities but are generally lower for larger populations. A limiting factor for these experiments is that the selection of cities influences the error substantially.
For a different sample of $>100K$ city pairs in Europe we obtained an average error of 22.8 $\pm 3.5$, for reference randomly shuffling the ground truth distances in this setting results in a relative error of 152 \%.

\def\rerun{}

\begin{table}[h!]
    \centering
    \footnotesize
    \resizebox{.6\textwidth}{!}{%
    \begin{tabular}{lllllll}
        \toprule
        Continent & \multicolumn{4}{c}{Population} \\
         & $<$ 20K & $<$ 100K & $<$ 500K ($\pm \sigma_5$) &  $>$ 500K  \\
         \midrule
         Europe & 17.2* & 19.3 & 12.4 $\pm 0.8$* & 14.2  \\
         North America & 51.0\rerun{} & 51.7 & 11.1 $\pm 0.7$ & 12.6\rerun{}  \\
         South America & 32.0 & 44.5\rerun{} & 28.6 $\pm 6.8$* & 20.4\rerun{} \\
        Asia & 33.7 & 27.3 & 24.8 $\pm 2.2$ & 18.3\rerun{}   \\
        Africa & 33.4\rerun{} &  25.9 & 23.5 $\pm 1.5$ & 18.3  \\
        Oceania &  21.1* & 25.7* & 12.4 $\pm 2.9$* & 0.6* \\
         \bottomrule \\
    \end{tabular}%
    }
    \hspace{0.2cm}
    \begin{tabular}{p{4cm}}
        * changed prompt since ChatGPT became reluctant to provide distances (possibly due to an updated configuration). \\
        \midrule
        $\sigma_5$ denotes standard deviation over five runs. \\
    \end{tabular}

    \caption{\textbf{Distance}. Average relative error between GPT-4's predicted distance and the actual distance for 40 city pairs sampled from different population and continent city groups. 
    }
    \label{tab:distances}
\end{table}

\begin{figure}
    \centering
    \includegraphics[width=4cm]{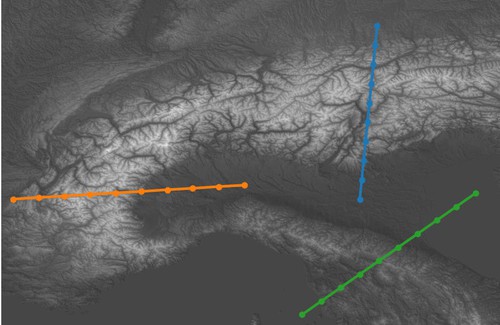}
    \includegraphics[width=9.6cm]{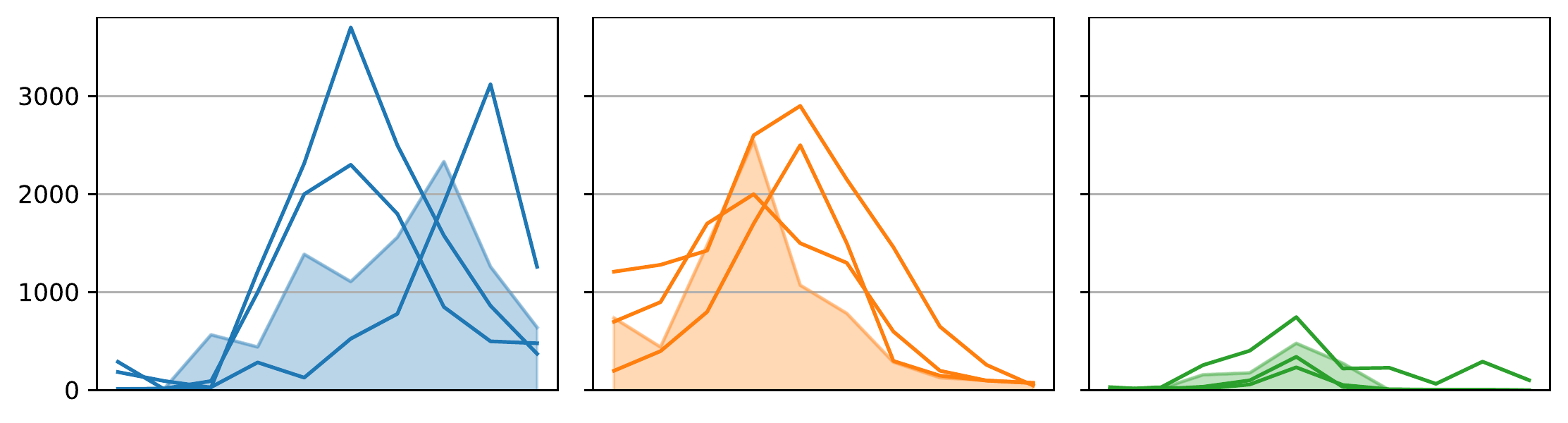}
    \caption{\textbf{Topography}. Predicted (lines) and actual elevations (shaded areas) along the trajectories depicted on the left (underlying region is in the Alps, brighter means more elevation).
   }
    \label{fig:elevation}
\end{figure}

\textbf{Topography}. We qualitatively assess GPT-4's topographical knowledge by considering three straight line trajectories (w.r.t. the geographic coordinate system) in the Alps and northern Italy. We simultaneously prompt for the elevation at ten equidistant points along each trajectory (Fig.~\ref{fig:elevation}, each prompt is repeated three times). To obtain ground truth, we use the ESA Copernicus Digital Elevation Model \cite{esa2021copernicus}. We build a geo-referenced elevation map that is sampled at the coordinates along the three trajectories. These coordinates are directly used as an input for GPT-4 using the following prompt:
\begin{formattedquote}
Provide a rough estimate of the elevation at the following coordinates to the best of your knowledge. Answer directly with a comma-separated list of elevations in meters only but without indicating the unit in the output. \\
45.00000, 11.20000\\
45.33333, 11.23333\\
...
\end{formattedquote}

The results indicate that GPT-4 has acquired a good sense of the elevation in this region of the world. However, it does not generate highly accurate predictions and tends to be sensitive to prompting (see \hyperref[app]{Appendix}).

\begin{figure}[htbp]
    
    \begin{subfigure}{0.24\textwidth}
        \centering
        \includegraphics[width=\linewidth]{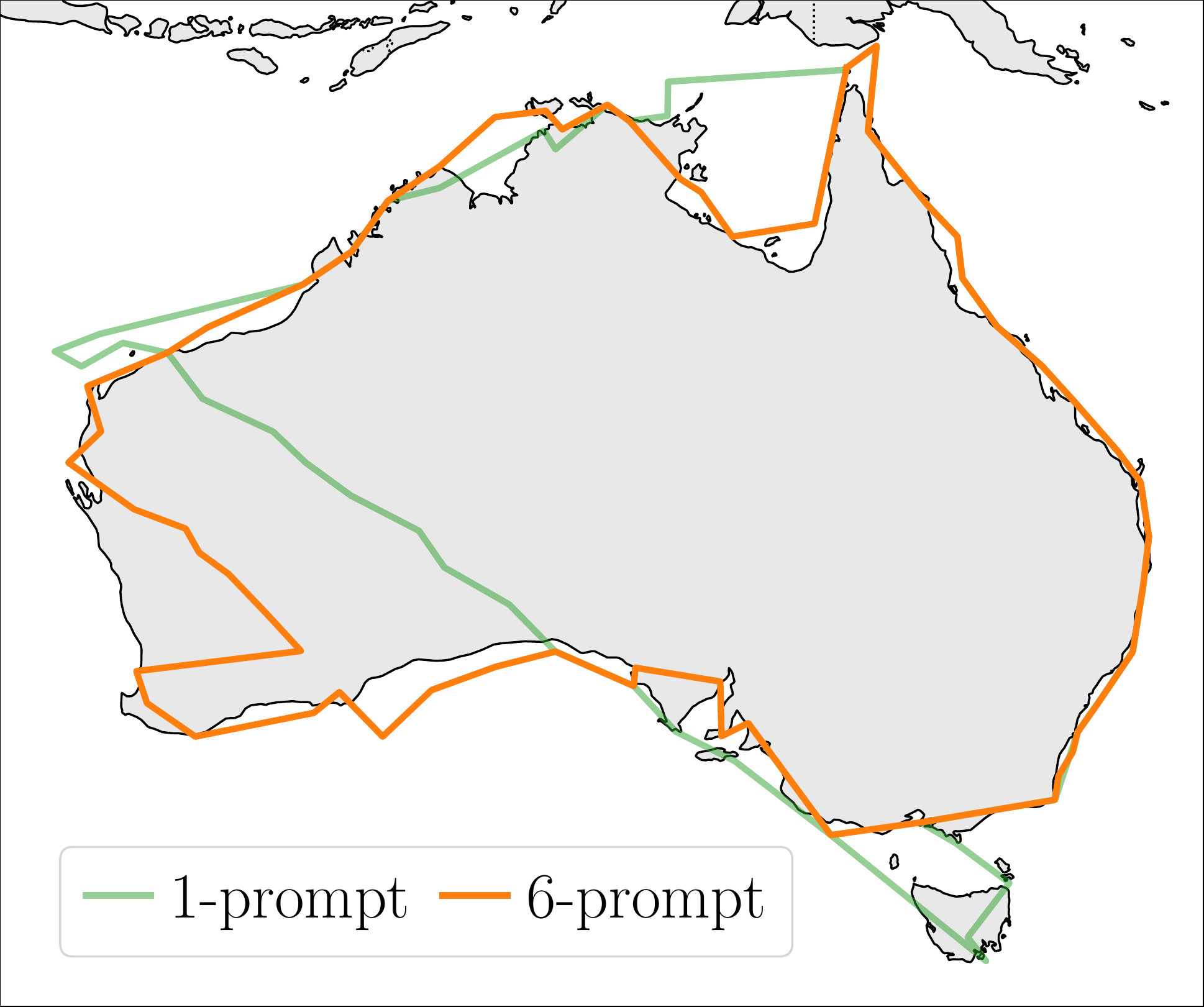}
        \caption{Australia.}
        \label{fig:o1}
    \end{subfigure}
    \begin{subfigure}{0.24\textwidth}
        \centering
        \includegraphics[width=\linewidth]{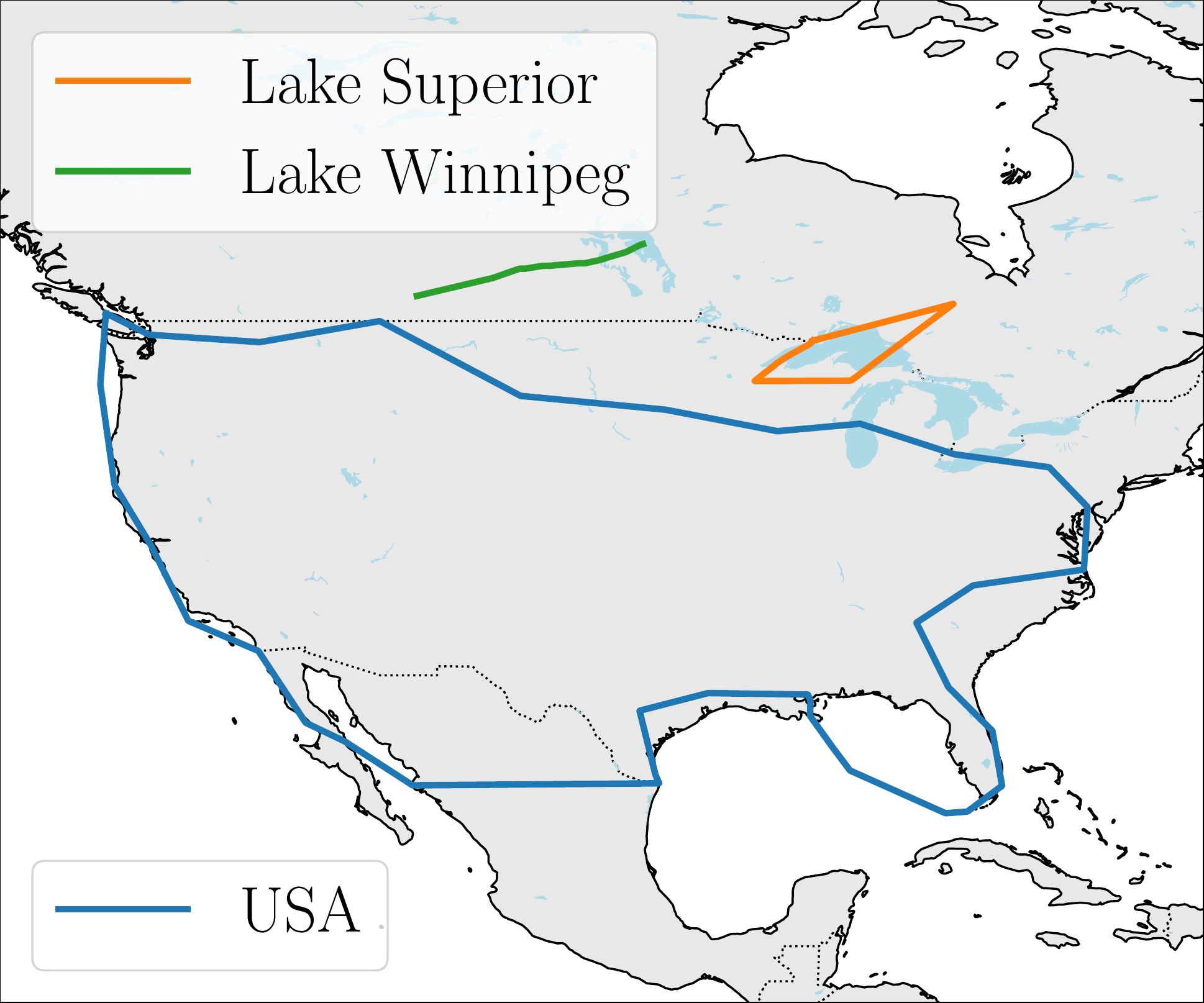}
        \caption{USA + Lakes.}
        \label{fig:o2}
    \end{subfigure}
    \begin{subfigure}{0.24\textwidth}
        \centering
        \includegraphics[width=\linewidth]{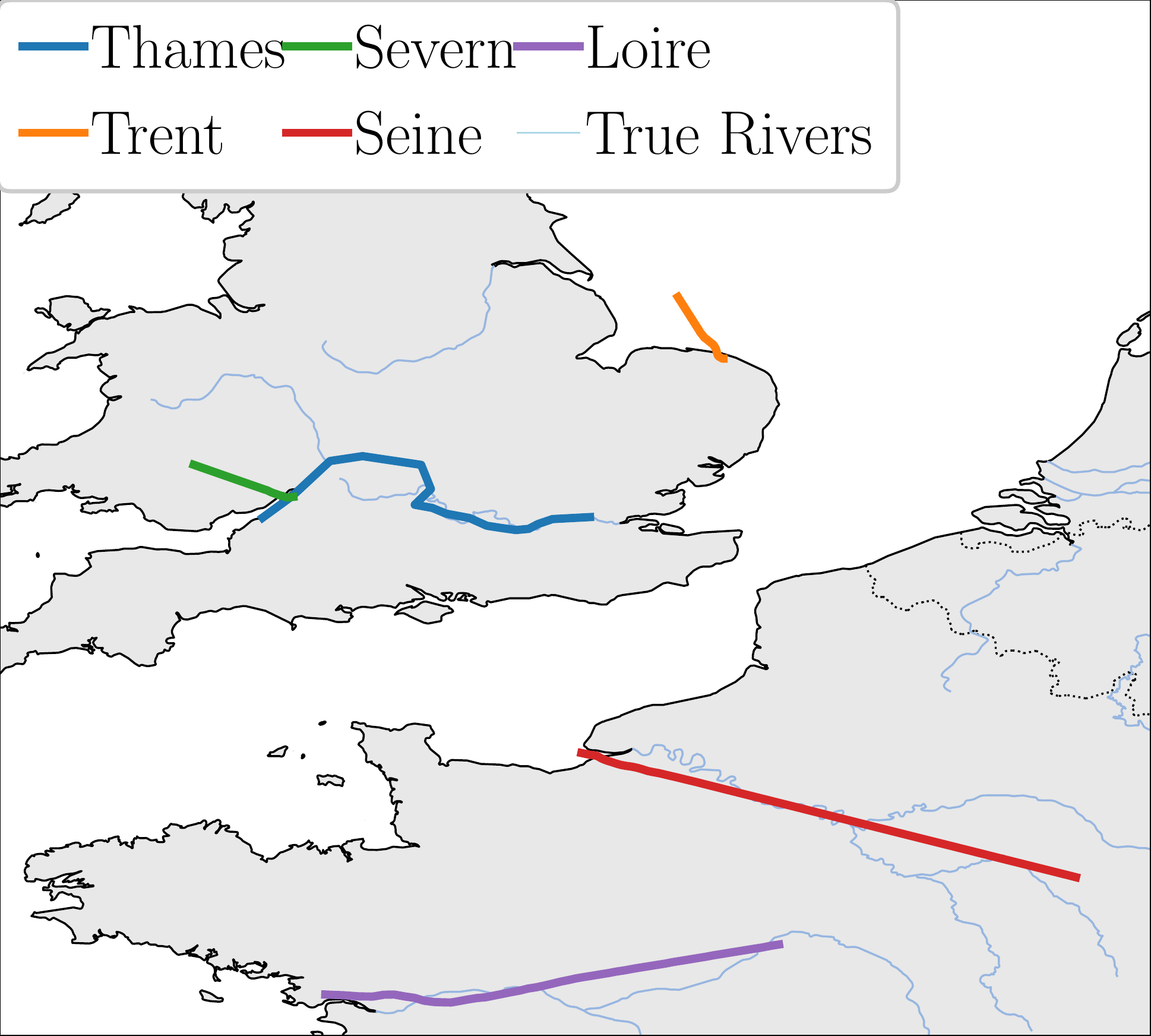}
        \caption{Rivers.}
        \label{fig:o3}
    \end{subfigure}
    \begin{subfigure}{0.24\textwidth}
        \centering
        \includegraphics[width=\linewidth]{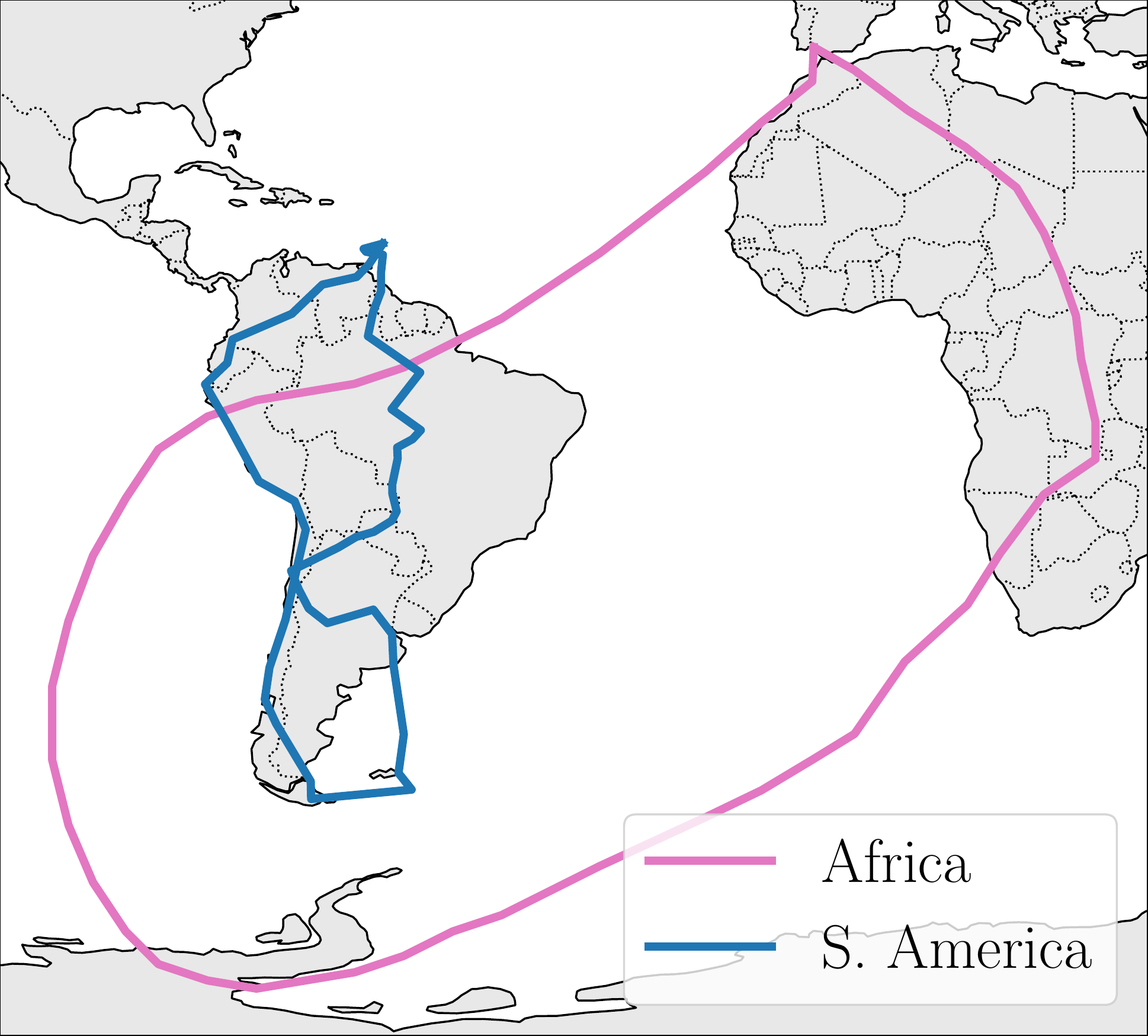}
        \caption{Continents.}
        \label{fig:o4}
    \end{subfigure}
    \caption{\textbf{Outlines} for various geographic features produced using coordinates given by GPT-4. Refinement with additional prompts improves the outlines, see (a) after 1 and 6 iterations.}
    \label{fig:mapping}
\end{figure}

\textbf{Outlines}. 
We task GPT-4 to provide coordinates for the outlines of countries, rivers, lakes and continents -- see Fig. \ref{fig:mapping}. We find a lot of inconsistency with the outputs. Generally, the outlines are geographically close to the target area and bear resemblance, but are frequently the wrong shape and the points crisscross (the outline for Africa is consistently inaccurate). Iteratively improving the response by providing feedback results in better outlines, such as for Australia. We structure the prompt in the same way when querying GPT-4 for the outlines of geographic features:

\begin{formattedquote}
    Please provide the lat/lon coordinates for the outline of \textit{<X>} as a Python list of tuples, consisting of approximately 50 points arranged clockwise. Due to output length limitations, only the coordinates should be returned.
\end{formattedquote}

\textbf{Languages.} To assess GPT4's knowledge of spoken language distribution we ask it to enumerate countries that match language-related criteria: (1) with at least three official, (2) with French and English being spoken and (3) with Romance and Germanic languages widely being spoken.
To this end, we used the following prompt template for the 1st prompt:
\begin{formattedquote}
    Name all countries that match \textit{<language criteria>}. Provide the output as a python list.
\end{formattedquote}

After obtaining a first set of countries we make use of the interactive chat setting and ask GPT-4 to extend the provided set by asking for any additional countries.

\begin{figure}[tb]
    \centering
    \includegraphics[width=14cm]{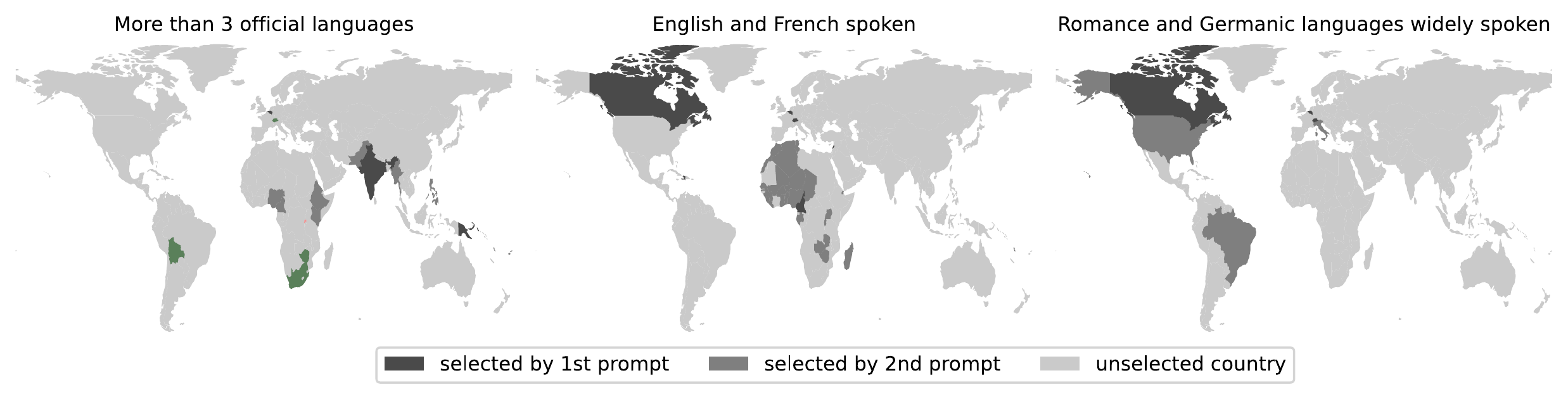}
    \caption{Distribution of countries that match different language-related criteria according to GPT-4. For the left plot, predictions that match Wikipedia list are indicated green, missing countries red.}
    \label{fig:langauges}
\end{figure}

Results (Fig.~\ref{fig:langauges}) indicate a good knowledge about the distribution of languages. For the task of identifying countries with more than three official languages GPT-4 misses only one country (Rwanda) from a Wikipedia list but suggest several additional countries. For these additional countries, the predictions still appear reasonable: For example, while India only has two official languages of the government, there are many more at the regional level.

\subsubsection*{Discussion}
We find that GPT-4 has strong capabilities for solving descriptive geographic tasks. It attains low errors in the majority of the factual knowledge retrieval experiments. The error increases as we move towards more difficult queries where the requested information is perhaps not directly seen during training and interpolation is needed. We observe no consistent variation in performance across different continents. We find GPT-4's output is significantly impacted by prompting: minor variations in output are observed using identical prompts, while there is a large variance for different prompts. Refining the answers with subsequent prompts can yield considerable improvements and serve as a technique for generating better answers.

\subsection{Application-Centric}

Having demonstrated GPT-4's underlying understanding of descriptive factual geography, we move towards more complex application-centric experiments that build upon this knowledge. 
We begin with exploring travel queries, potential downstream applications, as well as real-world and more abstract tasks that probe GPT-4's logical reasoning capabilities. 
These experiments further examine the model's ability to integrate knowledge from different sources. Lastly, we discuss our findings.

\begin{figure}
\centering
  \begin{subfigure}{0.49\textwidth}
    \centering
    \includegraphics[width=\textwidth, height=5cm]{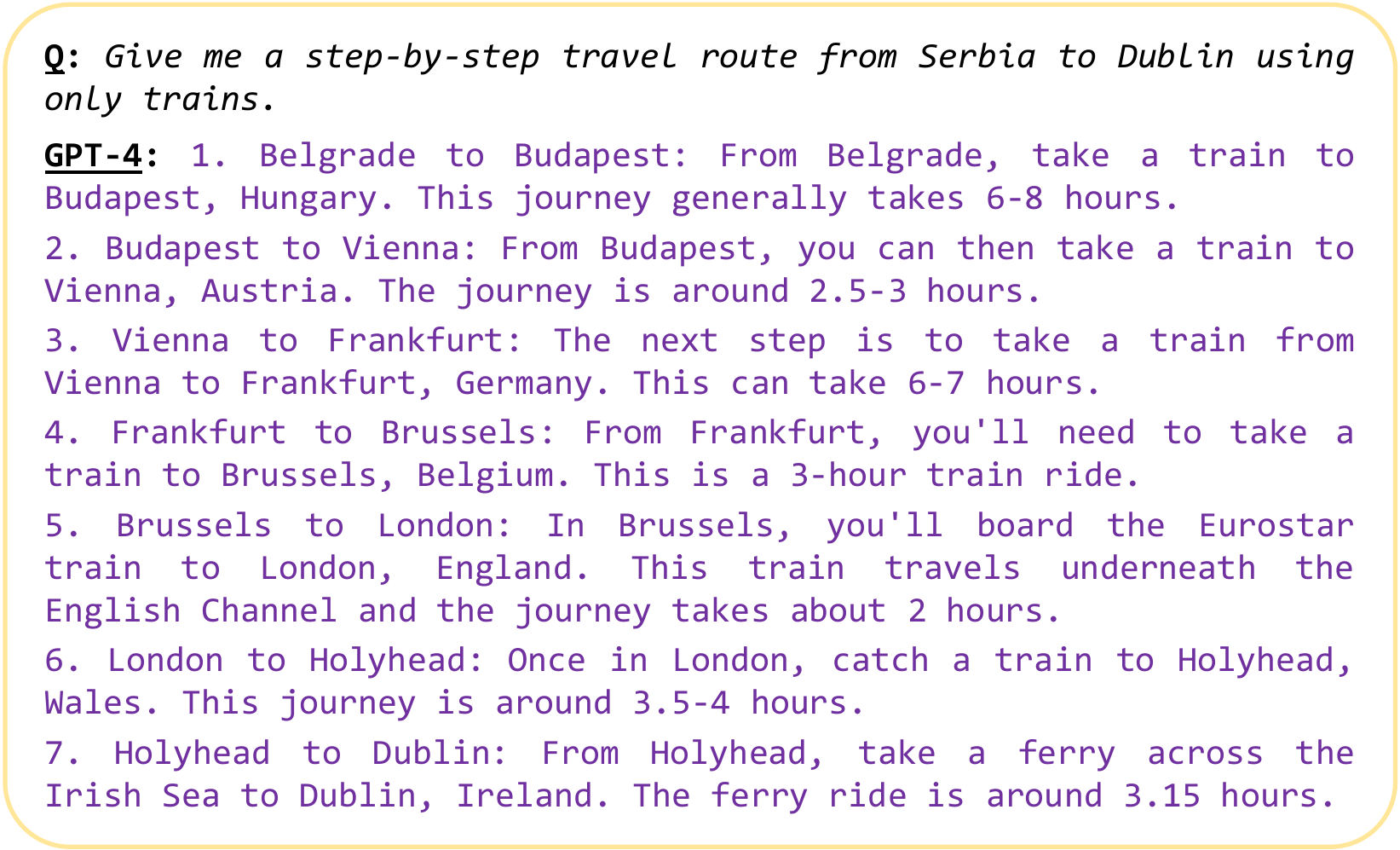}
    \caption{Travel route from \textit{Serbia} $\rightarrow$ \textit{Dublin}.}
    \label{prompt:serbia}
  \end{subfigure}
  \begin{subfigure}[b]{0.49\textwidth}
  \centering
    \includegraphics[width=\textwidth, height=5cm]{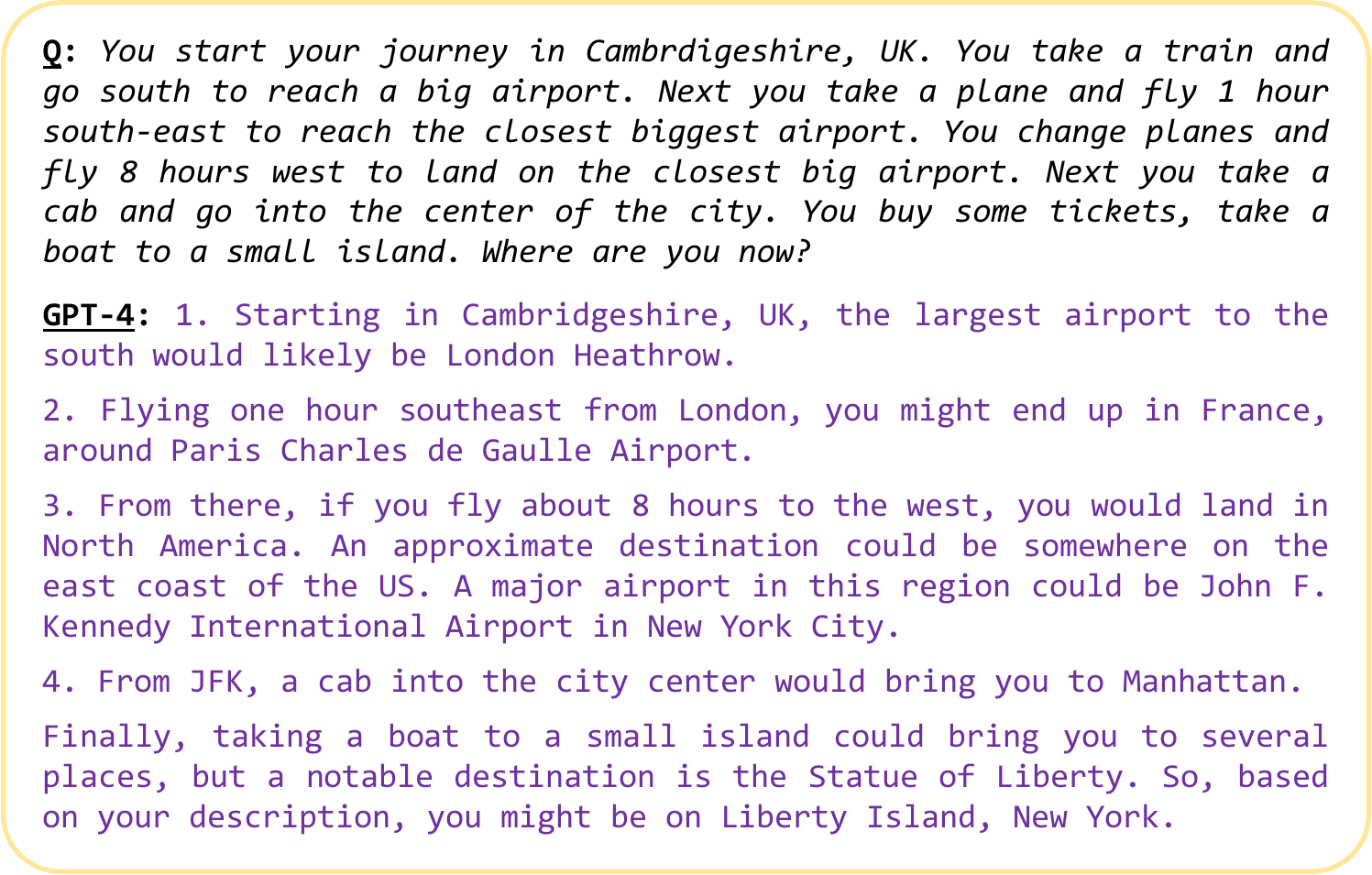}
    \caption{Navigation from \textit{Cambridge} $\rightarrow$ \textit{New York}.}
    \label{prompt:nav}
    \end{subfigure}
    \caption{\textbf{Route planning and navigation} responses from GPT-4. \vspace{-10pt}}
\end{figure}

\textbf{Route planning}. We query the model to see if it can provide plausible travel routes between specified places using prompts like:
\begin{formattedquote}
    Give me a step-by-step travel route from \textit{<Start\_Location>} to \textit{<End\_Location>} [(optional) using only \textit{<Mode\_of\_Transport>}]. 
\end{formattedquote}
Where the locations can be countries, cities, landmarks, names of streets or buildings, or lat/lon coordinates, and we specify modes of transport such as trains, buses, cars, and airplanes. We verify the accuracy of predictions using ground truth from Google Maps\footnote{https://www.google.com/maps}. The example in Fig. \ref{prompt:serbia} shows that GPT-4 can piece together information from multiple sources to provide a realistic travel plan from \textit{Serbia} to \textit{Dublin} using only trains. Similarly, when prompted to provide a travel route from \textit{Dallas, Texas} to \textit{The Swiss Alps}, it suggested multiple options including intermediate layovers, taking a combination of airlines, trains, and rental vehicles, as well as recommendations for alpine destinations.
We found GPT-4 to be capable of planning routes between short distances if it is allowed flexibility in its response. However, when constrained to urban routing with buses, we found some inaccuracies. Asking for bus routes within London yielded responses that matched %
the official Transport for London bus route, but exact bus numbers and stops were inaccurate. This might be because real-time bus routes change more frequently compared to airports, train stations, or road networks. Additional experiments and visualisations of the responses are provided in the \hyperref[app]{Appendix}.

\begin{figure}[h!]
    \centering
    \includegraphics[width=\textwidth, height=4.7cm]{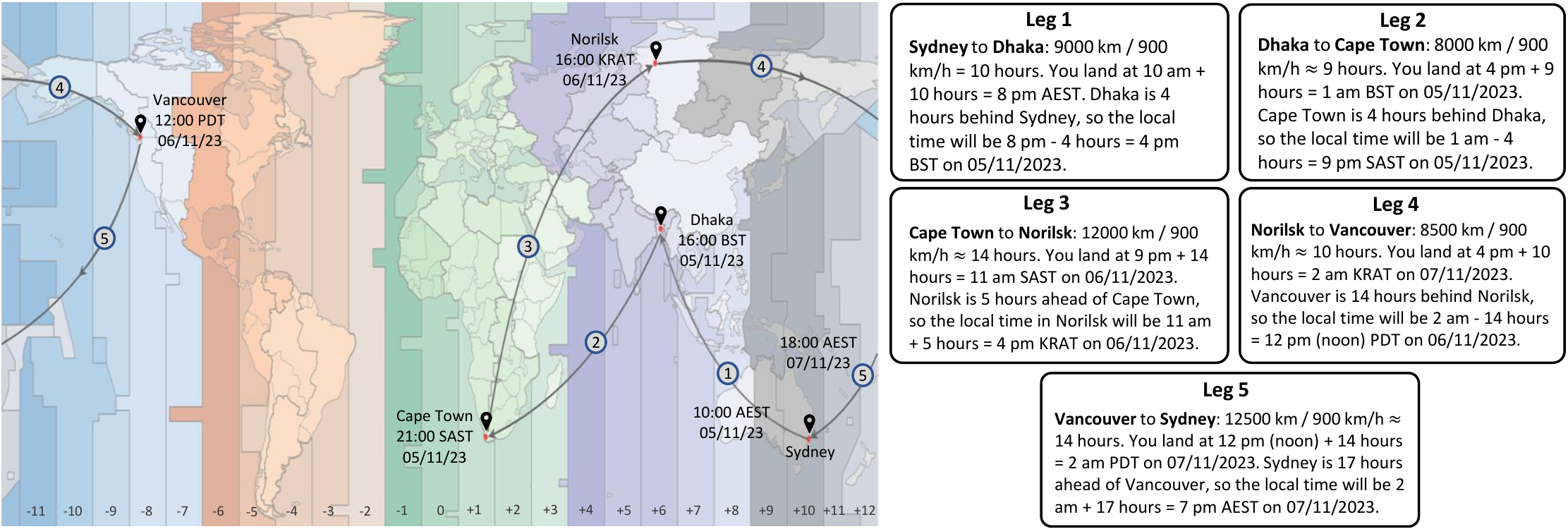}
    \caption{\textbf{Navigation}. GPT-4 successfully navigates a journey with multiple time zones.}
    \label{fig:timezone}
\end{figure}

\textbf{Navigation.} Directional navigation is more challenging than generic route planning because we constrict the intermediate objectives, thereby reducing the %
flexibility and encouraging GPT-4 to logically correlate the given data with its existing geographic knowledge rather than just recall established routes from memory.  We use the following style of prompts for these experiments:
\begin{formattedquote}
    You start your journey in \textit{<Location A>}. You take a \textit{<vehicle>} and go \textit{<direction>} for \textit{<duration>}. Where are you now?
\end{formattedquote}
Here, directions can be \textit{north, south, south-east}, etc. and duration can be \textit{minutes, hours,} or \textit{days}. Alternatively, we also use distance-based directions such as \textit{``I drive 20 km south"}. The prompt in Fig. \ref{prompt:nav} describes a journey from \textit{Cambridgeshire} $\rightarrow$ \textit{London} $\rightarrow$ \textit{Paris} $\rightarrow$ \textit{Manhattan} $\rightarrow$ \textit{Liberty Island}. Even without mentioning any specific information about these checkpoints, GPT-4 was able to follow the correct trajectory and reach the expected destination. In additional experiments (see \hyperref[app]{Appendix}) we extended the prompt to more complex journeys consisting of $15$ stops (e.g., from \textit{Sapporo, Japan} to \textit{Helsinki, Finland}). We found that unspecified waypoints resulted in different journeys for the same prompt as it is possible to reach many closely connected places given only the direction and time. However, refining the prompt with small amounts of information such as ``\textit{We land at a country that is famous for sausages}" allows GPT-4 to identify that we arrive in e.g., Germany. Moreover, the length of the route or the number of intermediary stages seems to have little impact on the accuracy of the response since GPT-4 breaks down and deciphers each leg of the journey independently. We found that GPT-4 can accurately track \textbf{short distances} within cities when given exact measurements. Additionally, we verified that for \textbf{long distance} routes between countries, GPT-4 can keep track of changing time zones and estimate travel/arrival times for complex journeys - see Fig. \ref{fig:timezone}; however, it struggles to account for daylight savings if not specified precisely.

\begin{figure}[h]
  \centering
\begin{framed}
  \begin{subfigure}{0.48\textwidth}
    \raggedright
    \includegraphics[width=0.9\textwidth]{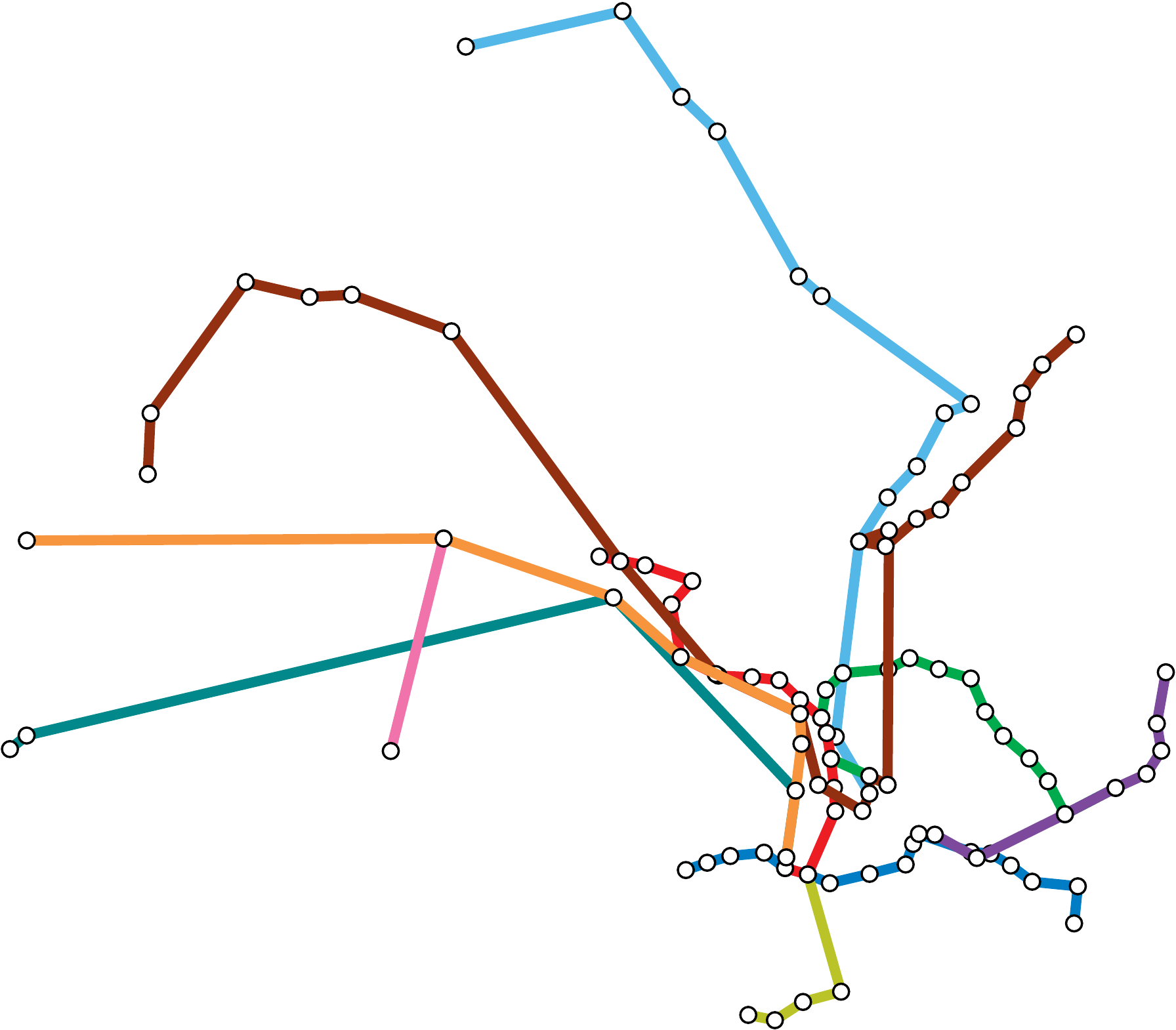}
    \caption{GPT-4.}
    \label{fig:hk_pred}
  \end{subfigure}
  \quad
  \begin{subfigure}{0.48\textwidth}
    \raggedleft
    \includegraphics[width=0.9\textwidth]{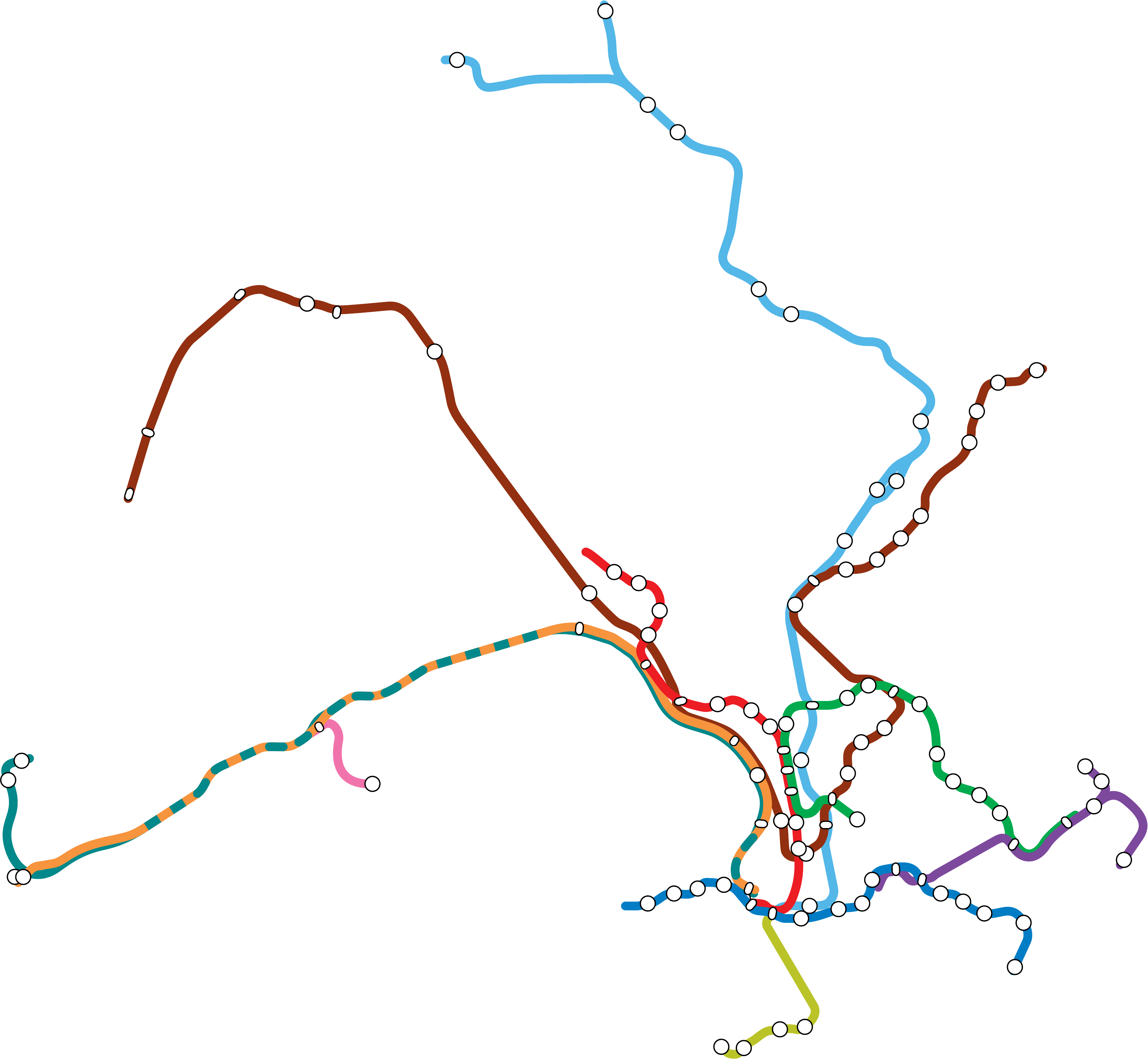}
    \captionsetup{position=below}
    \caption{Ground truth (2022)\footnotemark.}
    \label{fig:hk_true}
  \end{subfigure}
  \end{framed}
  \caption{\textbf{Hong Kong Mass Transit Railway (MTR) Network Map}.
  }
  \label{fig:both_figures}
\end{figure}

\textbf{Networks}. Having investigated GPT-4's route planning ability, we explore if it can recreate an entire travel network, e.g., the Hong Kong MTR network. Initially, we ask for a list of all the lines in the network. For each line, we ask for the lat/lon coordinates of the stations, in the correct order. Fig. \ref{fig:hk_pred} shows the resulting map that closely matches the ground truth (Fig. \ref{fig:hk_true}): all the stations are included (aside from those added after 2021) in the correct order (minus forks in the line), though there are inaccuracies in the positioning, especially at interchange stations. Here is the specific two-part prompt we use for each line of the Hong Kong MTR:

\begin{formattedquote}
    Provide a list of the names of the stations in order on the Hong Kong MTR
    \textit{<Line Name>} Line.
    \newline
    \newline
    Give the latitude and longitude coordinates for each of these as a python list of
    tuples. Maintain the same order.
\end{formattedquote}
\FrameSep2pt

\footnotetext{Adapted from https://commons.wikimedia.org/wiki/File:Hong\_Kong\_Railway\_Route\_Map\_en.svg}

\begin{figure}
    \centering
    \includegraphics[width=\textwidth]{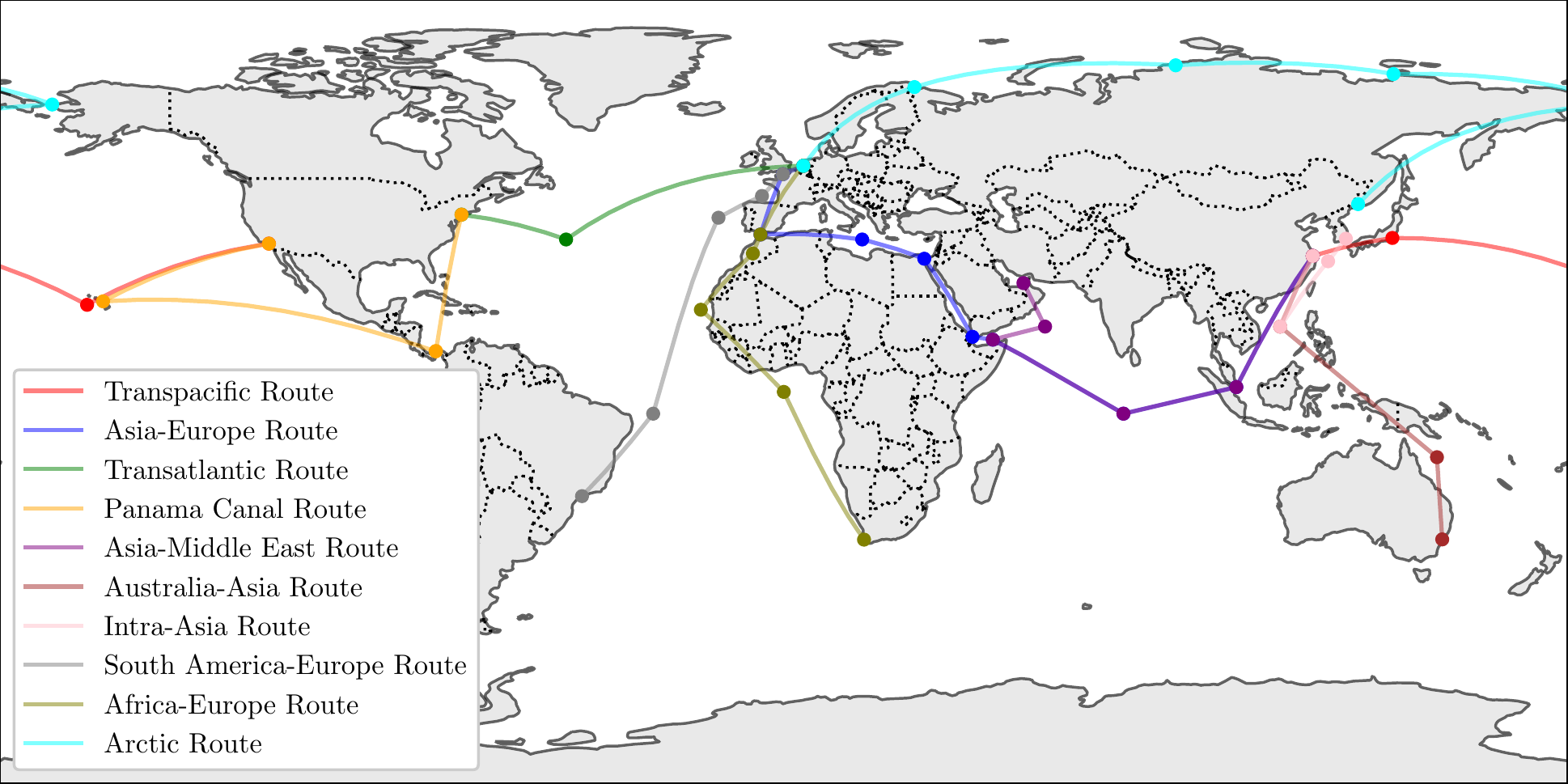}
    \caption{\textbf{Major international maritime shipping routes}.}
    \label{fig:maritime}
\end{figure}

We also generate the maritime shipping routes visualised in Fig. \ref{fig:maritime}. This shows the model's capabilities of generating coordinates that are not related to specific landmarks, as well as places in the middle of the ocean. We used the following prompt.
\begin{formattedquote}
    I want to plot the primary maritime shipping routes of the world. Please provide the lat/lon coordinates of each route. Indicate the start and finish and provide at least two or more coordinates for intermediate steps. For multiple routes provide separate lists of coordinates. Make sure that the paths do not intersect with any landmasses. Give the values as a list of python tuples and dictionaries
\end{formattedquote}
Recreations of flight and other rail networks can be found in the \hyperref[app]{Appendix}.

\textbf{Itinerary planning}. GPT-4 can act as a travel assistant by providing customised itinerary suggestions based on the provided requirements. We use the following prompt,
\begin{formattedquote}
    I am currently in \textit{<Source\_Location>}. I want to visit \textit{<Destinations>}, and I have a budget of \textit{<Budget>}. Provide a day-by-day step-by-step detailed itinerary plan for the whole trip with a breakdown of specific places to visit, foods to try out, as well as the required time, and money I need. Provide a breakdown of how to travel to the destinations and come back home.
\end{formattedquote}

\begin{figure}[h]
    \centering
    \includegraphics[width=\textwidth]{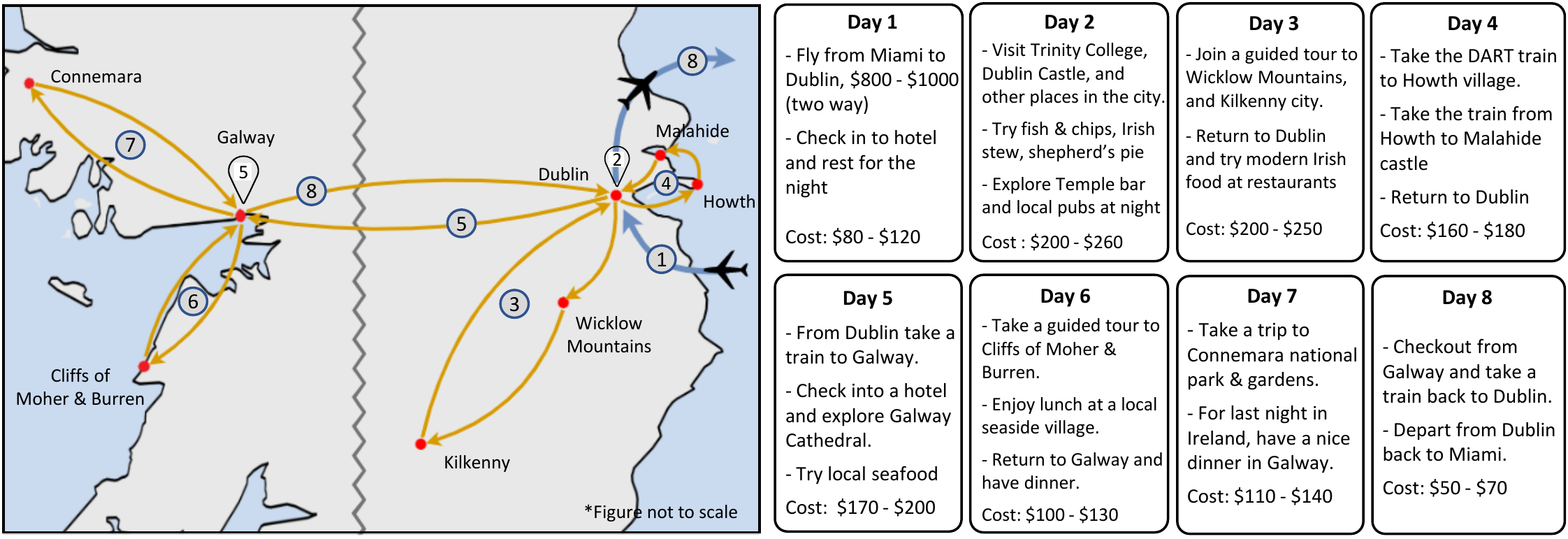}
    \caption{\textbf{Travel itinerary} suggestion for a 8-day trip in \textit{Ireland} starting from \textit{Miami}.}
    \label{fig:itinerary}
\end{figure}

Fig. \ref{fig:itinerary} visualizes the model's response of an 8-day itinerary for a holiday trip in \textit{Ireland}, for a fixed budget of \$2000. We also found that the model can accommodate to various constraints such as food allergies, requirements for children, and size of travel groups.

\begin{figure}[h!]%
    
    \begin{minipage}[b]{0.69\textwidth}
    \raggedright
        \begin{subfigure}[b]{0.29\textwidth}
        \raggedright
            \includegraphics[height=5cm]{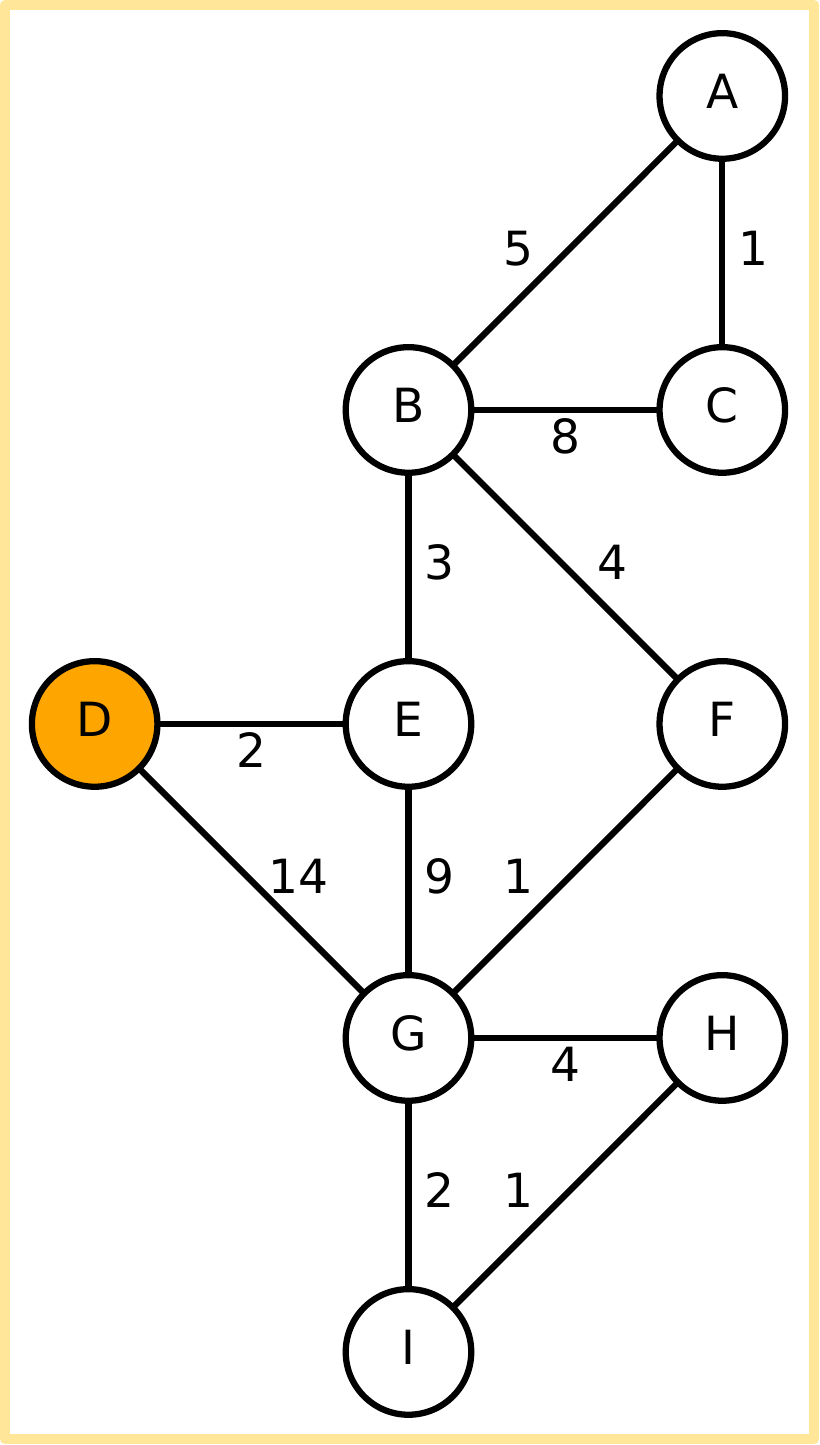}
            \caption{Graph.}
            \label{fig:graph}
        \end{subfigure}
        \begin{subfigure}[b]{0.31\textwidth}
        \raggedright
            \includegraphics[height=5cm]{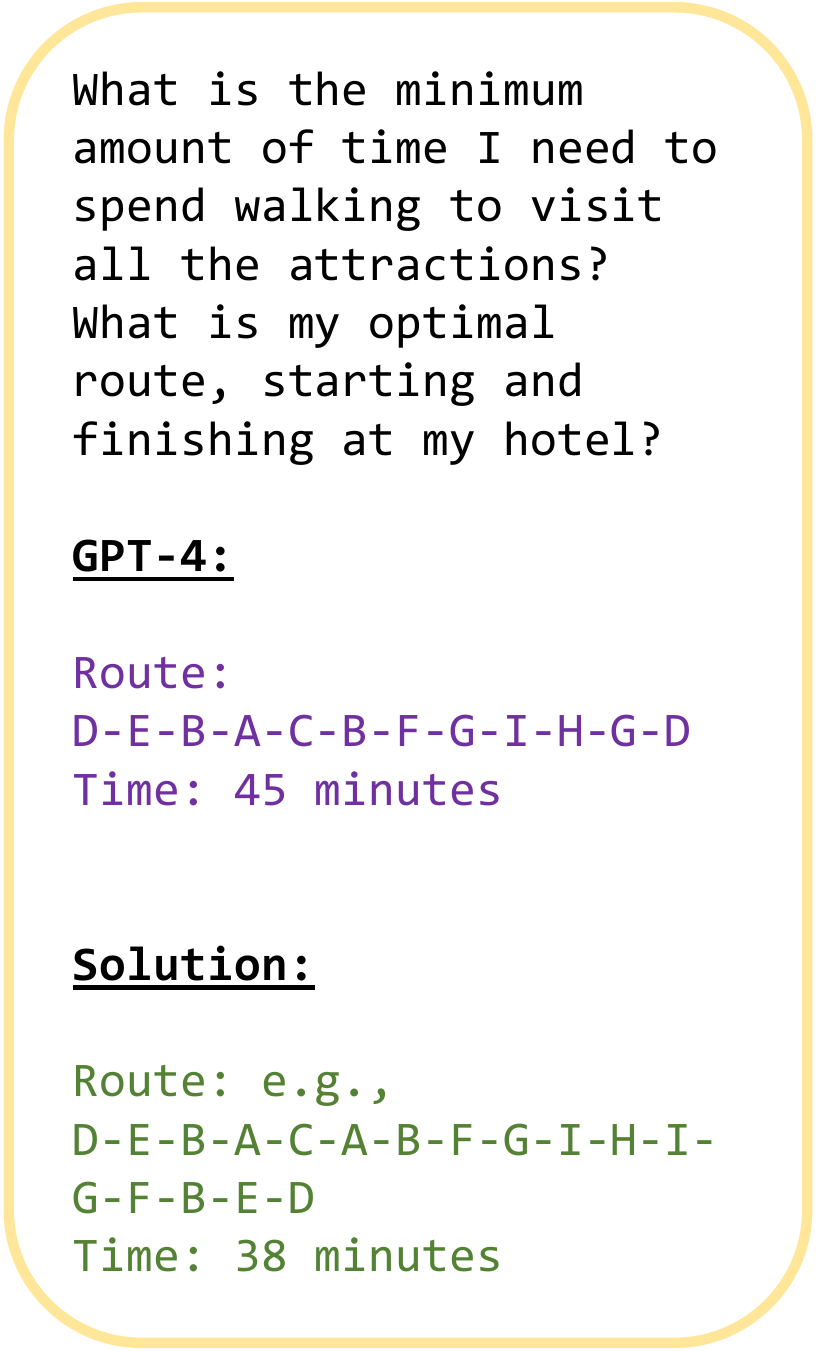}
            \caption{Every node.}
            \label{fig:node}
        \end{subfigure}
        \begin{subfigure}[b]{0.36\textwidth}
        \raggedright
            \includegraphics[height=5cm]{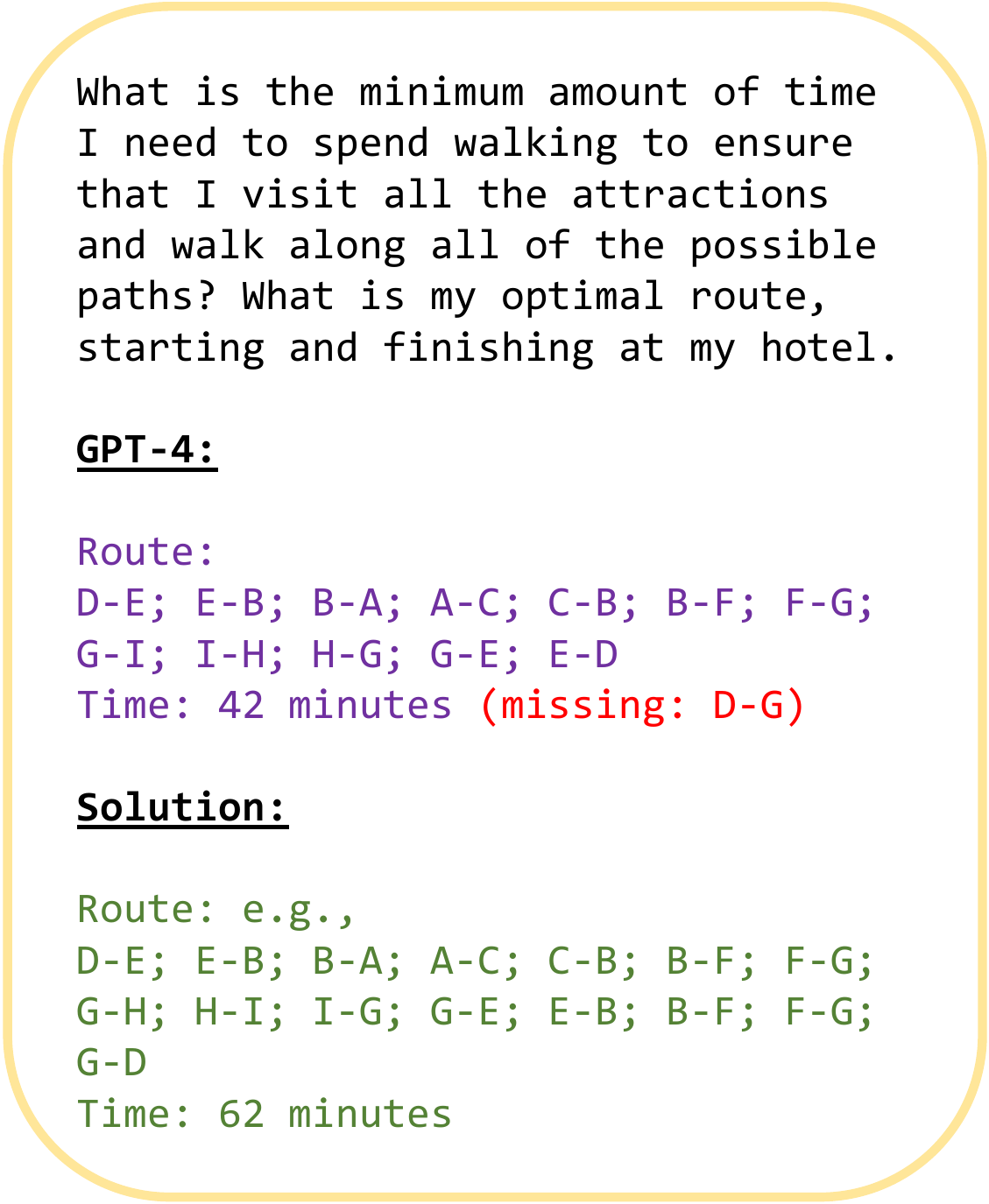}
            \caption{Every edge.}
            \label{fig:edge}
        \end{subfigure}
        \caption{\textbf{Abstract routing} based on the graph in (a).}
    \end{minipage}
    \begin{minipage}[b]{0.29\textwidth}
    \raggedleft
    \begin{subfigure}{\textwidth}
        \raggedleft
        \includegraphics[height=5cm]{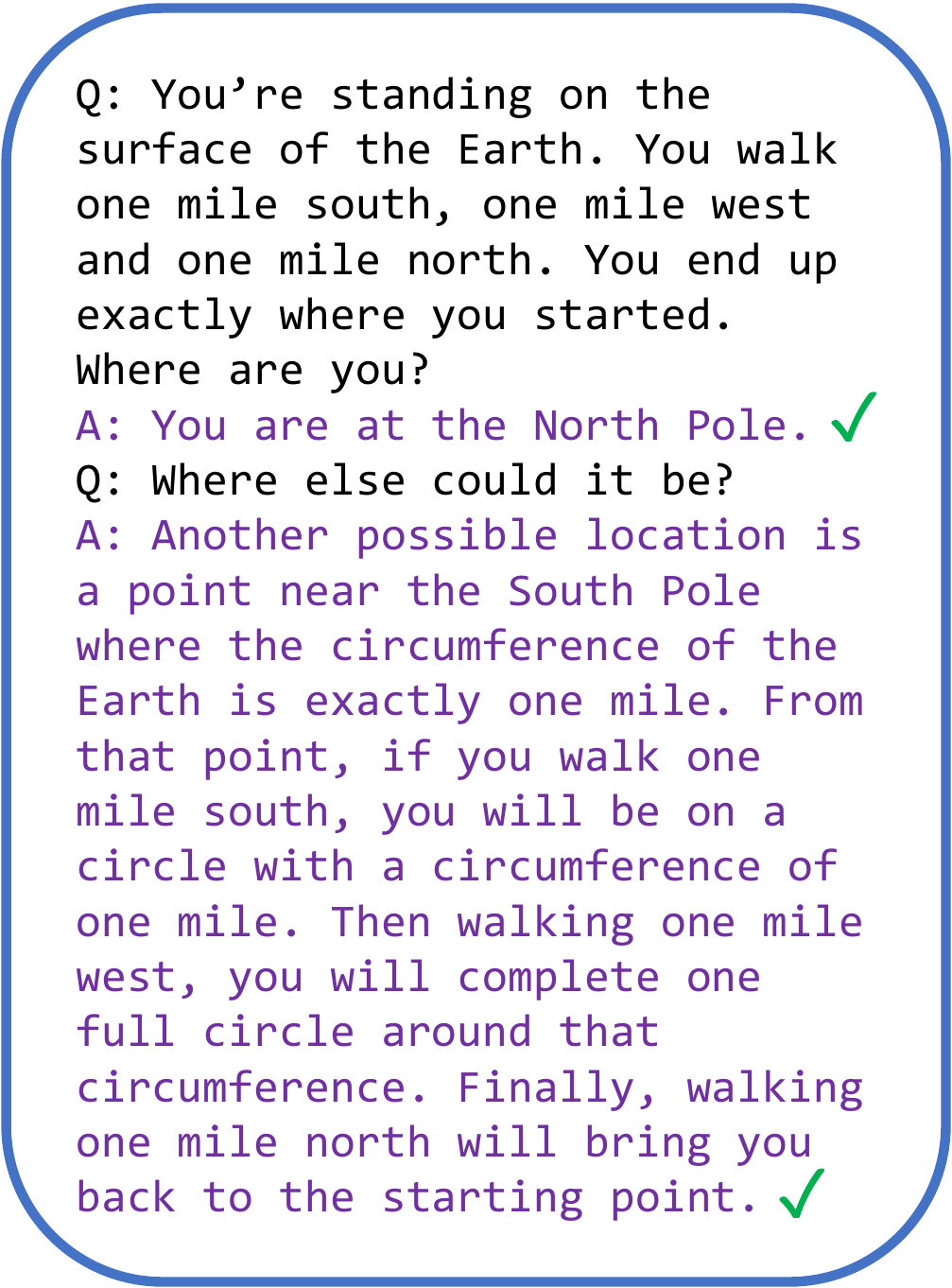}
        \caption*{}
    \end{subfigure}
        \caption{\textbf{Riddle}.}
    \label{fig:riddle}
    \end{minipage}
\end{figure}

\textbf{Abstract routing}. We probe the underlying route planning logic of GPT-4 by evaluating its performance in an abstract setting. We construct a distributed set of nodes (e.g., attractions) connected by edges (e.g., paths) with associated weights (e.g., times); one of the nodes is specified as the start and end point (e.g., hotel) -- see Fig. \ref{fig:graph}. 
We ask GPT-4 to find an optimal route that visits every node\footnote{A variation of the travelling salesman problem.} (Fig. \ref{fig:node}) and every edge\footnote{i.e., Chinese postman problem.} (Fig. \ref{fig:edge}), and returns to the start. For both, GPT-4 fails to find an optimal route, proposing solutions that needlessly take expensive paths or that miss required paths. Fig. \ref{fig:riddle} shows a 2-part geographic riddle that GPT-4 answers correctly, even if the question is rephrased.

\begin{figure}
  \centering
  \begin{subfigure}[b]{0.23\textwidth}
  \centering
    \includegraphics[height=2.5cm]{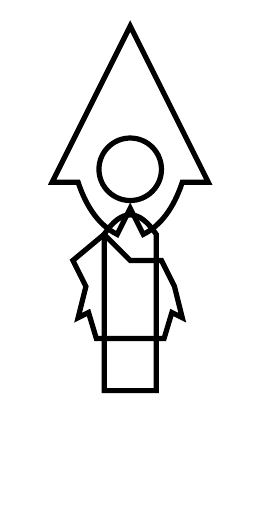}
    \caption{Statue of Liberty.}
    \label{fig:landmark1}
  \end{subfigure}
  \begin{subfigure}[b]{0.23\textwidth}
  \centering
    \includegraphics[width=3cm]{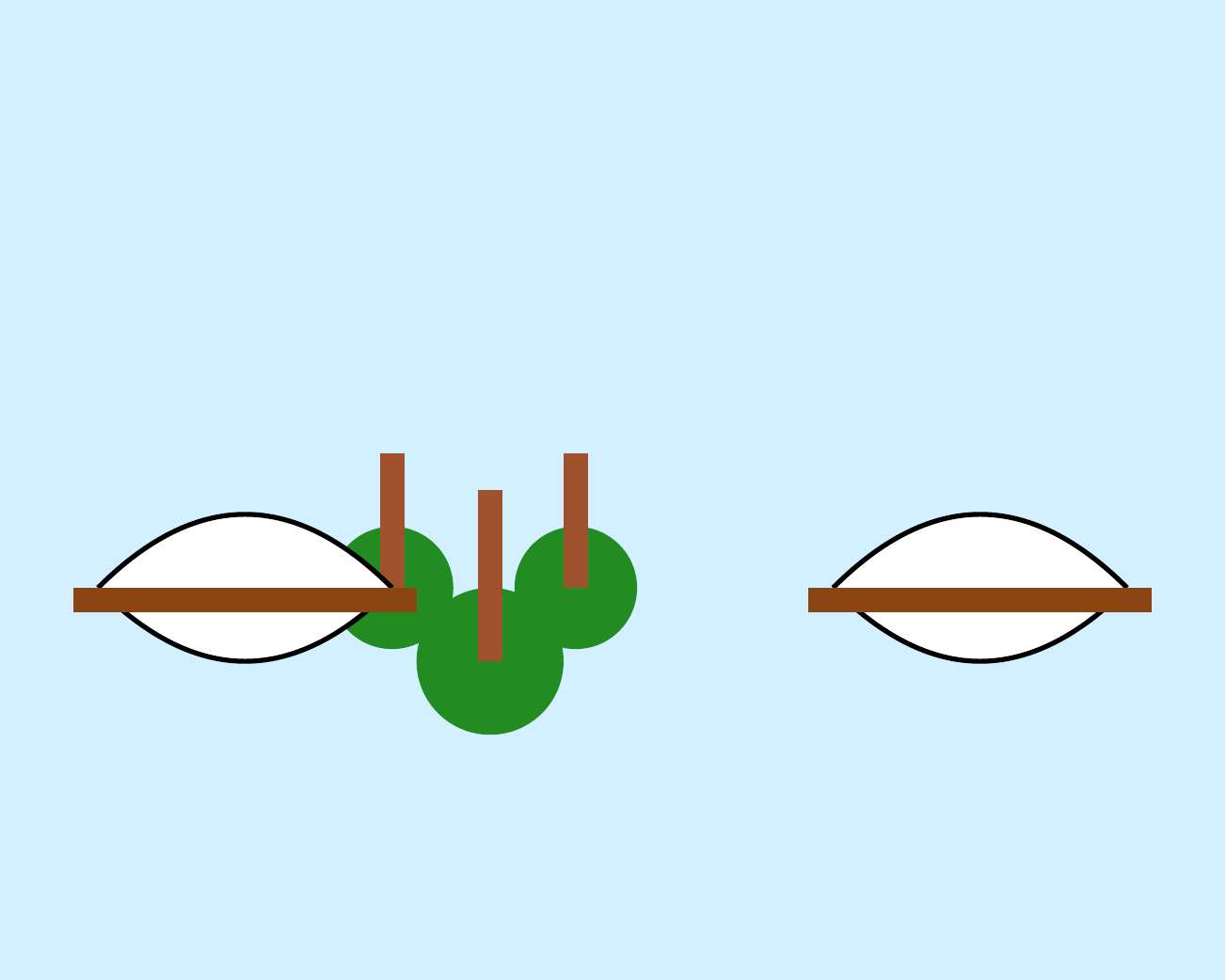}
    \caption{Gardens by the Bay.}
    \label{fig:landmark2}
    \end{subfigure}
    \begin{subfigure}[b]{0.23\textwidth}
    \centering
    \includegraphics[width=3cm]{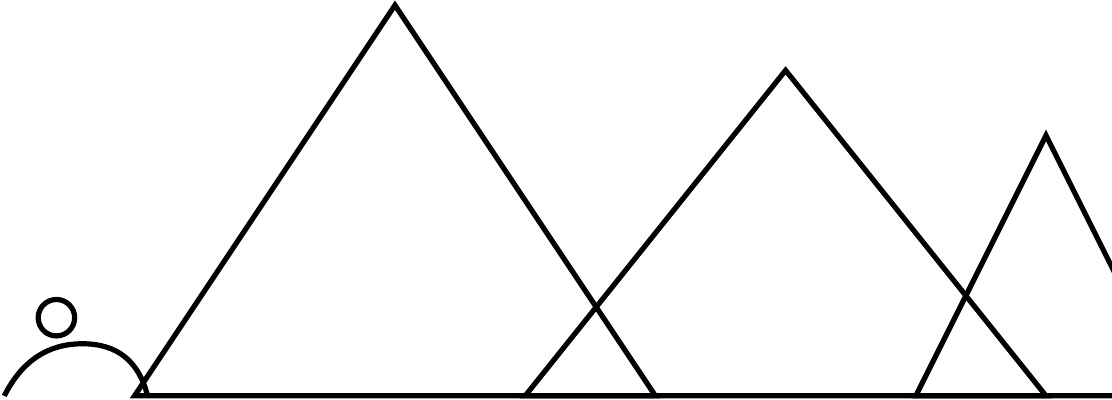}
    \caption{Pyramids Complex.}
    \label{fig:landmark3}
  \end{subfigure}
\begin{subfigure}[b]{0.23\textwidth}
  \centering
    \includegraphics[width=3cm]{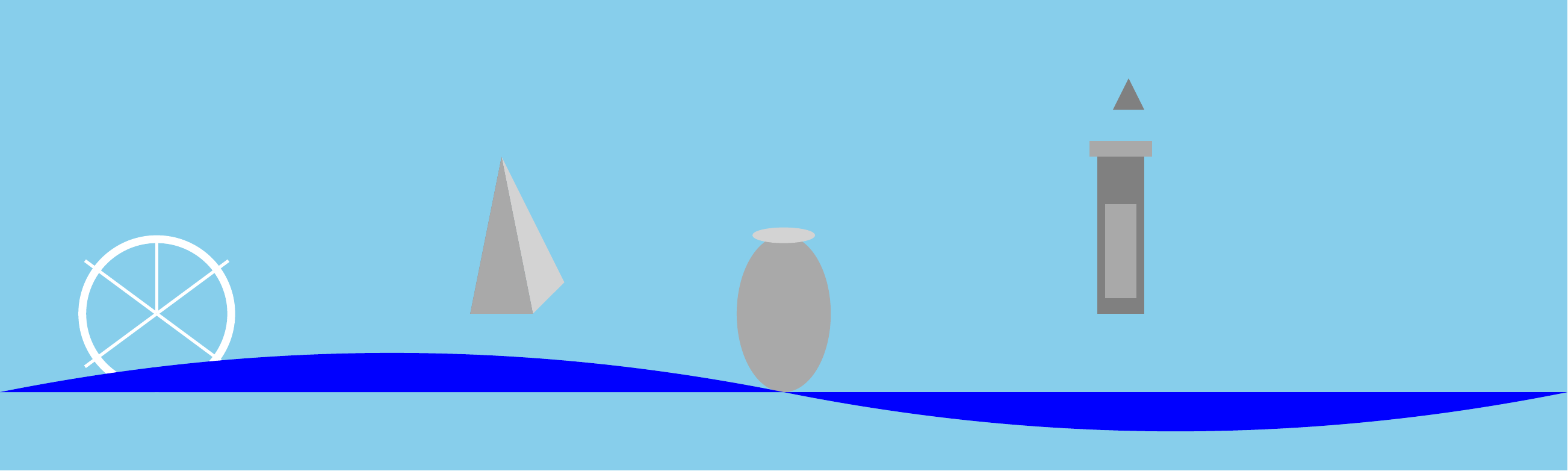}
    \caption{London Skyline.}
    \label{fig:landmark4}
  \end{subfigure}
  \caption{\textbf{Landmark SVGs}. (Countries (a)-(d): USA, Singapore, Egypt, UK).}
  \label{fig:landmarks}
\end{figure}

\textbf{Landmarks}. We find that GPT-4 not only has a remarkable ability to accurately guess landmarks from coordinates but can also produce code for SVG outlines of landmarks. Generated from a single prompt, these illustrations are clearly recognisable (though slightly abstract) -- see Fig. \ref{fig:landmarks}. In cases where landmarks are relatively isolated, we use the following prompt:
\begin{formattedquote}
    What famous landmark is at this location: \textit{<Lat/Lon\_Coordinates>}? Provide the SVG code for a detailed outline of the landmark.
\end{formattedquote}

\begin{figure}
    \centering
    \includegraphics[width=1\textwidth]{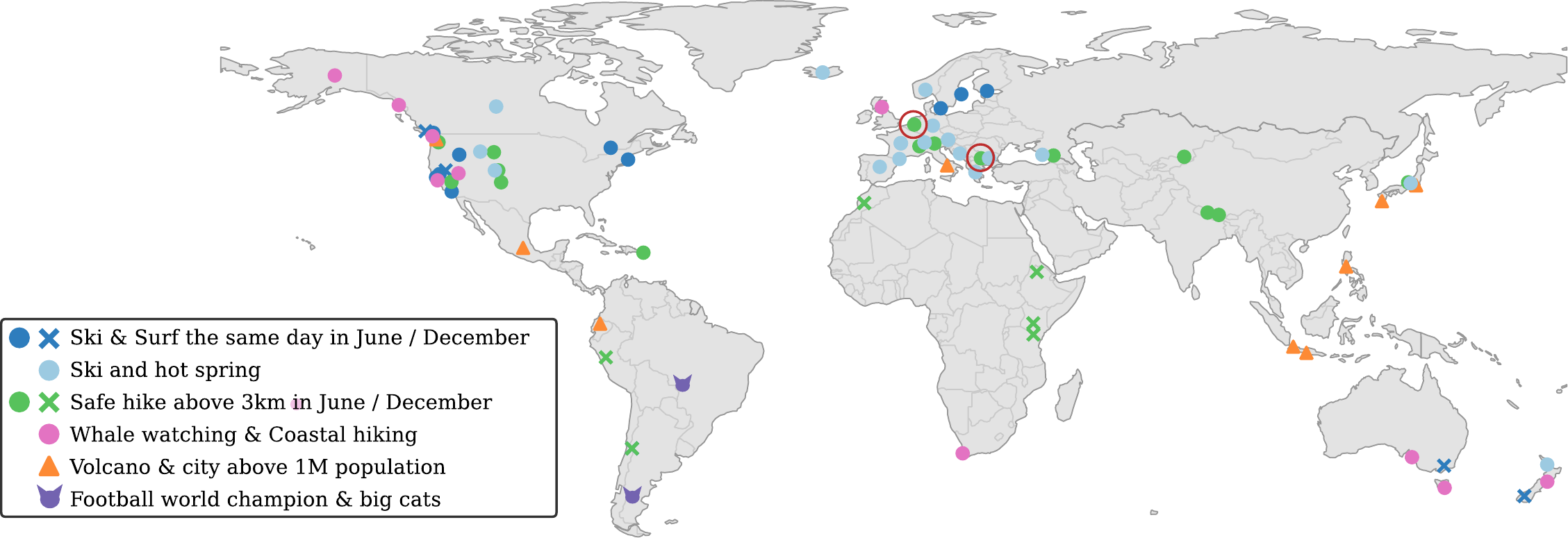}
    \caption{\textbf{Multi-criteria place retrieval}. We ask GPT-4 which places %
    satisfy different criteria.}
    \label{fig:activities}
\end{figure}

\textbf{Multi-criteria place retrieval}. 
We task GPT-4 with six prompts where the goal is to generate coordinates that match specific criteria (Fig.~\ref{fig:activities}) to assess its capability to connect different geographic information sources.
We used the following prompt for our experiment on multi-criteria place retrieval (for exact prompts see \hyperref[app]{Appendix}):
\begin{formattedquote}
    Name all places in the world where \textit{<X>}. Provide a python list in the format [0.00000N, 0.00000E].
\end{formattedquote}

The predictions are mostly correct, with some errors in details, e.g., the red circles in Fig.~\ref{fig:activities} denote places where a mountain height of over 3 km is absent. 
Furthermore, there are potential places matching the criteria that are missed, e.g., Mount Teide on Tenerife for hiking in December. 
Generally, the results indicate good skills in connecting different sources of knowledge and making plausible predictions based on somewhat vague, multi-criteria prompts.

\begin{figure}
    \centering
    \includegraphics[width=\linewidth]{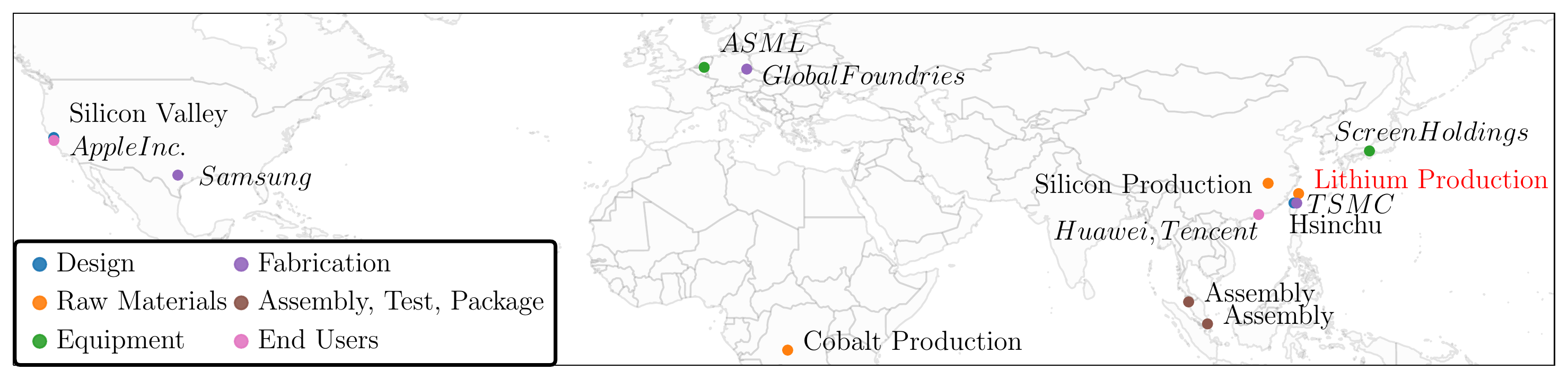}
    \vspace{-0.5cm}
    \caption{\textbf{Global semiconductor supply chain}. Companies in \textit{italics}, errors in \textcolor{red}{red}.}
    \label{fig:supplychain}
\end{figure}

\textbf{Supply chains}. To probe GPT-4's ability to reason across multiple information sources we ask for key stages and elements in the global semiconductor supply chain, along with locations. From a single prompt-answer, we construct the map shown in Fig. \ref{fig:supplychain}. The map includes the majority of the critical components at the correct locations of the supply chain \cite{miller2022chip}, with the exception of lithium production, which is labelled as Australia (a major producer) though given coordinates near China. We create the map using just a single prompt:

\begin{formattedquote}
    I want to construct a map of the semiconductor supply chain, end-to-end. Please
    provide the lat/lon coordinates and names of the key elements in
    the supply chain, including design, manufacturing, materials, equipment + tools,
    etc. If you don’t know any coordinates exactly just estimate, every point needs
    coordinates.
\end{formattedquote}

\begin{figure}[b]

\centering
  \begin{subfigure}[b]{0.38\textwidth}
    \raggedright
    \includegraphics[height=3.86cm]{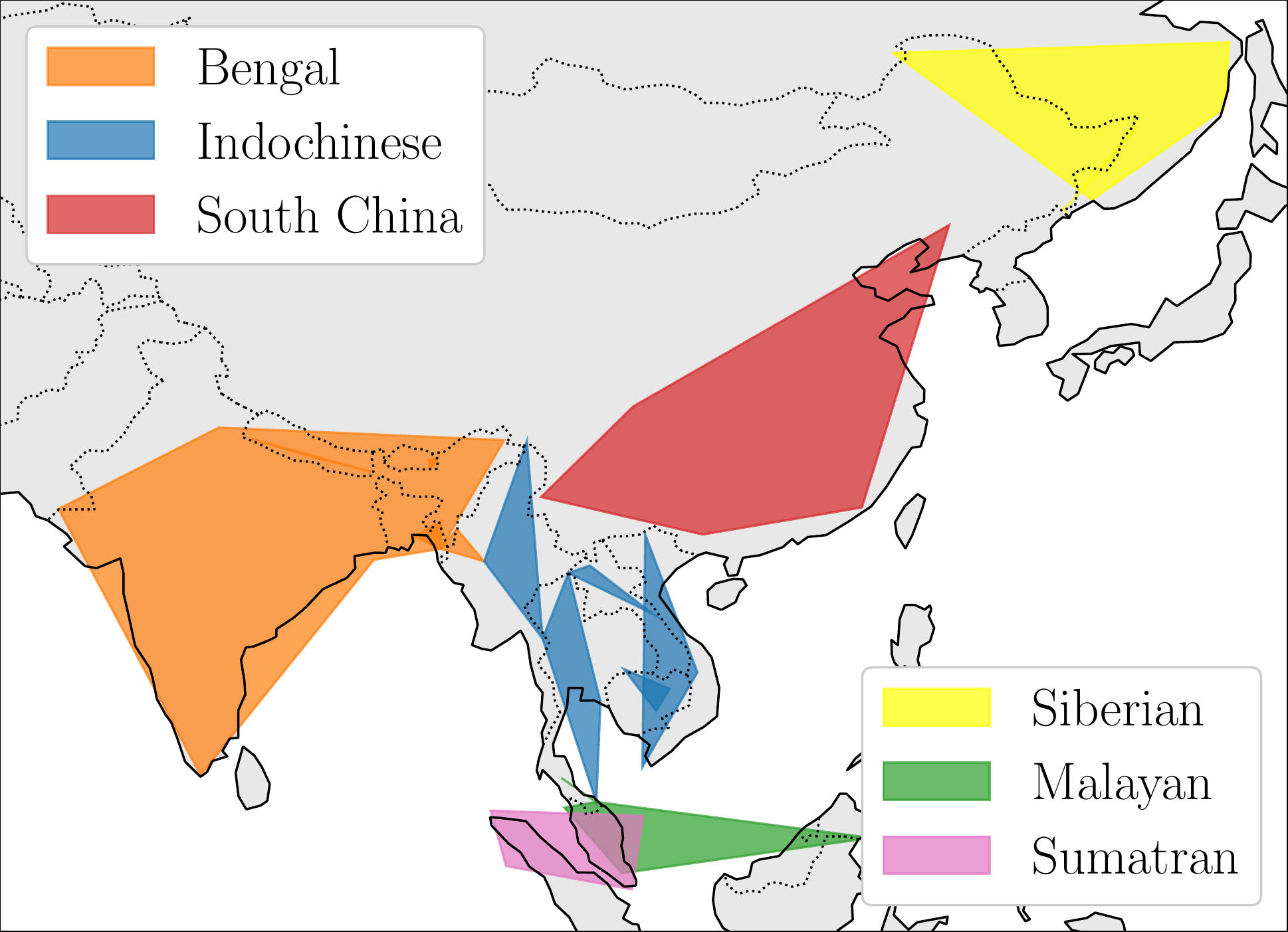}
    \caption{Tiger subspecies range.}
    \label{fig:t1}
  \end{subfigure}
  \begin{subfigure}[b]{0.6\textwidth}
    \raggedleft
    \includegraphics[height=3.86cm]{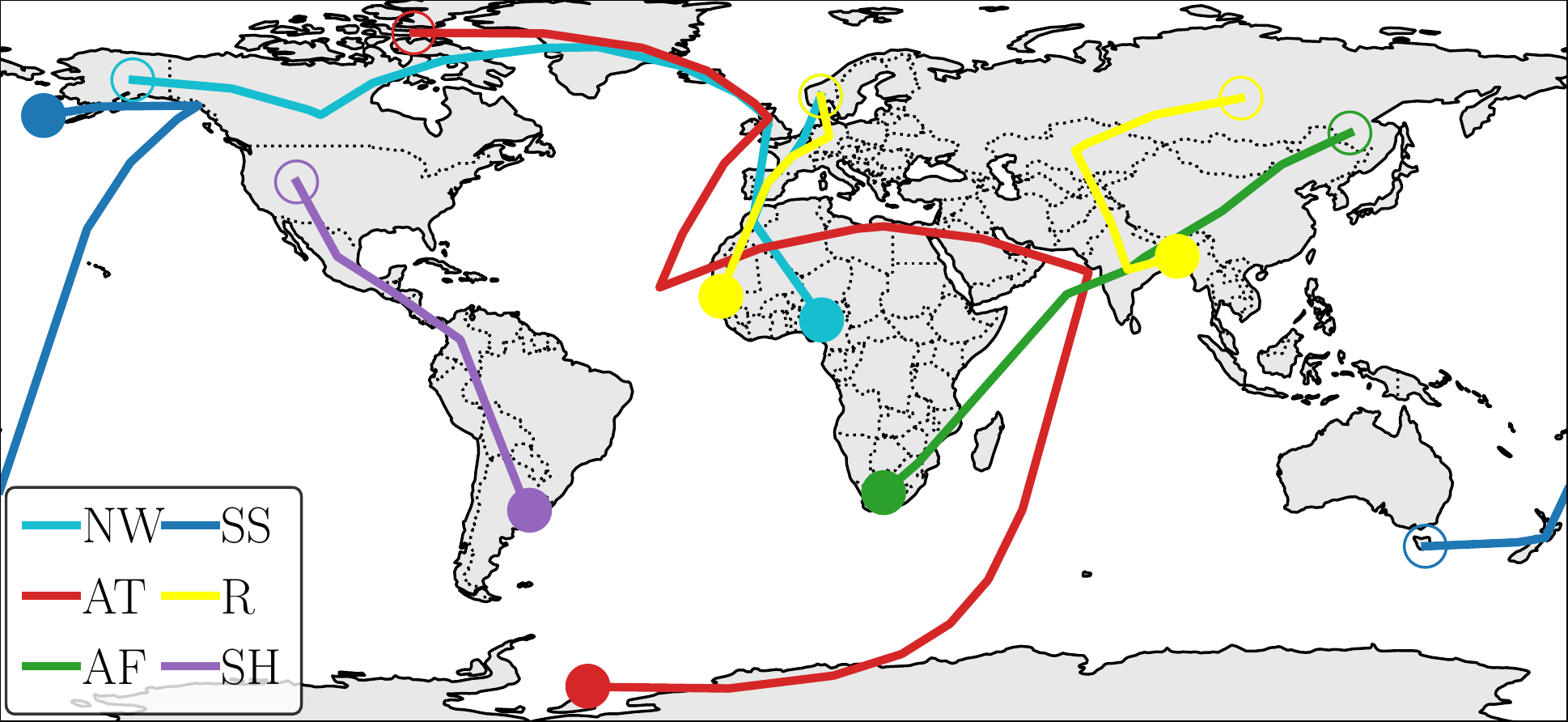}
    \caption{Bird migratory routes.}
    \label{fig:b1}
  \end{subfigure}
  \caption{\textbf{Natural world} ranges and routes. (Bird species: NW=Northern Wheatear, AT=Arctic Tern, AF=Amur Falcon, SS=Short-tailed Shearwater, R=Ruff, SH=Swainson's Hawk).}
  \label{fig:natural_world}

\end{figure}

\textbf{Natural world}. We briefly investigate GPT-4's understanding of wildlife ranges and migrations (see Fig. \ref{fig:natural_world}). For tiger subspecies, GPT-4 correctly identifies the 6 living subspecies and their relative ranges. However, the ranges are a combination of current and historic, and are partly incorrect with some (e.g., Malayan) largely in the ocean. For bird migrations, start and end locations are generally correct and the route is highly accurate in some cases (e.g., SH, AF, NW). In a few cases, additional routes are missed. GPT-4 provides reasonable suggestions for how the routes will shift with climate change (e.g., wintering further north, at higher elevations or earlier). To obtain coordinates for the migratory routes we use the following prompt:

\begin{formattedquote}
    I want to plot the migratory routes of various bird species. Please provide the latitude and longitude coordinates of the migratory route of the \textit{<bird species>}. Indicate the start and finish and provide coordinates for intermediate steps if necessary. If multiple routes exist, provide separate lists of coordinates.
    Output the coordinates as a python list of tuples.
\end{formattedquote}

\subsubsection*{Discussion}
GPT-4 proves to be skillful at solving a variety of application-centric tasks, further highlighting its potential for downstream tasks. The model can come up with creative, plausible travel routes (though it can be inaccurate with specifics) and shows strong capabilities of direction-based navigation. 
In plotting the landmark outlines, we illustrate GPT-4's ability to ``see" despite being a language model (similar to \cite{bubeck2023sparks}); although correct in aspect and content, deficits in composition can be observed. 
GPT-4 has a clear strength in tasks involving integrating knowledge from various (possibly unstructured) sources across different domains. 
In practice, the output can serve as proposals that need to be checked in a second stage.
On the other hand, we find that GPT-4 struggles with abstract optimisation reasoning without relation to knowledge, leading to the question: to what degree are tasks solved via memorisation rather than reasoning? Given the variability of tasks it solves it seems unlikely to memorise everything, but some things appear to be memorised.

\section{Conclusions}
\label{sec:iand_fd}

We find GPT-4 to have a remarkable understanding of the world, as evidenced by many examples showcasing strong underlying knowledge and logical reasoning. 
We hope our investigation contributes to useful applications of large language models in the geospatial domain.
If provided with access to geographical data feeds, it may ultimately be possible to create tools that improve travel planning and navigation. 
Despite these impressive examples, it is important to acknowledge that GPT-4 has knowledge gaps and remains prone to hallucination. 
A key research question highlighted by this work is the degree to which tasks are solved via memorisation or by reasoning.

\textbf{Broader Impacts}. The knowledge and reasoning abilities of GPT-4 towards geographic tasks 
demonstrate its potential to benefit myriad professions, especially in the travel industries. 
However, as well as serving as a tool to augment human workers, utilisation of GPT-4, or successors, could also replace human jobs, potentially causing significant economic and social challenges.
Looking to the future, if frontier models beyond GPT-4 continue to advance in capabilities, the geographic knowledge and planning abilities present in the current model may later evolve to represent a significant risk, through misuse or misalignment~\cite{shevlane2023model}.

\textbf{Limitations}. 
Full reproducibility is not possible for this work due to the stochastic nature of GPT-4\footnote{Full reproducibility could be achieved by setting the temperature parameter to 0,
however, constraining the model to be deterministic reduces its creativity, hence does not provide a true measure of capability.} and the GPT-4 updates. This, coupled with the breadth of the geographic field, necessitates the mostly qualitative approach we take, preventing us from making comprehensive claims about what GPT-4 knows about the world. Limited access to the model, and limited output and context windows, further restricts the details to which we can conduct our experiments. Finally, access to the visual component of GPT-4 could enable further capabilities; it would be interesting to see how it performs on remote sensing imagery, e.g., the SATIN benchmark \cite{roberts2023satin}.

\section*{Acknowledgements}
This work was supported by the UKRI Centre for Doctoral Training in Application of Artificial Intelligence to the study of Environmental Risks (reference EP/S022961/1), an Isaac Newton Trust grant, an EPSRC HPC grant, the Deutsche Forschungsgemeinschaft (DFG, German Research Foundation) - Project-ID 494541002, the Hong Kong Research Grant Council - Early Career Scheme (Grant No. 27208022), HKU Seed Fund for Basic Research, and the Bangabandhu Science and Technology Fellowship Trust. Samuel would like to acknowledge the support of Z. Novak and N. Novak in enabling his contribution.

\bibliographystyle{unsrt}
\bibliography{sample}

\section*{Appendix}
\renewcommand{\thesubsection}{\Alph{subsection}}{\label{app}}

We structure this Appendix to our main paper submission into two parts. First, we provide general remarks on our experimentation (Sec. \ref{sec:app-general_remarks}). Next, we include additional information on the investigations we carried out including: information required for reproducibility, prompts and extra supportive experiments (Sec. \ref{sec:app-extra_details}).

\subsection{General Remarks}
\label{sec:app-general_remarks}

A number of comments can be made about behaviours we observed throughout our experimentation. 

\textbf{Counting}. A common issue was GPT-4's inability to consistently accurately count. This was especially apparent when asking for specific numbers of outputs (e.g., coordinates for 50 points along the outline of a country or when asking for the distance between numerous pairs of coordinates), in which GPT-4 would frequently return a different length output. 
    
\textbf{Output format}. Another inconsistency is in the formatting of the output. For the majority of our experiments, we ask for output in python syntax, expecting the relevant part of the answer to be returned in a codeblock environment (when using the ChatGPT interface). However, output was sometimes returned as a python list in the code block (as expected) but sometimes as plaintext with or without linebreaks.

\textbf{Answering}. With some tasks, convincing GPT-4 to provide the desired out was difficult, requiring many re-runs. We found this typically for tasks that are perhaps not obviously expected of an LLM, such as providing the coordinates for area outlines or SVG code for landmarks.

\subsection{Additional Details, Prompts and Supportive Experiments}
\label{sec:app-extra_details}

Here, we progress through each investigation from Section 4 of the main paper. We include details of the specific prompts used and, where relevant, supplementary details to aid in the understanding of how we conducted each experiment. Furthermore, we present results for additional experiments that were carried out as part of this work that, due to length constraints, we were unable to incorporate into the main paper.

\subsubsection{\texorpdfstring{Population, Life Expectancy and CO\textsubscript{2} Emissions}{Population, Life Expectancy and CO2 Emissions}}

\textbf{Population}

We used the following prompt for country population estimation:
\begin{formattedquote}
    For each of the following countries, provide their population in \textit{<Year>} as a python list in the following format: 
    \\\
    [Population\_of\_Country\_1, \# Country 1 \\
    Population\_of\_Country\_2, \# Country 2, ...]
    \\ \\
    \textit{[<Country\_1>, <Country\_2>, ...]}
\end{formattedquote}

Country names were taken from the ground truth \cite{worldbank2023pop}.

\textbf{Life Expectancy}

We used the following prompt for country life expectancy estimation:
\begin{formattedquote}
    For each of the following countries, provide an estimate of the life expectancy at birth, as of 2020. Provide the life expectancies in years as a python list in the following format:
    \newline
    [Country1\_Life\_Expectancy, \# Country 1 Name \newline
    Country2\_Life\_Expectancy, \# Country 2 Name, \newline
    ... ]
    \newline
    Note: life expectancy at birth indicates the number of years a newborn infant would live if prevailing patterns of mortality at the time of its birth were to stay the same throughout its life.
    \\ \\
    \textit{[<Country\_1>, <Country\_2>, ...]} \newline
    Just to length constraints, output the python list, nothing else.
\end{formattedquote}

Country names were taken from \cite{worldbank2023le}. The ground-truth data contained numerous entries for regions that are not countries, such as territories (e.g., Cayman Islands), special administrative regions (e.g., Macao), and other categories (e.g., Heavily indebted poor countries (HIPC)). We disregarded the estimations for these regions.

\textbf{CO\textsubscript{2} Emissions}

We used the following prompt for country CO\textsubscript{2} emissions estimation:

\begin{formattedquote}
    For each of the following countries, provide an estimate for the CO2 emissions (in metric tons per capita) from the year 2019. CO2 emissions are defined as: Carbon dioxide emissions are those stemming from the burning of fossil fuels and the manufacture of cement. They include carbon dioxide produced during consumption of solid, liquid, and gas fuels and gas flaring.
    \newline
    Output a python list of the form: \newline
    [CO2\_Emissions\_Country1, \#Country1 \newline
    CO2\_Emissions\_Country2, \# Country2 \newline
    ...]
    \newline
    For queried regions that are not countries, return None. 
    \newline
    Countries: \newline
    \textit{[<Country\_1>, <Country\_2>, ...]} \newline
\end{formattedquote}

Country names were taken from the ground truth \cite{worldbank2023co2}. As before, the ground-truth data contained numerous entries for regions that are not countries. GPT-4 successfully returned `None' for these.

\subsubsection{Area}

We used the following prompt for country area estimation:

\begin{formattedquote}
    For each of the following countries, provide the land area in sq. km as of \textit{<Year>}. Provide the areas as a python list in the following format:
    \\\
    [Area\_of\_Country\_1, \# Country 1 \\
    Area\_of\_Country\_2, \# Country 2, ...]
    \\ \\
    \textit{[<Country\_1>, <Country\_2>, ...]}
\end{formattedquote}

Country names were taken from the ground truth \cite{worldbank2023area}. As before, the ground-truth data contained numerous entries for regions that are not countries. GPT-4 successfully returned `None' for non-country categories (e.g., Heavily indebted poor countries (HIPC)).

\subsubsection{Height}

We used the following prompt for mountain height estimation:

\begin{formattedquote}
    Return the height in metres for the following mountains. Provide the heights in the following format: \newline
    [Height\_Mountain\_1, \# Mountain\_1\_Name, \newline
    Height\_Mountain\_2 \# Mountain\_2\_Name,...]
    \\ \\
    \textit{[<Country\_1>, <Country\_2>, ...]}
\end{formattedquote}

Mountain names were taken from the ground truth \cite{kaggle_mountains}.

\subsubsection{Location}

\textit{Name $\rightarrow$ Coordinate}

\begin{formattedquote}
    In a code block, provide a python list of tuples for the latitude and longitude coordinates for each of the following settlements in the format - [(Lat,Lon), \# Settlement 1, ...]. Maintain the same order.
    \\ \\
    \textit{[<Settlement\_1>, <Settlement\_2>, ...]}
\end{formattedquote}

Country names were taken from the ground truth \cite{kaggle_cities}.

\textit{Coordinate $\rightarrow$ Name}

\begin{formattedquote}
    In a code block, provide a python list of the name of the settlement at each of these coordinates- e.g., [Settlement1, Settlement2, ...]. Maintain the same order.
    \\ \\
    \textit{[<(Lat,Lon)>,...]}
\end{formattedquote}

Country lat/lon coordinates were taken from the ground truth \cite{kaggle_cities}.

\subsubsection{Distance Estimation}

We used the following prompt for distance estimation:
\begin{formattedquote}
    What are the straight-line distances between the following pairs of cities? Output a csv list of the distances only in km.\\\
    [\textit{<City 1 (Country)>} to \textit{<City 2 (Country)>},\\
    ... \textit{(repeated for all 40 pairs)}]
\end{formattedquote}
In cases where the number of outputs did not match the number of provided cities, we repeated the prompt.

We had to change the prompt because ChatGPT started to refuse to provide distances (possibly as a consequence of backend changes to ChatGPT's configuration).
On results indicated by asterisk (*) were obtained using the following prompt. 
\begin{formattedquote}
    Estimate the straight-line distances between the following pairs of cities. Output only a comma-separated Python list of rough distance estimates in km.\\\
    [\textit{<City 1 (Country)>} to \textit{<City 2 (Country)>},\\
    ... \textit{(repeated for all 40 pairs)}]
\end{formattedquote}

\subsubsection{Topography}

For the elevation plots shown in the main paper, %
we use the following prompts:

\begin{formattedquote}
Provide a rough estimate of the elevation at the following coordinates to the best of your knowledge. Answer directly with a comma-separated list of elevations in meters only but without indicating the unit in the output. \\
45.00000, 11.20000\\
45.33333, 11.23333\\
...
\end{formattedquote}

The respective coordinates for the lines are\\
blue: \\
{
\small
\texttt{
45.00000, 11.20000;
45.33333, 11.23333;
45.66667, 11.26667;
46.00000, 11.30000;
46.33333, 11.33333;
46.66667, 11.36667;
47.00000, 11.40000;
47.33333, 11.43333;
47.66667, 11.46667;
48.00000, 11.50000;
}
}
\\
\\
orange:\\
{
\small
\texttt{
45.00000, 5.20000;
45.02667, 5.64444;
45.05333, 6.08889;
45.08000, 6.53333;
45.10667, 6.97778;
45.13333, 7.42222;
45.16000, 7.86667;
45.18667, 8.31111;
45.21333, 8.75556;
45.24000, 9.20000;
}
}
\\
\\
green:\\
{
\small
\texttt{
43.00000, 10.20000;
43.23333, 10.53333;
43.46667, 10.86667;
43.70000, 11.20000;
43.93333, 11.53333;
44.16667, 11.86667;
44.40000, 12.20000;
44.63333, 12.53333;
44.86667, 12.86667;
45.10000, 13.20000;
}
}

We found that explicitly asking for coordinates works better than specifying trajectory endpoints only:
For the prompt
\begin{formattedquote}
    Provide a rough estimate of the elevation between the following coordinates. Please answer with a comma-separated list of 10 elevations in meter only.
between 43.00000, 10.20000 and 45.10000, 13.20000
\end{formattedquote}
GPT-4 predicted 200, 350, 450, 600, 800, 1000, 1200, 1400, 1600 and 1800, which is substantially worse than the predictions above.

\subsubsection{Outlines}

This section outlines the prompts we used for the Outlines experiment in the main paper. The prompts are arranged in order from L-R for subfigures in the figure in the main paper.

\textbf{Australia}

When creating the outline of Australia, we experiment with providing feedback on the outlines produced by GPT-4 in an attempt to incrementally update and improve the outline.

Here is our starting prompt:

\textit{Iteration 0}.
\begin{formattedquote}
    Please provide the lat/lon coordinates for the outline of Australia as a Python list, consisting of approximately 75 points arranged clockwise. Due to output length limitations, only the coordinates should be returned.
\end{formattedquote}

These are the incrementally suggested improvements.

\textit{Iteration 1}. 
\begin{formattedquote}
    That's a decent outline. However it could be improved. Currently the south-west of the Western Australia state is not included in the outline, please correct this to include it. Also, the outline includes the Gulf of Carpentaria as if it were part of the Australian landmass, please include update the outline to match the coastline.
\end{formattedquote}

\textit{Iteration 2}.
\begin{formattedquote}
    The new outline is improved. Currently the outline has very little detail around Spencer Gulf, please add this in. Furthermore, the outline could still be improved around Western Australia. Extend the outline so that it includes settlements such as Albany and Busselton. Return an outline with these improvements.
\end{formattedquote}

\textit{Iteration 3}.
\begin{formattedquote}
    Those changes further improved the outline. Extend the outline near Western Australia to include Geraldton. Also, the outline around the north of Western Australia, specifically from Exmouth to Broome needs correcting. Please update the outline with these changes.
\end{formattedquote}

\textit{Iteration 4}.
\begin{formattedquote}
    The outline around the Western Australia coastline is still very jagged. Smooth this out to better match the coastline.
\end{formattedquote}

\textit{Iteration 5}.
\begin{formattedquote}
    The Eastern half of the outline is very accurate. However, the Western half needs to be improved. The very Western edge of the outline includes numerous jagged sections despite the actual coastline being fairly smooth. In the North, more detail needs to be added on the coast near the Timor Sea and Arafura Sea. Update the outline coordinates with these changes.
\end{formattedquote}

\textit{Iteration 6}.
\begin{formattedquote}
    That's improved some parts but made some worse. From Shark Bay in the West down towards Perth should be a smooth coastline. The outline you proposed juts into the centre of Western Australia. Correct the outline so that it reflects the true coastline. Include all the previous improvements.
\end{formattedquote}

\textbf{Continental USA}

\begin{formattedquote}
    Please provide the lat/lon coordinates for the outline of the Continental United States as a Python list of tuples, consisting of approximately 50 points arranged clockwise. Due to output length limitations, only the coordinates should be returned.
\end{formattedquote}

\textbf{Lake Winnipeg}

\begin{formattedquote}
    I want to draw an outline of Lake Winnipeg using at least 50 points. Provide just a python list of tuples of the lat/lon coordinates for the outline of the lake clockwise from the North. Due to length limitations, just provide the coordinates with no additional text or comments.
\end{formattedquote}

\textbf{Lake Superior}

\begin{formattedquote}
    I want to draw an outline of Lake Superior using at least 50 points. Provide just a python list of tuples of the lat/lon coordinates for the outline of the lake clockwise from the North. Due to length limitations, just provide the coordinates with no additional text or comments.
\end{formattedquote}

\textbf{UK Rivers}

\begin{formattedquote}
    I want to draw an outline of the \textit{<river\_name>} in the UK using at least 50 points. Provide just a python list of tuples of the lat/lon coordinates for the outline of the river from East to West. Approximate the river as a 2D line, ignoring its width. Due to length limitations, just provide the coordinates with no additional text or comments.
\end{formattedquote}
Where \textit{\textless river\_name\textgreater} = Thames, Trent, Severn.

\textbf{France Rivers}

\begin{formattedquote}
    I want to draw an outline of the \textit{<river\_name>} in France using at least 50 points. Provide just a python list of tuples of the lat/lon coordinates for the outline of the river from North to South. Approximate the river as a 2D line, ignoring its width. Due to length limitations, just provide the coordinates with no additional text or comments.
\end{formattedquote}
Where \textit{\textless river\_name\textgreater} = Loire, Seine.

\textbf{Continents}

\begin{formattedquote}
    Please provide the lat/lon coordinates for the outline of Africa (ignoring any islands) as a Python list, consisting of approximately 50 points arranged clockwise. Due to output length limitations, only the coordinates should be returned.
\end{formattedquote}

\begin{formattedquote}
    I want to draw an outline of South America using at least 50 points. Provide just a python list of tuples of the lat/lon coordinates for the outline of the continent clockwise from the North. Due to length limitations, just provide the coordinates with no additional text or comments.
\end{formattedquote}

\subsubsection{Languages.} 

To assess GPT4's knowledge of spoken language distribution we used the following prompts (1st prompt):
\begin{formattedquote}
    Name all countries with more than 3 official languages in 2021. Provide the output as a python list.
\end{formattedquote}

\begin{formattedquote}
Name all countries in the world where both a Romance language and a Germanic language were widely spoken in 2021. Provide the output as a python list.
\end{formattedquote}

\begin{formattedquote}
Name all countries where both English and French were widely spoken in 2021. Provide the output as a python list.
\end{formattedquote}

\subsubsection{Route Planning}

\begin{figure}
    \centering
    \includegraphics[width=\textwidth]{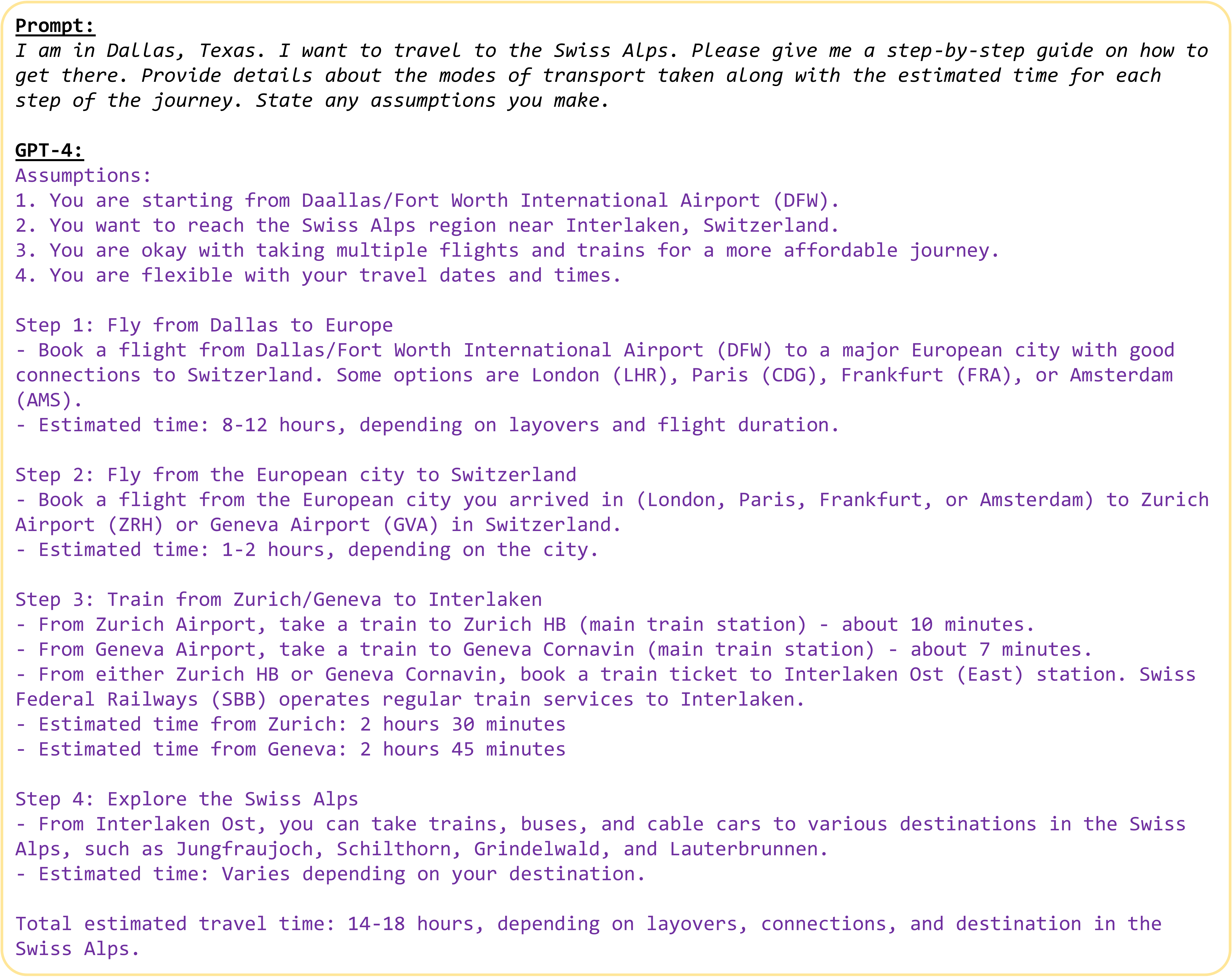}
    \caption{Model response for the journey prompt between \textit{Dallas, Texas} to \textit{The Swiss Alps}.}
    \label{prompt:swiss}
\end{figure}

\paragraph{General Planning.} General route planning involves specifying a source and requesting a path to reach the required destination. GPT-4 is adept at these queries because they do not impose any constraints, and allows flexibility in choosing the mode of travel. Fig. \ref{prompt:swiss} shows the prompt and response for the trip from \textit{Dallas, Texas} to \textit{The Swiss Alps}. The model is capable of breaking down the journey into multiple segments and solve the problem similar to how a human would.

\begin{figure}[!h]
    \centering
    \includegraphics[width=\textwidth]{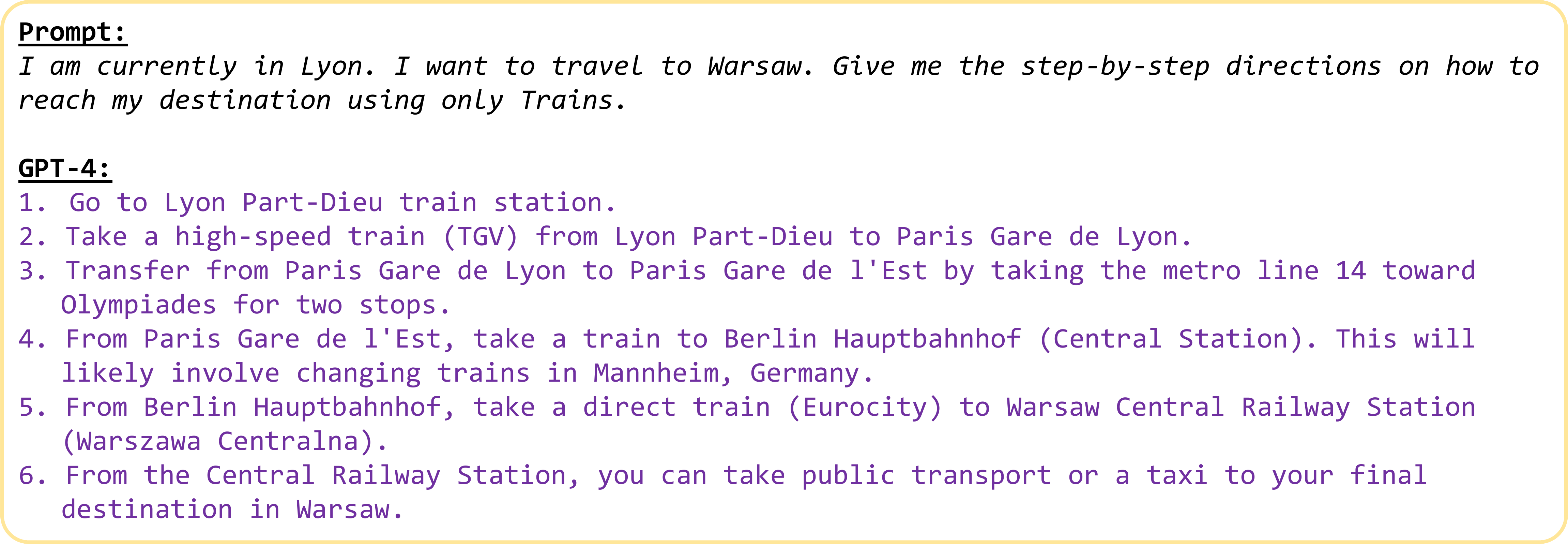}
    \caption{Prompt for the travel path from \textit{Lyon, Paris} to \textit{Warsaw, Poland} as shown in Fig. \ref{map:lyon}.}
    \label{prompt:lyon}
\end{figure}

\begin{figure}[!h]
    \centering
    \includegraphics[width=\textwidth]{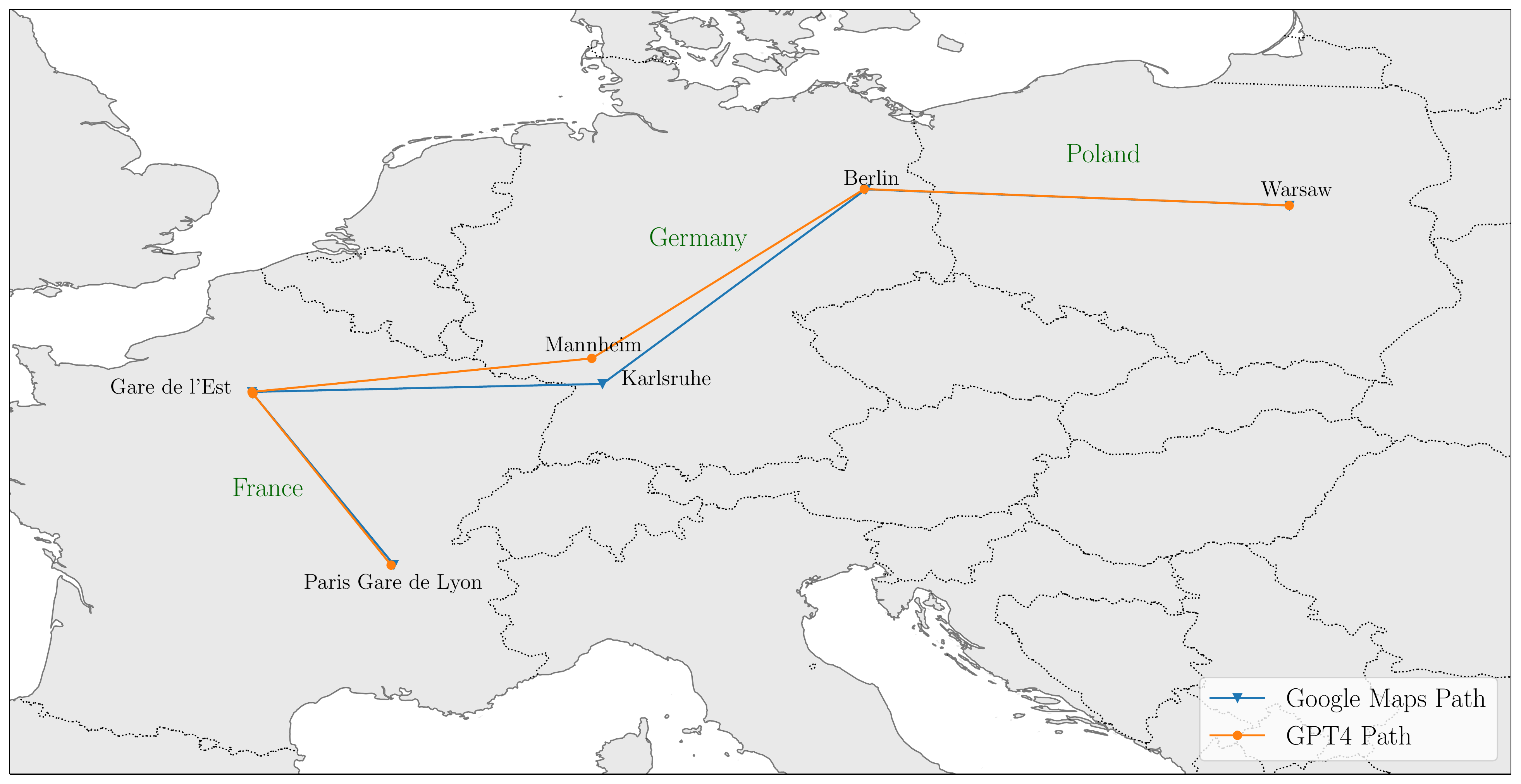}
    \caption{Comparison of travel route prediction from \textit{Lyon, France} to \textit{Warsaw, Poland} using only trains for the prompt in Fig. \ref{prompt:lyon}. The GPT-4 predicted route is one of many plausible ones, and matches very closely with the best path predicted by Google Maps.}
    \label{map:lyon}
\end{figure}

\begin{figure}[!h]
    \centering
    \includegraphics[width=\textwidth]{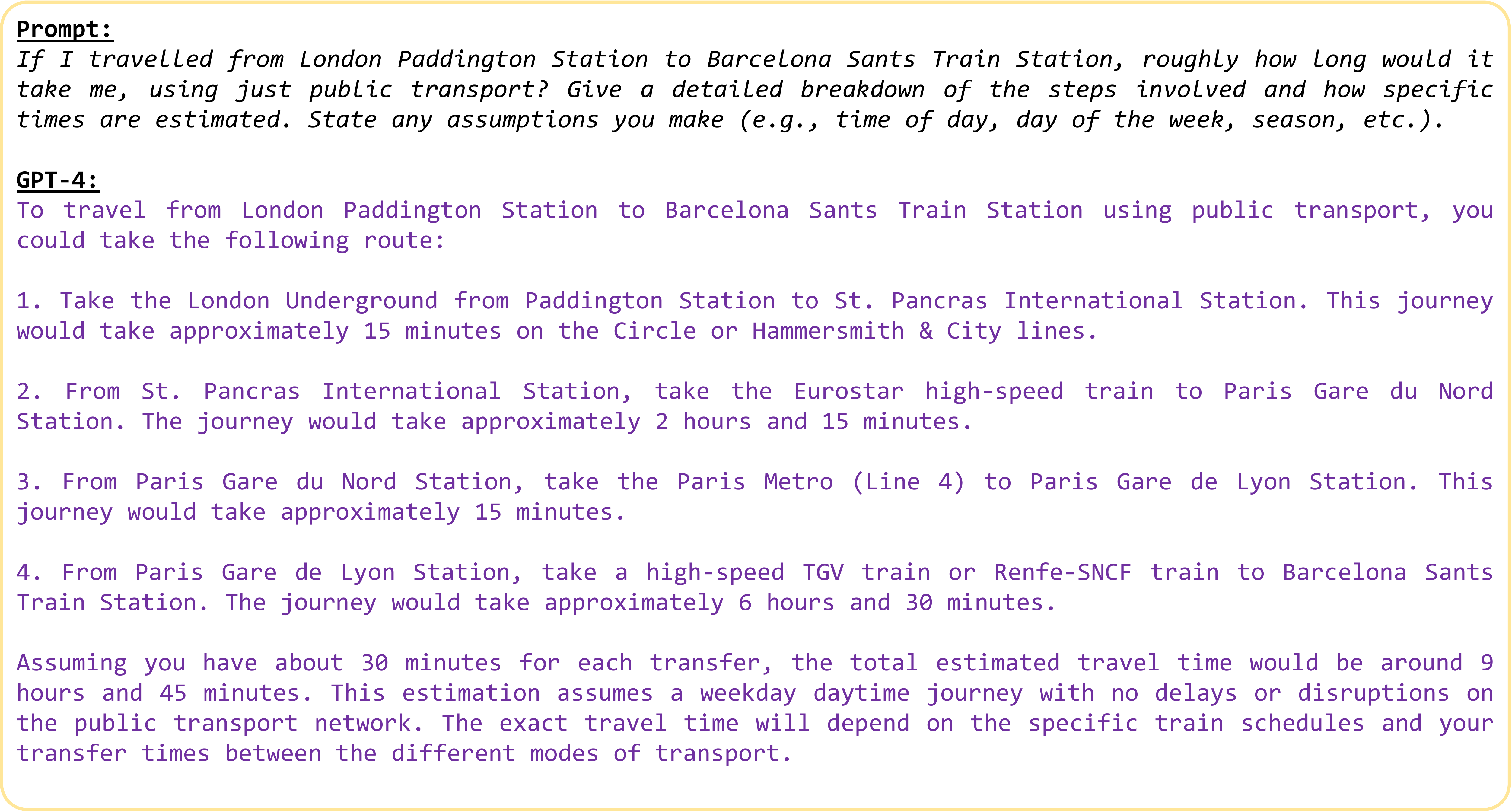}
    \caption{Travel between \textit{London Paddington Station} to \textit{Barcelona Sants Station}. The estimated time and intermediate steps match accurately with predictions by Google Maps. \vspace{-10pt}}
    \label{prompt:travel_time}
\end{figure}

\paragraph{Trains.} In Fig. \ref{prompt:lyon} we prompt the model to provide a travel route using only trains from \textit{Lyon} to \textit{Warsaw}. The model's suggested path matches closely with the recommended route by Google Maps. Based on the current availability of buses and trains, the google maps api forecasts a number of potential routes. In contrast, GPT-4 most of the time only specifies one probable route. However, the model was able to correctly identify the individual train stations and lines that must be taken in order to get to the desired location, which closely fits the groundtruth prediction (Fig. \ref{map:lyon}). Additionally, we asked the model to predict how long each leg of a trip from \textit{London Paddington Station} to \textit{Barcelona Sants Station} would take (Fig. \ref{prompt:travel_time}). The model's predictions for each leg of the journey agree with Google Maps' best estimates very well.

\begin{figure}
\centering
  \begin{subfigure}[b]{0.49\textwidth}
  \centering
    \includegraphics[width=\textwidth, height=8.5cm]{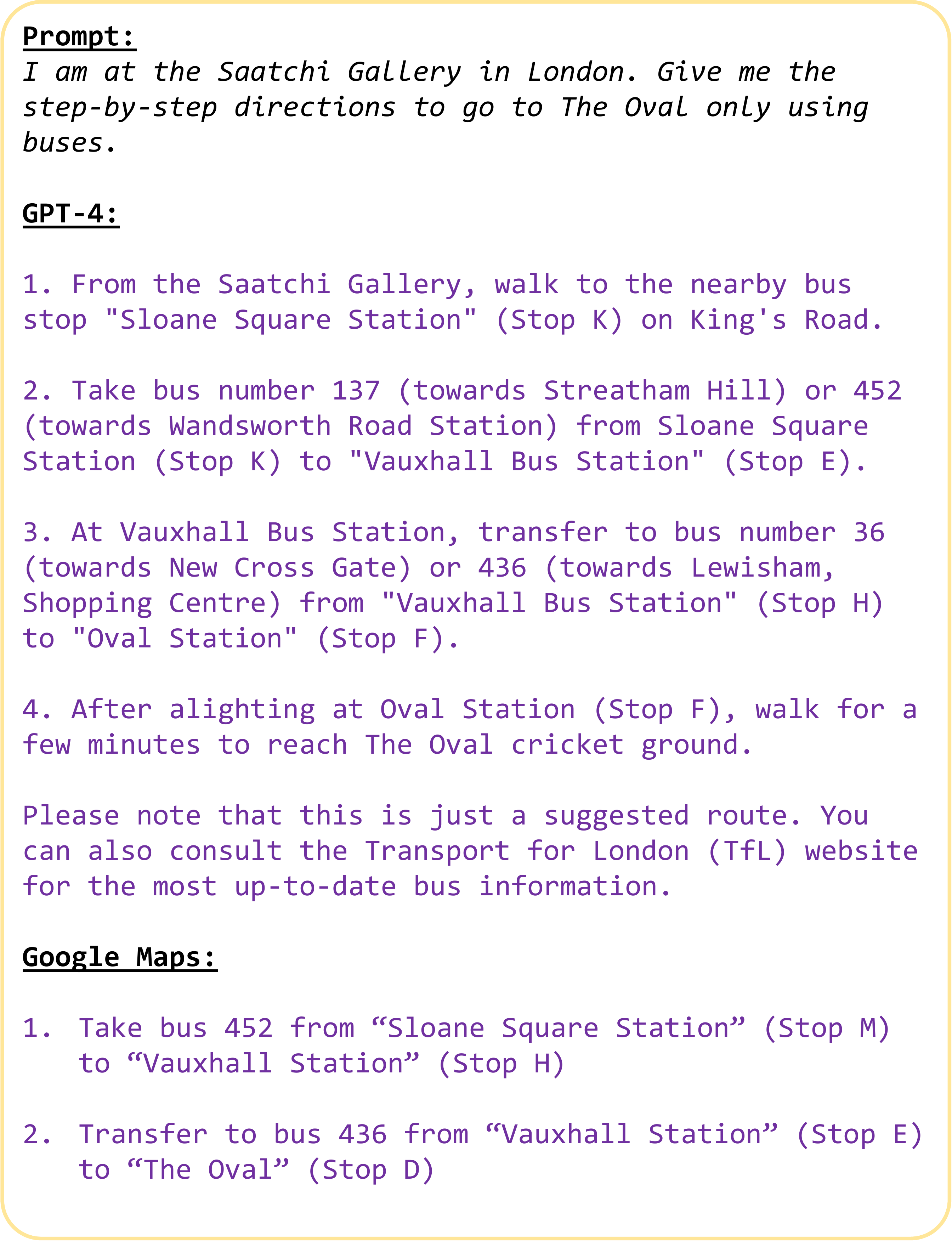}
    \caption{Bus route, \textit{Saatchi Gallery} $\rightarrow$ \textit{The Oval}}
    \label{prompt:saatchi}
    \end{subfigure}
    \begin{subfigure}{0.47\textwidth}
    \centering
    \includegraphics[width=\textwidth, height=8.4cm]{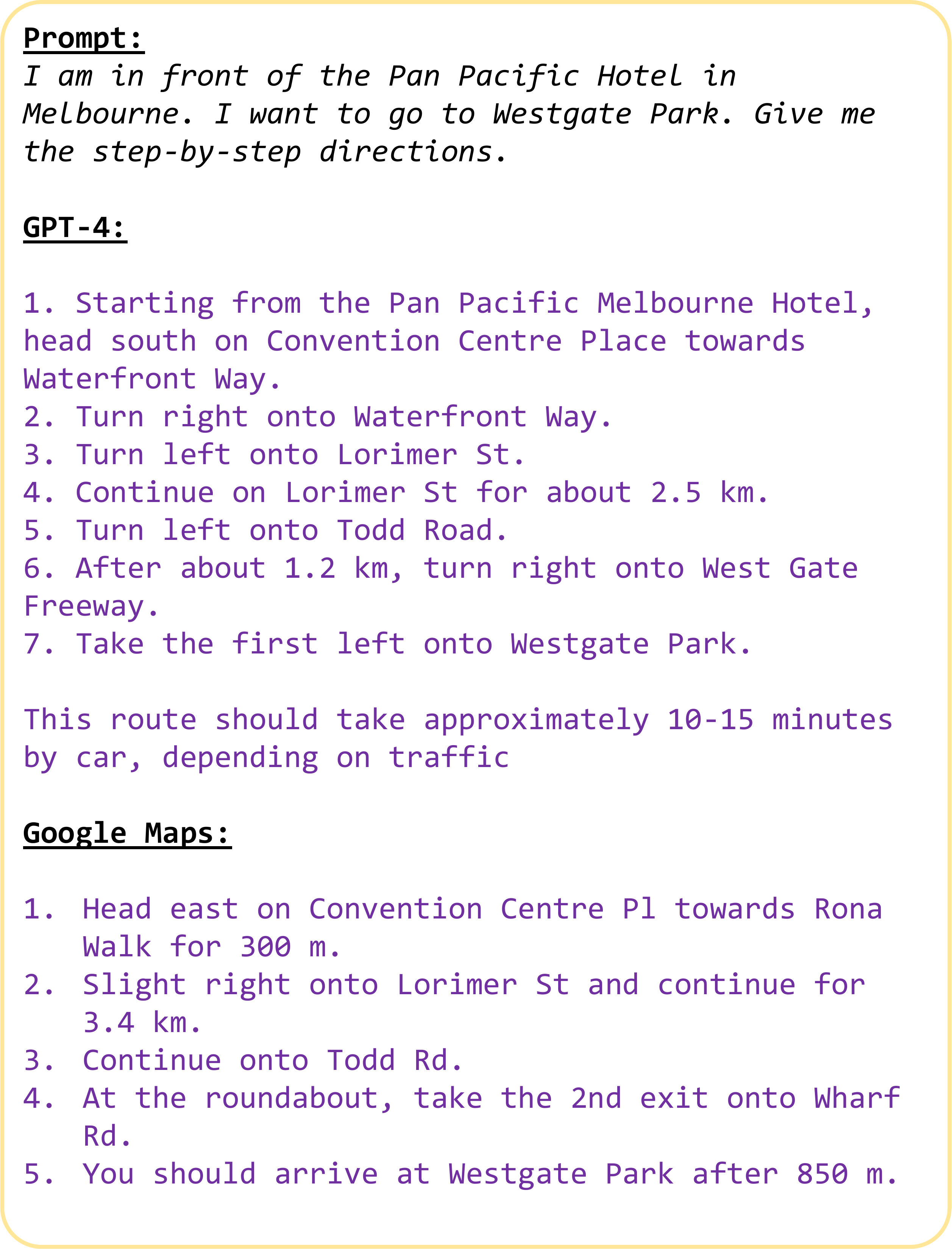}
    \caption{Car route, \textit{Pan Pacific Hotel} $\rightarrow$ \textit{Westgate Park}}
    \label{prompt:pan_pacific}
  \end{subfigure}
    \caption{Travel routes for car and bus travel to specific destinations within different cities.}
\end{figure}

\begin{figure}
    \centering
    \includegraphics[width=0.93\textwidth]{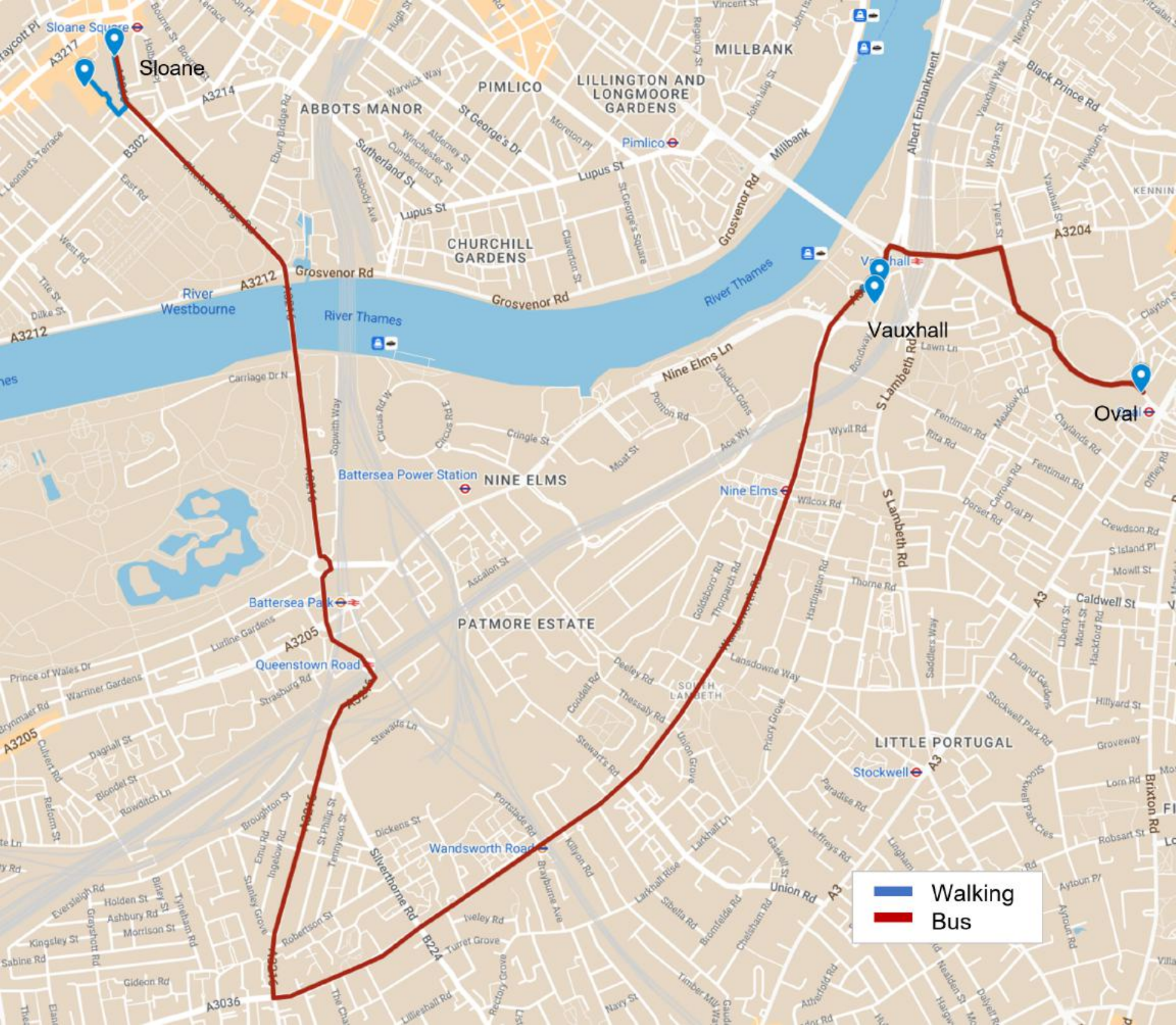}
    \caption{The planned bus route proposed by GPT-4 in the prompt Fig. \ref{prompt:saatchi} as verified by the official TFL (Transport for London) database.The model correctly predicts where the buses need to be changed as well as waking route to the station.}
    \label{fig:bus_travel}
\end{figure}

\paragraph{Buses.} Compared to train or aviation routes, navigation with buses are more difficult because they can change overtime. Since GPT-4 has a knowledge cutoff after 2021, it does not know the realtime bus route information (which it is aware of). In Fig. \ref{prompt:saatchi} we prompt the model to generate a bus route from  \textit{Saatchi Gallery, London} to \textit{The Oval, London}. The output includes a combination of walking and travel by bus along with specifications on where to change buses and the individual stoppages (Fig. \ref{fig:bus_travel}). We verified the proposed route with the official TFL (Transport for London) bus route database and found that the bus specified in the response runs between the specified bus stops. However, the exact stop number is inaccurate. The current database shows that the $452$ bus leaves from stop \textit{M} whereas GPT-4 suggested stop \textit{K}. 

\begin{figure}
    \centering
    \includegraphics[width=\textwidth]{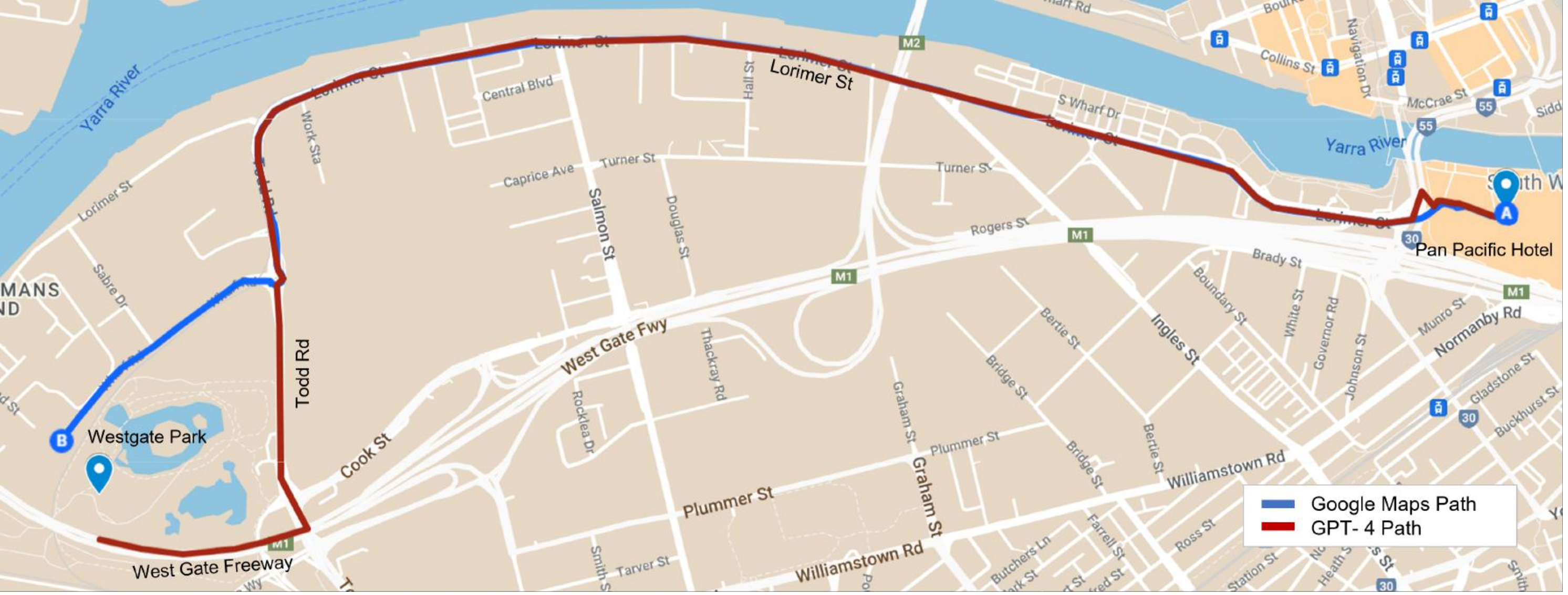}
    \caption{Comparison of predicted paths by GPT-4 and Google Maps for travel by car from \textit{The Pan Pacific Hotel, Melbourne} to \textit{Westgate Park} for the prompt Fig. \ref{prompt:pan_pacific}.}
    \label{fig:panpacific}
\end{figure}

\paragraph{Driving.} Intra or Inter-city driving route prediction in comparison to other modes of transport are fairly trivial because road networks remain consistent over time, and is easy to locate in terms of distance measure and landmark positioning. Even with a very simple prompt such as \textit{Tell me how to go from 'A' to 'B'} is enough to generate a fairly accurate travel plan. We show a comparison between GPT-4 output (Fig. \ref{prompt:pan_pacific}) and Google Maps prediction for a driving route from \textit{The Pan Pacific Hotel, Melbourne} to \textit{Westgate Park, Melbourne} (Fig. \ref{fig:panpacific}).

\subsubsection{Navigation}

\begin{figure}
\centering
  \begin{subfigure}[b]{0.49\textwidth}
  \centering
    \includegraphics[width=\textwidth, height=14.7cm]{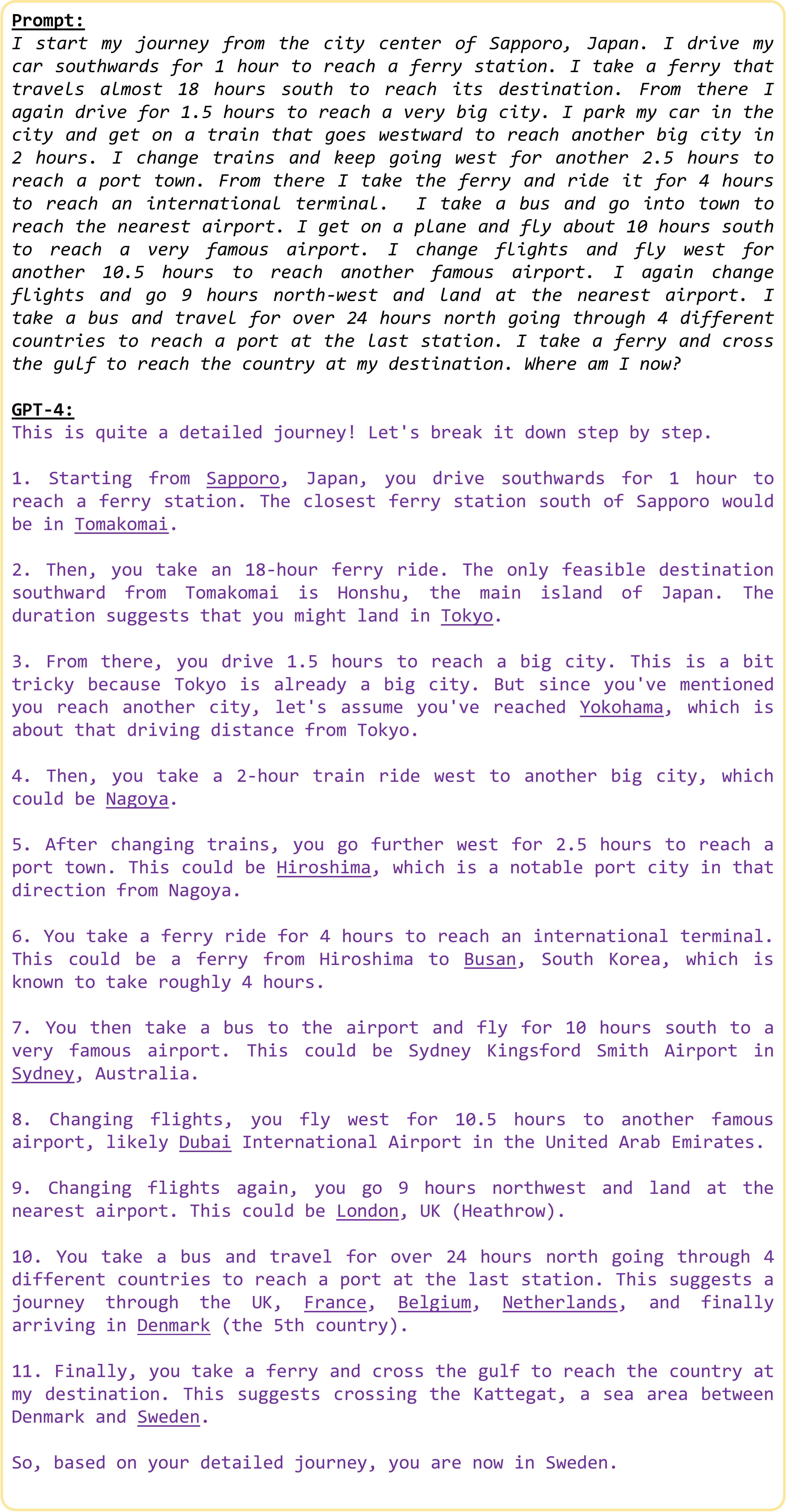}
    \caption{Initial prompt.}
    \end{subfigure}
    \hfill
    \begin{subfigure}{0.49\textwidth}
    \centering
    \includegraphics[width=\textwidth, height=14.5cm]{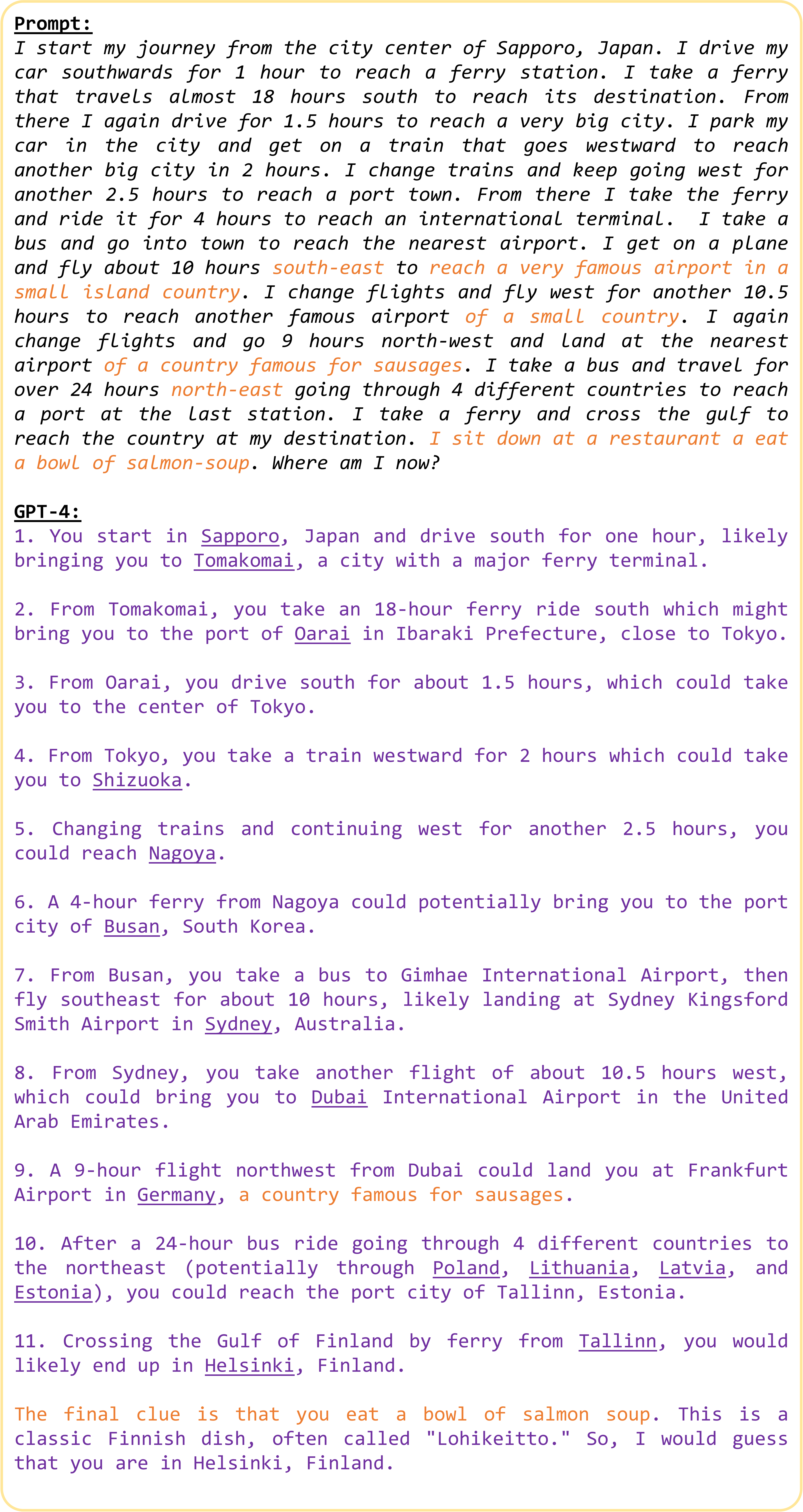}
    \caption{Refined prompt.}
  \end{subfigure}
    \caption{Responses for the initial and refined prompts for the more complex journey illustrated in Fig. \ref{fig:country_journey}. Difference between the prompts are highlighted in \textcolor{orange}{orange}.}
    \label{prompt:sapporo}
\end{figure}

\begin{figure}
    \centering
    \includegraphics[width=\textwidth]{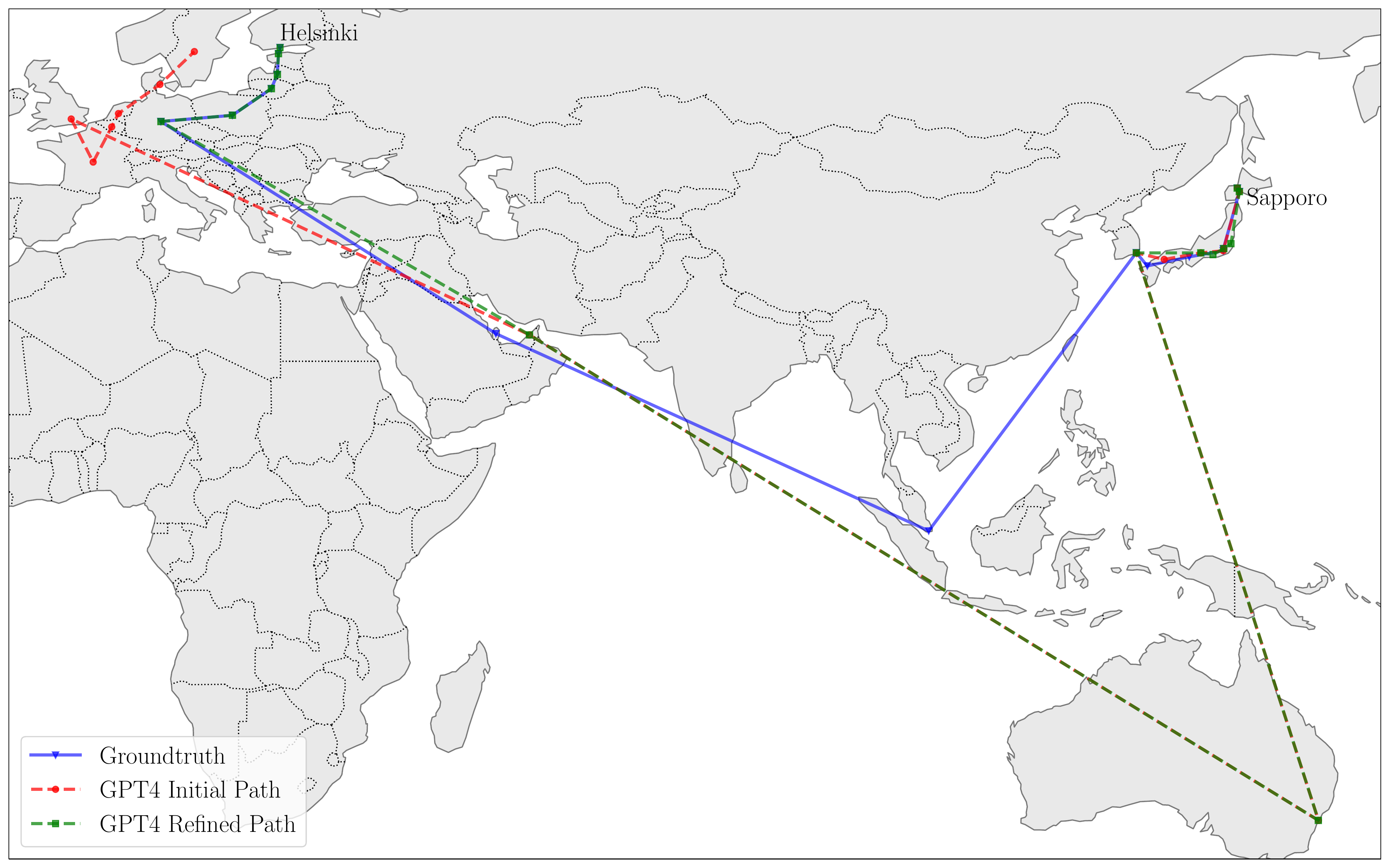}
    \caption{Illustration of the journey and intermediate stops outlined by GPT-4 against the groundtruth. We see that after refining the prompt for more specific information about the possible destinations, GPT-4 was better at identifying the described path.}
    \label{fig:country_journey}
\end{figure}

\paragraph{Long Distance Travel.} This includes navigation between multiple countries and involving different modes of transport given only direction or distance based queries.  Fig. \ref{prompt:sapporo} shows the prompt for journey from \textit{Sapporo, Japan} to \textit{Helsiniki, Finland} consisting of 15 intermediate stops. We found that extending the prompt with additional steps has minimal effect on the model's response as long as it keeps within the context window. Since the model is breaking down each leg of the journey and evaluating step-by-step, long journeys are the same as shorter ones. However, there is a high degree of uncertainty in the response if the prompt contains unspecified directional locations. Such as, it is possible to reach both \textit{Kyoto} and \textit{Osaka} via a two hour train ride from \textit{Tokyo}. This type of randomness is observable when travelling between countries in Europe, states in the US, and islands in South Asia. If the prompts are refined with small amounts of information regarding the particular waypoints, the model can better differentiate between these neighboring regions. GPT-4 was able to correlate \textit{Salmon-Soup} with \textit{Helsinki} and accurately arrived at the final location. The whole journey is visualized in Fig. \ref{fig:country_journey}. 

\begin{figure}
    \centering
    \includegraphics[width=\textwidth]{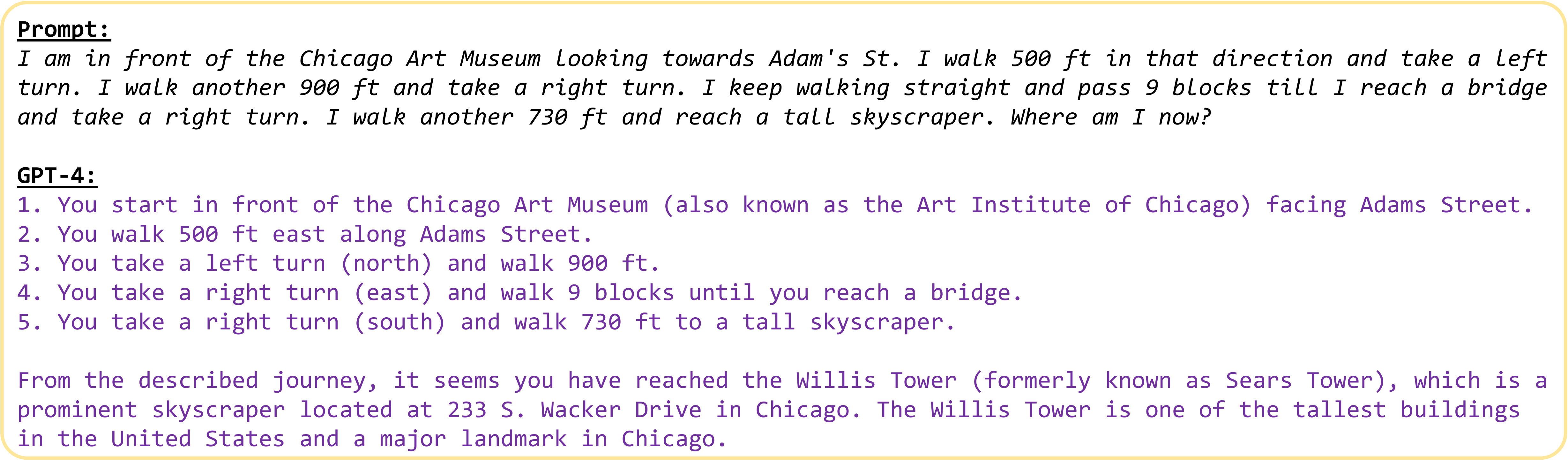}
    \caption{Response for short distance based navigation in Chicago. The journey is visualized in Fig. \ref{fig:chicago}.}
    \label{prompt:chicago}
\end{figure}

\paragraph{Short Distance Travel.} In addition to travel between countries, we also evaluated if GPT-4 can identify short immediate distances between buildings within a city (Fig. \ref{prompt:chicago}). We specify the distances in \textit{feet (ft)} and direction in the perspective of someone walking down the road. We found that the model can accurately track the distances and turns along the road to reach the destination in more well documented cities like \textit{New York} or \textit{Chicago}, but not for small paces like \textit{Daventry, UK}.

\begin{figure}[!t]
    \centering
    \includegraphics[width=\textwidth]{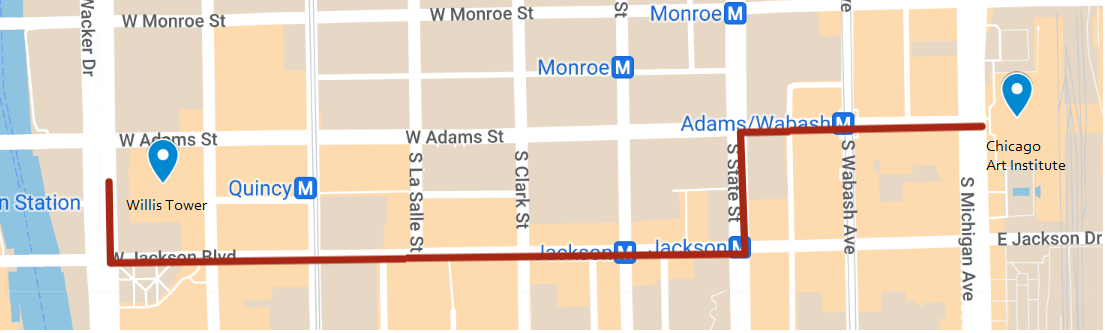}
    \caption{Illustration of the short distance travel in Chicago from \textit{The Chicago Art Institute} to \textit{Willis Tower} described in prompt Fig. \ref{prompt:chicago}.}
    \label{fig:chicago}
\end{figure}

\paragraph{Time zones.} We use the following prompt to specify the journey $-$ $Sydney \rightarrow Dhaka \rightarrow Cape Town \rightarrow Norilsk \rightarrow Vancouver \rightarrow Sydney$, and evaluate if GPT-4 can keep track of changing time zones.

\begin{formattedquote}
    I have my own private jet and I can land anywhere in the world I wish. I start a journey from Sydney Australia. I take a flight and go to Dhaka, Bangladesh. From there, I go to Cape Town South Africa. Next, I go to Norilsk, Russia. Next, I go to Vancouver, Canada. From there I return to Sydney. I began my journey on 05/11/2023 (dd/mm/yyyy) at 10 am in Sydney. State the local times and dates for each country when I landed there. At what time will I return to Sydney? I spend no time at any of the stops. Break down your calculations and assumptions. 
\end{formattedquote}

\clearpage
\subsubsection{Networks}

\textbf{Hong Kong MTR Network}

We begin by prompting GPT-4 for an ordered list of the stations on each line. We find if we ask for coordinates for the stations without this initial prompt, the model misses 30-40\% of the stations on the line when we ask for coordinates.
\begin{formattedquote}
    Provide a list of the names of the stations in order on the Hong Kong MTR \textit{<Line\_Name>} Line. Give the latitude and longitude coordinates for each of these as a python list of tuples. Maintain the same order.
\end{formattedquote}

\textbf{Airport Network}

\begin{figure}[h]
    \includegraphics[width=\textwidth]{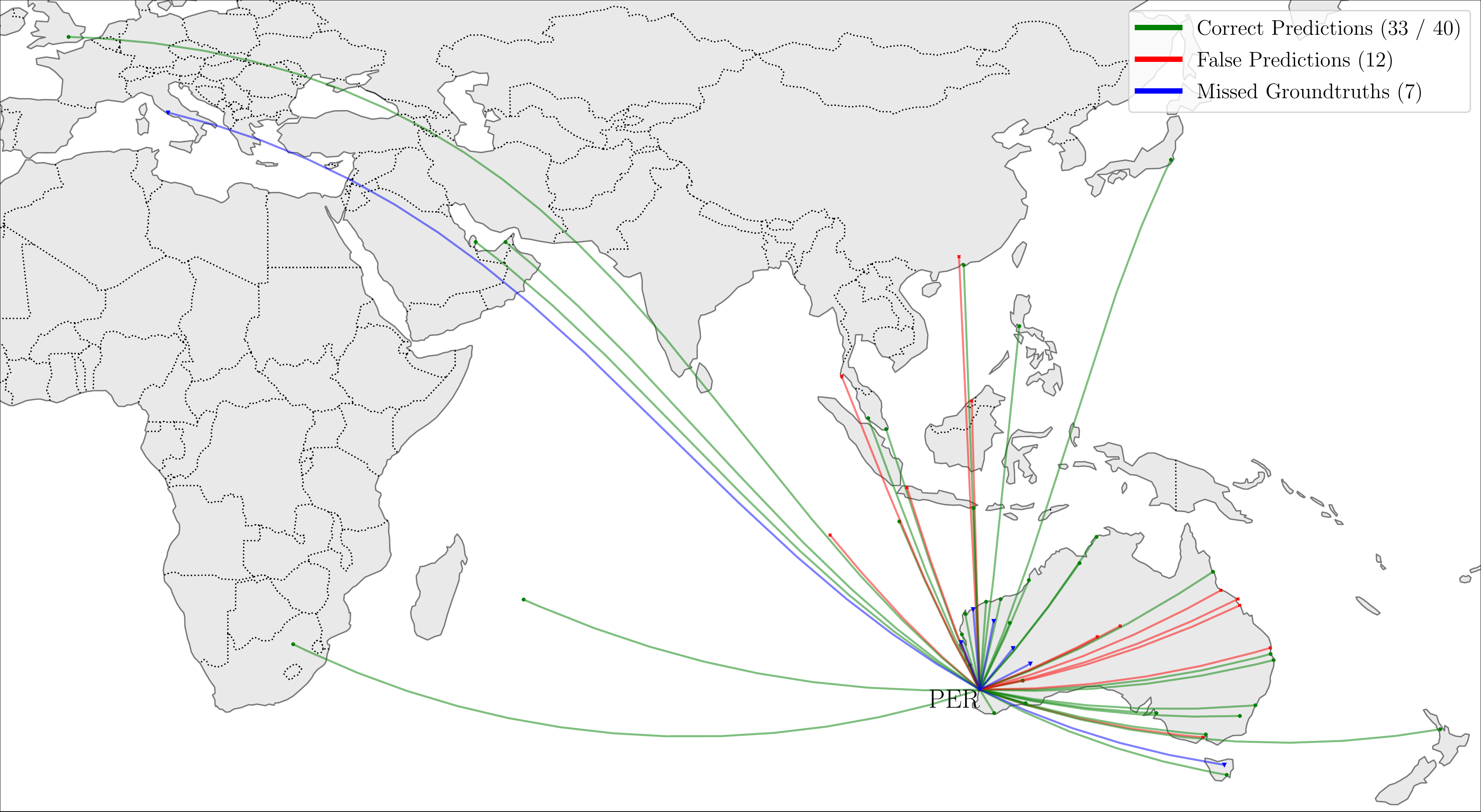}
  \caption{Visualizing the knowledge of GPT-4 about direct flight routes from \textit{Perth International (PER), Australia} to other countries.}
  \label{fig:perth}
\end{figure}

We tried to create the network of all airports that are directly connected to \textit{Perth International Airport (PER)}. We first used the following prompt,
\begin{formattedquote}
    Give me the list of coordinates for all airports directly connected with Perth International Airport.
\end{formattedquote}
However, the network only listed a few locations and responded that the full list of airports are too large for the response. So, we broke down the prompt into two query segments - list of all internal airport, and a list of all external airports. This was prompted as,
\begin{formattedquote}
    Give me a list of every airport <inside/outside> Australia that has direct flights from Perth Airport. Also provide their lat/lon coordinates.
\end{formattedquote}
The result when plotted gave the airport network in Fig. \ref{fig:perth}. GPT-4 was able to identify 33 out of the 40 airports, but also made 12 false predictions.

\begin{figure}[!h]
    \centering
    \includegraphics[width=\textwidth]{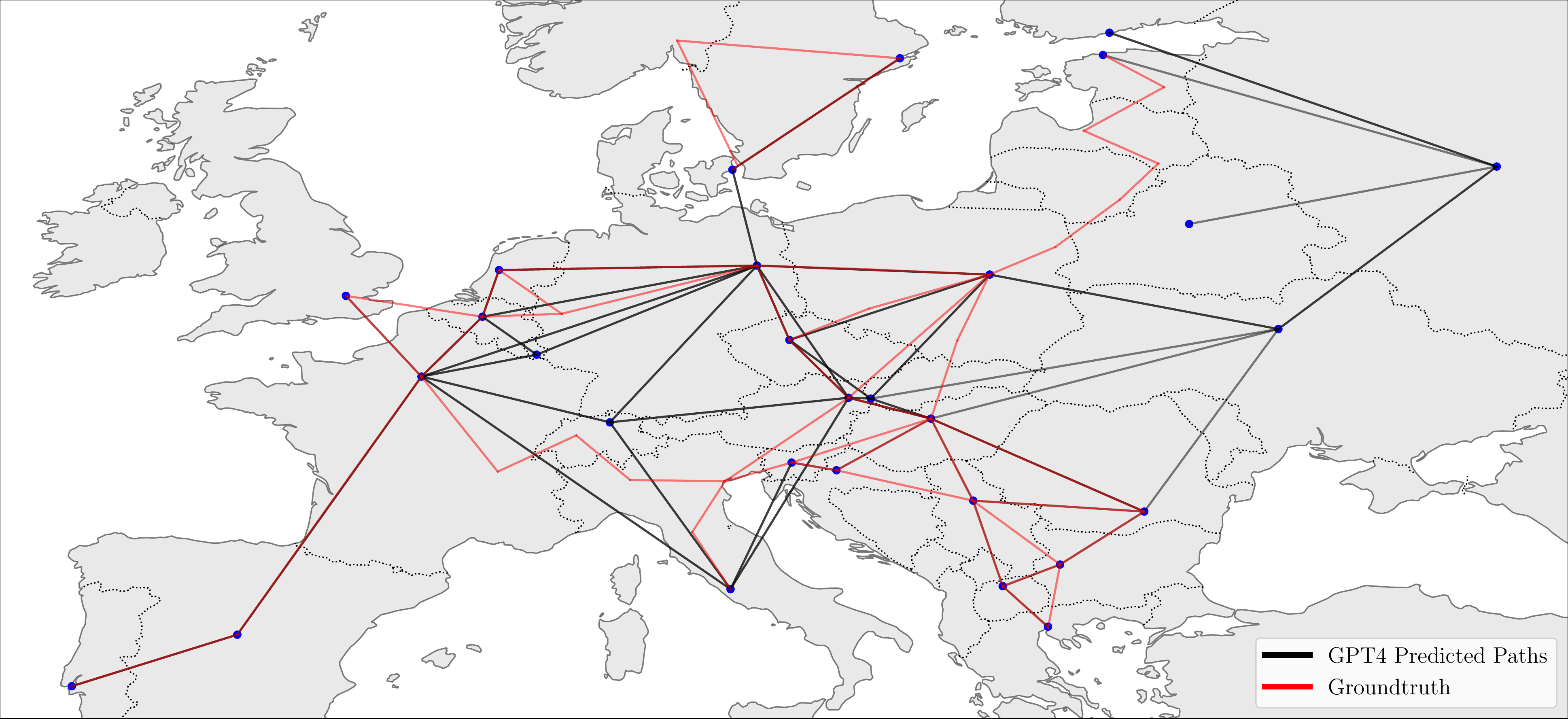}
    \caption{Subset of train network between the major cities of European countries including Russia.}
    \label{fig:euro_train}
\end{figure}

\textbf{Rail Network.} We plot the European Railway Network in Fig. \ref{fig:euro_train}. We first queried for the pairwise list of countries that are connected with each other, and next the coordinates for a major train station in each country. The two part prompt is as follows,
\begin{formattedquote}
    Give me a list of pairwise countries in Europe that have direct railway connections between them.
    \\ \\
    Give me the lat/lon coordinates of the train station for each country in the list
\end{formattedquote}
Response:
\begin{formattedresponse}
    Here's a list of pairwise countries with direct railway connections between them. \\
    1. Austria - Germany\\
    2. Austria - Switzerland\\
    3. Austria - Italy\\
    .\\
    75. Ukraine - Romania\\
    76. United Kingdom - France\\
    \\
    Here's the list of coordinates for stations.\\
    1. Austria - Vienna Central Station (Wien Hauptbahnhof): Latitude: 48.1853, Longitude: 16.3759\\
    2. Belgium - Brussels Central Station (Bruxelles-Central): Latitude: 50.8450, Longitude: 4.3571\\
    .\\
    25. United Kingdom - London St Pancras International: Latitude: 51.5314, Longitude: -0.1262
\end{formattedresponse}

We plotted the coordinates and connected the lines between the pairwise countries. Contrary to the groundtruth, GPT-4 kept Russia as part of Europe. Moreover, we only ask for a single rail station from each country. This removes some major stations like Munich, Frankfurt, and Lyon.

\subsubsection{Itinerary Planning}

\begin{figure}[h]
    \centering
    \includegraphics[width=\textwidth]{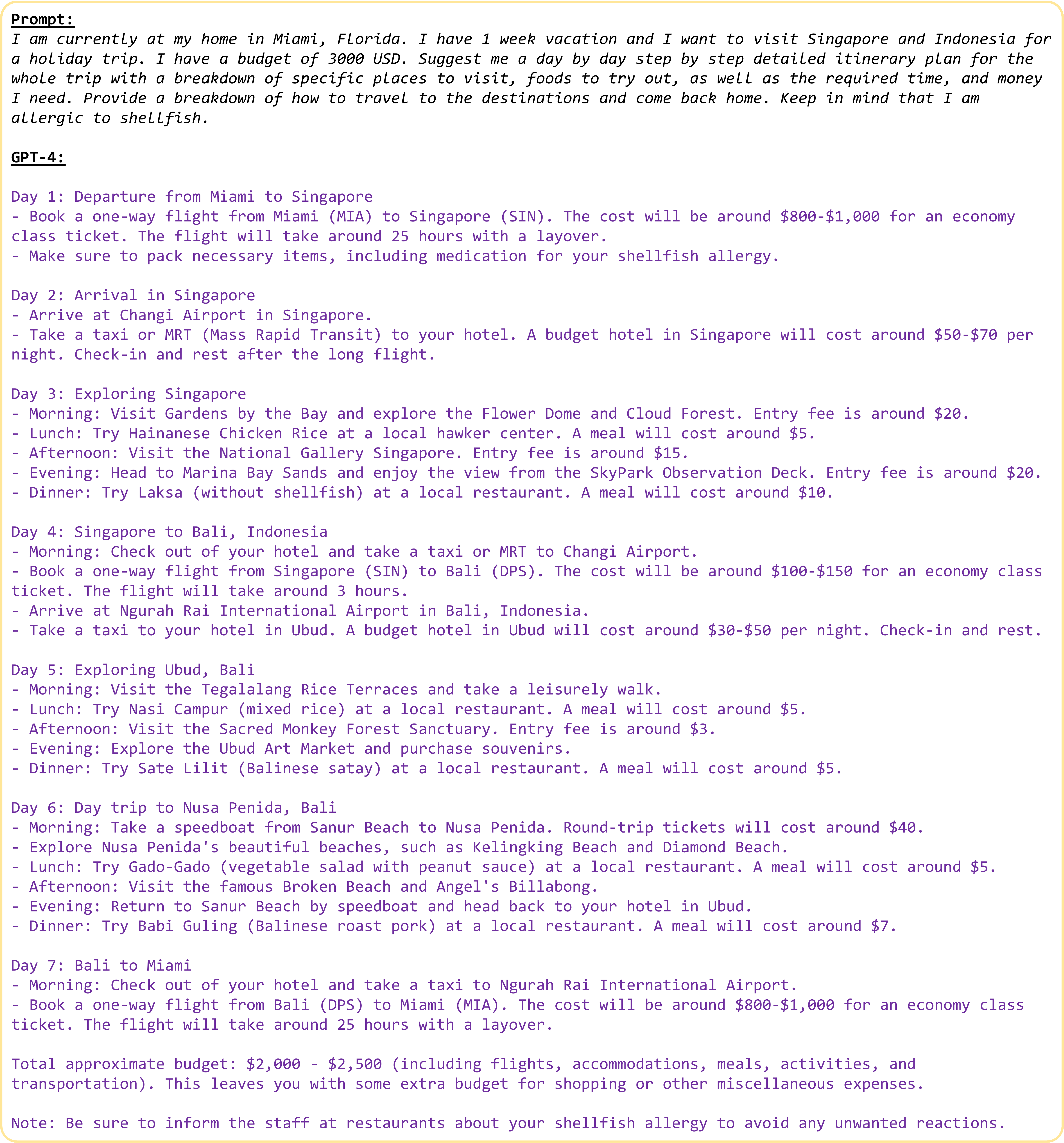}
    \caption{GPT-4 response for a 1 week itinerary for a trip to Singapore and Indonesia.}
    \label{fig:bali}
\end{figure}

Itinerary prompts involve specifying the start location, destination location, intermediate waypoints, budget limitations, duration of travel, and any additional  constraints. The prompt for the 8 days itinerary in Ireland is as follows:
\begin{formattedquote}
    I am currently at my home in Miami, Florida. I have 1 week vacation and  a budget of 2000 USD. Suggest me a travel itinerary for a trip to Ireland. Provide day by day step by step detailed itinerary plan for the whole trip with a breakdown of specific places to visit, foods to try out, as well as the required time, and money I need. Provide a breakdown of how to travel to the destinations and come back home.
\end{formattedquote}
We also query for a one week vacation combining Singapore and Indonesia, starting from Miami with more specific prompt and constraints. Fig. \ref{fig:bali} shows the response which includes detailed guide on travel, food, cost per day, as well as recommendations adhering to the allergy and food constraints. This shows GPT-4's capabilities as a personalized travel agent.

\clearpage

\subsubsection{Abstract Routing}

\textbf{Every node}

We used the following prompt for finding the optimum route that visits every edge in the graph.

\begin{formattedquote}
    I am visiting Hong Kong for a holiday. There are 8 tourist attractions I'd like to see, all are within walking distance.  The attractions are labelled A,B,C,E,F,G,H,I. For each attraction I will state the other attractions its directly linked to and how long it takes to walk between the two. My hotel is at point D. I will be starting and finishing from this point.
    \newline
    A-B = 5
    A-C = 1
    B-C = 8
    B-E = 3
    B-F = 4
    D-E = 2
    D-G = 14
    E-G = 9
    F-G = 1
    G-H = 4
    G-I = 2
    I-H = 1
    \newline
    Note, each of these pathways is bidirectional, and it takes the same time to walk in each direction.
    \newline
    What is the minimum amount of time I need to spend walking to visit all the attractions? What is my optimal route, starting and finishing at my hotel.
\end{formattedquote}

\textbf{Every edge}

We used the following prompt for finding the optimum route that visits every node in the graph.

\begin{formattedquote}
    I am visiting Hong Kong for a holiday. There are 8 tourist attractions I'd like to see, all are within walking distance. The attractions are labelled A,B,C,E,F,G,H,I. For each attraction I will state the other attractions its directly linked to (via a walking path) and how long it takes to walk between them. My hotel is at point D. I will be starting and finishing from this point. I also want to walk along all of the walking paths available.
    \newline
    A-B = 5
    A-C = 1
    B-C = 8
    B-E = 3
    B-F = 4
    D-E = 2
    D-G = 14
    E-G = 9
    F-G = 1
    G-H = 4
    G-I = 2
    I-H = 1
    \newline
    Note, each of these pathways is bidirectional, and it takes the same time to walk in each direction.
    \newline
    What is the minimum amount of time I need to spend walking to ensure that I visit all the attractions and walk along all of the possible paths? What is my optimal route, starting and finishing at my hotel.
\end{formattedquote}

\textbf{Graph Description}
We verify that GPT-4 correctly interprets the description of the graph by asking for the adjacency matrix from the descriptions above:

Prompt:
\begin{formattedquote}
    Can you give me the adjacency matrix for the graph
\end{formattedquote}

Response:
\begin{formattedresponse}
    ~ ~A  B  C  D  E  F  G  H  I \\
    A  0  5  1 -1 -1 -1 -1 -1 -1\\
    B  5  0  8 -1  3  4 -1 -1 -1\\
    C  1  8  0 -1 -1 -1 -1 -1 -1\\
    D -1 -1 -1  0  2 -1 14 -1 -1\\
    E -1  3 -1  2  0 -1  9 -1 -1\\
    F -1  4 -1 -1 -1  0  1 -1 -1\\
    G -1 -1 -1 14  9  1  0  4  2\\
    H -1 -1 -1 -1 -1 -1  4  0  1\\
    I -1 -1 -1 -1 -1 -1  2  1  0
\end{formattedresponse}

GPT-4 also correctly returns the adjacency list:

Prompt:
\begin{formattedquote}
    Convert this into an adjacency list
\end{formattedquote}

Response:
\begin{formattedresponse}
    An adjacency list representation of the given adjacency matrix is as follows:\\
    A: [(B, 5), (C, 1)] \\
    B: [(A, 5), (C, 8), (E, 3), (F, 4)] \\
    C: [(A, 1), (B, 8)] \\
    D: [(E, 2), (G, 14)] \\ 
    E: [(B, 3), (D, 2), (G, 9)] \\ 
    F: [(B, 4), (G, 1)] \\
    G: [(D, 14), (E, 9), (F, 1), (H, 4), (I, 2)] \\
    H: [(G, 4), (I, 1)] \\
    I: [(G, 2), (H, 1)] \\
\end{formattedresponse}

\textbf{Alternative wording for Riddle}

We reword the riddle used in the main paper, altering the context to provide a second assessment of GPT-4's ability to answer the riddle correctly with less chance of the exact question/answer being seen during training.

Prompt:
\begin{formattedquote}
    You’re driving on the
    surface of the Earth. After driving
    one mile south, one mile west
    and one mile north, you end up
    exactly where you started.
    Where are you?
\end{formattedquote}

Response:
\begin{formattedresponse}
    There are actually multiple locations from which you could start and end up at the same place after driving one mile south, one mile west, and one mile north.

    The North Pole: If you start a bit more than one mile south of the North Pole, you can drive one mile south to reach the North Pole, then drive one mile west (which would actually just have you moving in a circle around the North Pole, because all directions point south from there), and then drive one mile north to return to your original starting point.
    
    Certain points near the South Pole: This is more complex. Imagine a circle around the South Pole that has a circumference of 1 mile. If you start 1 mile north of this circle, you can drive 1 mile south to reach the circle, then drive 1 mile west, which will take you all the way around the circle and back to the same point (because the circumference is 1 mile), and then drive 1 mile north to return to your original starting point. This same principle applies to any circle near the South Pole that has a circumference less than or equal to 1 mile. For those smaller circles, driving ``west" for 1 mile would involve making one or more complete laps around the circle.
    
    Note that these answers are based on an idealized, spherical model of the Earth. The real world is more complicated due to the Earth's oblate shape, and the presence of continents and other geographical features.
\end{formattedresponse}

\subsubsection{Landmarks}

We use two different prompts to generate SVG code for the outline of landmarks.

\textbf{Isolated Landmarks}

For landmarks that are isolated and far from other landmarks, we used these prompts:

\begin{formattedquote}
    What famous landmark is at this location: \textit{<Landmark\_Lat/Lon\_Coordinates>}? Provide the SVG code for a detailed outline of the landmark.
\end{formattedquote}

and

\begin{formattedquote}
    What famous landmark is at this location: \textit{<Landmark\_Lat/Lon\_Coordinates>}? Provide the SVG code for a simple outline of the landmark.
\end{formattedquote}

Where \textit{\textless Landmark\_Lat/Lon\_Coordinates\textgreater} are the coordinates for the Statue of Liberty and Giza Pyramids Complex. Coordinates were taken from Wikipedia and Google Maps. For these landmarks, we find GPT-4 is able to accurately identify the landmark from the coordinates and provide the SVG code for a recognisable outline.

\textbf{Dense Landmarks}

For landmarks that are located closely with others, we specify the landmark name in  the prompt rather than providing a location. For example, ``London skyline'' is not possible to specify using a single coordinate, and the Gardens by the Bay are located too close to the Marina Bay Sands hotel for a single coordinate to distinguish the two. 

For the London Skyline, we used this prompt:

\begin{formattedquote}
    Provide the code for an SVG of the detailed outline of the London skyline. Use colour where relevant. Ensure that it's clearly recognisable as London from the outline. Include the most famous buildings.
\end{formattedquote}

For the Gardens by the Bay, we used this prompt:

\begin{formattedquote}
    Provide the code for an SVG outline for the Gardens by the Bay in Singapore.
\end{formattedquote}

\begin{figure}[h]
  \centering
  \begin{subfigure}[b]{0.23\textwidth}
  \centering
    \includegraphics[height=2.5cm]{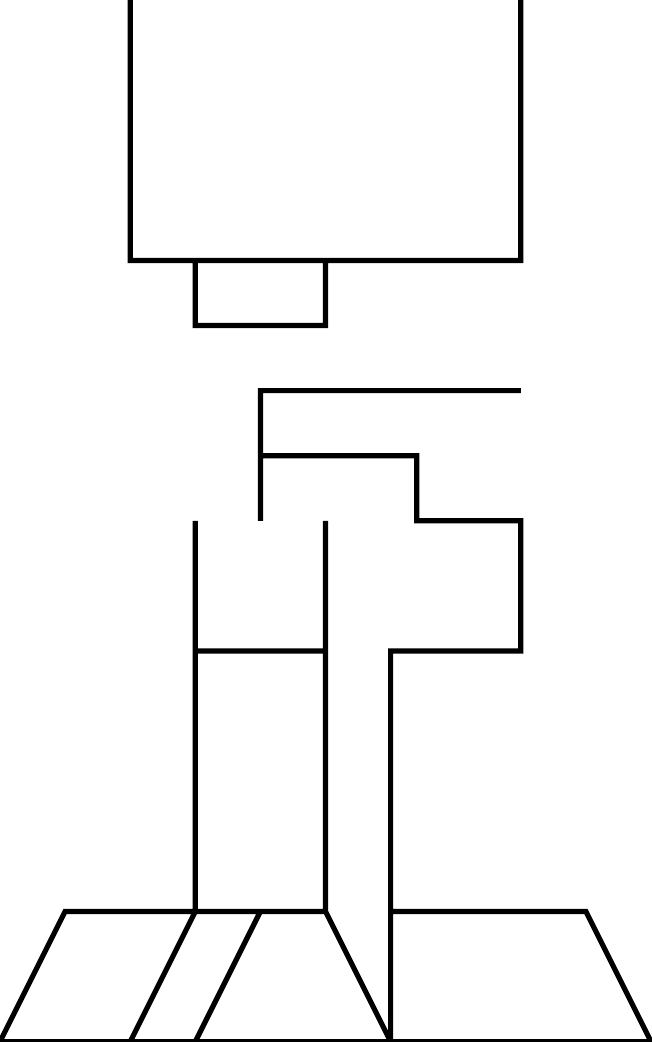}
    \caption{Eiffel Tower.}
    \label{fig:l1}
  \end{subfigure}
  \begin{subfigure}[b]{0.23\textwidth}
  \centering
    \includegraphics[width=3cm]{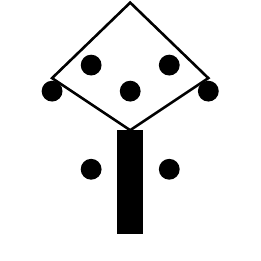}
    \caption{St. Basil's Cathedral.}
    \label{fig:l2}
    \end{subfigure}
    \begin{subfigure}[b]{0.23\textwidth}
    \centering
    \includegraphics[width=3cm]{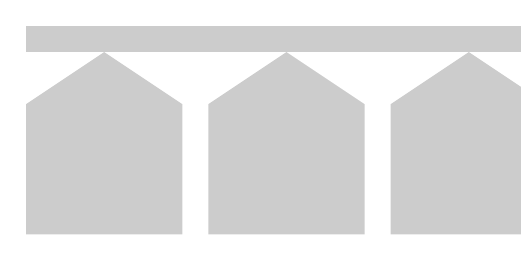}
    \caption{Marina Bay Sands.}
    \label{fig:l3}
  \end{subfigure}
\begin{subfigure}[b]{0.23\textwidth}
  \centering
    \includegraphics[width=3cm]{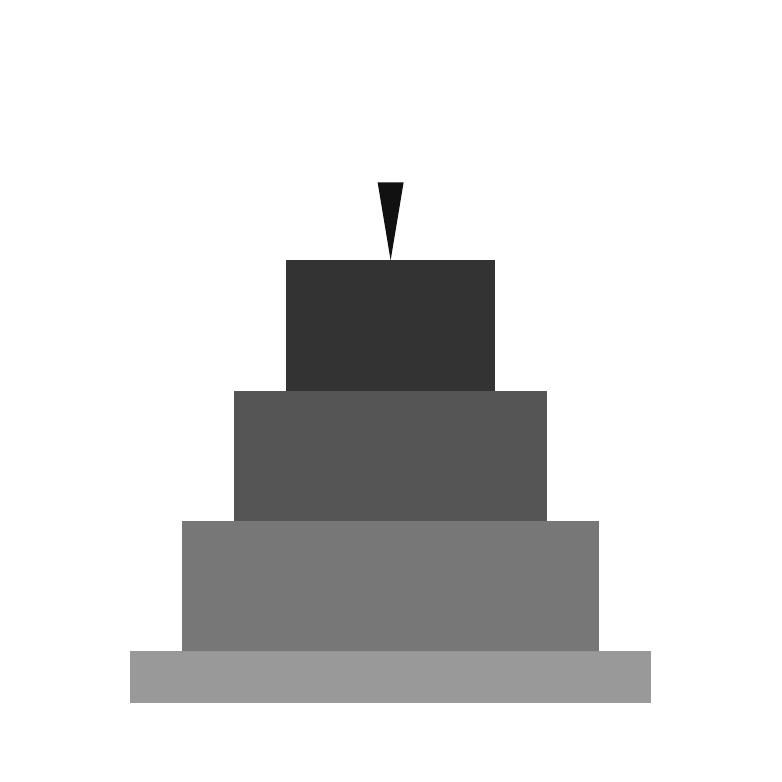}
    \caption{Temple of Heaven.}
    \label{fig:l4}
  \end{subfigure}
  \caption{\textbf{Additional Landmark SVGs}. (Countries (a)-(d): France, Russia, Singapore, China).}
  \label{fig:landmarks_additional}
\end{figure}

We include GPT-4's SVG outlines for additional landmarks that we experimented with in Fig. \ref{fig:landmarks_additional}.

\subsubsection{Multi-criteria Place Retrieval}

We used the following prompt for our experiment on multi-criteria place retrieval:
\begin{formattedquote}
    Name all places in the world where skiing and surfing is possible at the same day in December? Provide a python list in the format [0.00000N, 0.00000E].
\end{formattedquote}
\begin{formattedquote}
    Name all places in the world where skiing and taking a bath in natural hot springs is possible. Provide a python list in the format [0.00000N, 0.00000E].
\end{formattedquote}
\begin{formattedquote}
    Name all places in the world where safe hiking above 3000m is possible in December. Provide a python list in the format [0.00000N, 0.00000E].
\end{formattedquote}
\begin{formattedquote}
    Name all places in the world close to a volcano and a city with a population above 1 million. Provide a python list in the format [0.00000N, 0.00000E].
\end{formattedquote}
\begin{formattedquote}
    Name all places in the world that lie in a country that has won FIFA World Cup and where a large cat be seen in the wild. Provide a python list in the format [0.00000N, 0.00000E].
\end{formattedquote}
\begin{formattedquote}
    Name all places in the world where whales can be watched and a coastal hike is possible. Provide a python list in the format [0.00000N, 0.00000E].
\end{formattedquote}

We found that asking for a specific coordinate format helped as GPT-4 would sometimes generate inconsistent coordinates otherwise. If the output was not only provided as a Python list, we extracted the coordinates manually.

\subsubsection{Supply Chains}

We used the following prompt for analysing the global semiconductor supply chain:

\begin{formattedquote}
    I want to construct a map of the semiconductor supply chain, end-to-end. Please provide the latitude and longitude coordinates and names of the key elements in the supply chain, including design, manufacturing, materials, equipment + tools, etc. If you don't know any coordinates exactly just estimate, every point needs coordinates. 
\end{formattedquote}

We then ask for assistance with plotting:
\begin{formattedquote}
    Provide a python script that uses the Cartopy library to plot these locations on a world map. Use different colours for the steps 1-6. Annotate each location with the company/industry located there.
\end{formattedquote}

\subsubsection{Natural World}

\textbf{Wildlife Ranges}

We start with the initial prompt:
\begin{formattedquote}
    What are the different subspecies of tiger?
\end{formattedquote}

We then repeatedly use the following prompt, replacing \textit{tiger subspecies} with each subspecies returned in the first answer:
\begin{formattedquote}
    I want to create an outline of the range of the \textit{tiger subspecies} that I can overlay on a world map that I have. Provide me with latitude and longitude coordinates for this approximate outline. Output the coordinates as a python list of tuples. Due to length constraints, just output the list. Ensure the outline does not cross over itself. If the range is not continuous, provide multiple outlines as separate lists.
\end{formattedquote}

\textbf{Bird Migrations}

We used the following prompt for evaluating GPT-4's knowledge of bird migrations:

\begin{formattedquote}
    I want to plot the migratory routes of various bird species. Please provide the latitude and longitude coordinates of the migratory route of the \textit{bird species}. Indicate the start and finish and provide coordinates for intermediate steps if necessary. If multiple routes exist, provide separate lists of coordinates.
    \newline
    Output the coordinates as a python list of tuples.

\end{formattedquote}

\end{document}